\documentclass[10pt,twocolumn,letterpaper]{article}

\usepackage{iccv}
\usepackage{times}
\usepackage{epsfig}
\usepackage{graphicx}
\usepackage{amsmath}
\usepackage{amssymb}

\usepackage{amssymb}%
\usepackage{pifont}%
\usepackage{xcolor}

\usepackage{colortbl}

\usepackage{graphicx}
\usepackage{caption}
\usepackage{subcaption}
\usepackage{multirow}
\usepackage{amsmath}
\usepackage{amsfonts}
\usepackage{booktabs}
\usepackage{siunitx}
\usepackage{wrapfig}
\usepackage{bbm}
\usepackage{relsize}
\usepackage{color}

\definecolor{almond}{rgb}{0.94, 0.87, 0.8}

\usepackage[accsupp]{axessibility}  %

\usepackage[pagebackref=true,breaklinks=true,letterpaper=true,colorlinks,bookmarks=false]{hyperref}

\iccvfinalcopy %

\def\iccvPaperID{7072} %

\ificcvfinal\fi

\usepackage{cleveref}[capitalise]
\crefname{section}{Section}{Sections}
\crefname{figure}{Figure}{Figures}
\crefname{table}{Table}{Tables}
\crefname{equation}{Eq.}{Eqs.}
\Crefname{section}{Section}{Sections}

\begin{document}

\title{In-Style: Bridging Text and Uncurated Videos with Style Transfer \\for Text-Video Retrieval}

\author{%
    Nina Shvetsova\thanks{Equal contribution.} $^{1,2   ,3}$ \quad
    Anna Kukleva$^{*2}$  \quad
    Bernt Schiele$^{2}$  \quad
    Hilde Kuehne$^{1,3,4}$ \\
    \small{
    $^1$Goethe University Frankfurt,   
    $^2$Max-Planck-Institute for Informatics,
    $^3$University of Bonn,
    $^4$MIT-IBM Watson AI Lab
    } \\
    \small{
    \texttt{\{nshvetso,akukleva\}@mpi-inf.mpg.de}}  
}

\maketitle
\ificcvfinal\thispagestyle{empty}\fi

\newcommand{\myparagraph}[1]{\vspace{2pt}\noindent{\bf #1}}

\vspace{-1mm}
\begin{abstract}

Large-scale noisy web image-text datasets have been proven to be efficient for learning robust vision-language models. However, when transferring them to the task of video retrieval, models still need to be fine-tuned on hand-curated paired text-video data to adapt to the diverse styles of video descriptions. To address this problem without the need for hand-annotated pairs, we propose a new setting, text-video retrieval with uncurated \& unpaired data, that during training utilizes only text queries together with uncurated web videos without any paired text-video data. To this end, we propose an approach, In-Style, that learns the style of the text queries and transfers it to uncurated web videos. Moreover, to improve generalization, we show that one model can be trained with multiple text styles. To this end, we introduce a multi-style contrastive training procedure that improves the generalizability over several datasets simultaneously. We evaluate our model on retrieval performance over multiple datasets to demonstrate the advantages of our style transfer framework on the new task of uncurated \& unpaired text-video retrieval and improve state-of-the-art performance on zero-shot text-video retrieval.
\footnote{\href{https://github.com/ninatu/in_style }{github.com/ninatu/in\_style} \\ To be published at ICCV 2023. Cite as:
Nina Shvetsova, Anna Kukleva, Bernt Schiele, Hilde Kuehne. 
``In-Style: Bridging Text and Uncurated Videos with Style Transfer for Text-Video Retrieval''.
In: \textit{Proceedings of the IEEE/CVF International Conference on Computer Vision}, 2023.
}

\end{abstract}
\section{Introduction}
\vspace{-1mm}
\label{section:intro}

\begin{figure}[!t]
\begin{center}
\includegraphics[scale=0.15]{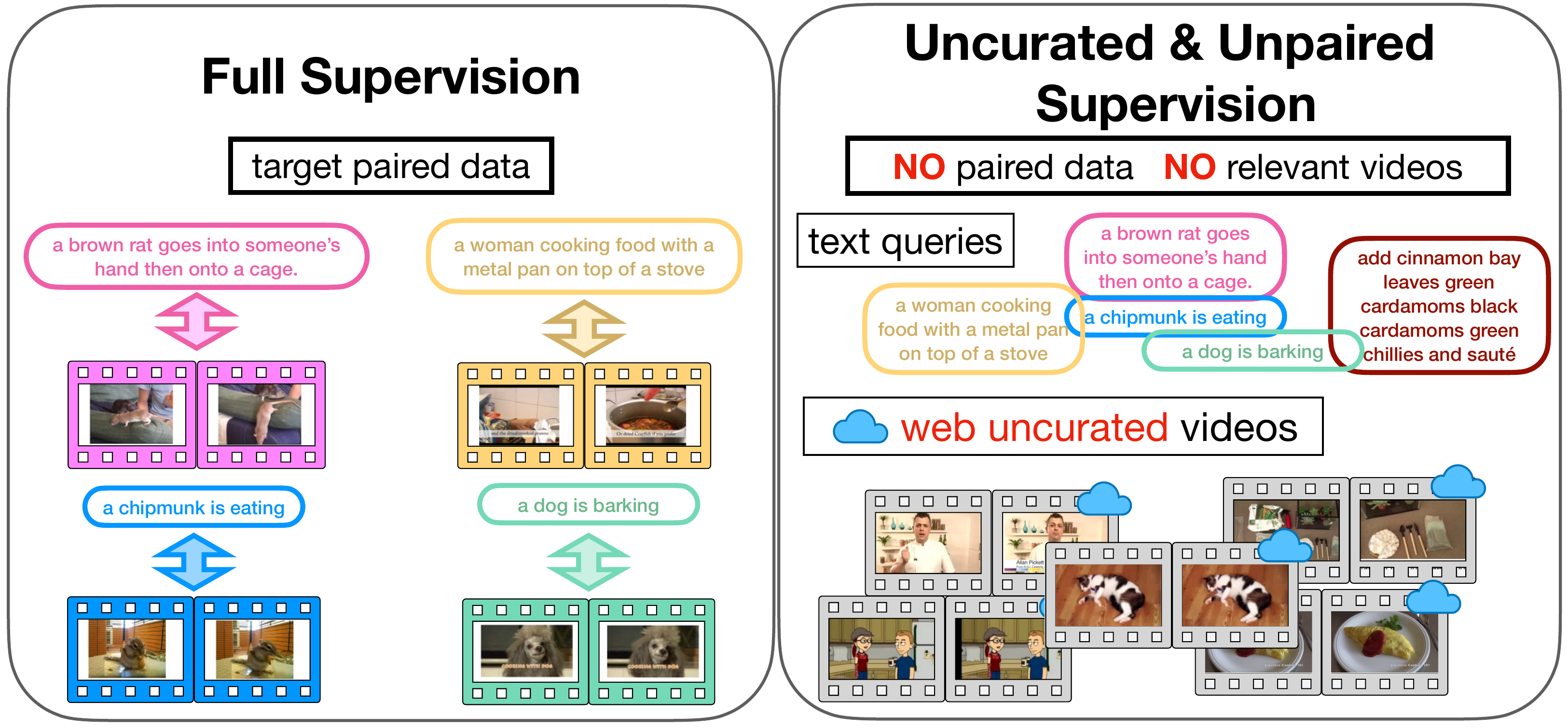}
\end{center}
\vspace{-1.3em}
\caption{ \small{\textbf{Training data for supervised and our uncurated \& unpaired settings for text-video retrieval.} \textbf{Left}: standard supervised text-video retrieval. Aligned and paired data is given for training with the same distribution as the target test set. \textbf{Right}: our uncurated \& unpaired text-video retrieval. No paired data is available during training, only text queries, whereas to support training, we use uncurated web videos.}
}
\vspace{-1.5em}
\label{fig:teaser}
\end{figure}

Vision-language retrieval refers to the task of retrieving an image or a video from a large data pool given a textual description of the content. The field of text-image retrieval has seen remarkable progress, mainly spurred by the combination of image and text models trained on large-scale web collections~\cite{radford2021learning, li2022blip} of image-text pairs. While advances in video retrieval also rely on pre-trained image-language models, which serve for better task transfer, most systems still require a fine-tuning on downstream data. This requires hand-annotated text-video pairs, namely trimmed segments of larger videos that are precisely described by the corresponding texts, for the training and testing of each target downstream dataset. 
Collecting such aligned pairs of text and videos can be time- and cost-intensive, and particularly gathering videos that comply with national regulations and copyright can be a challenge. Also, in the case of relying on free web content, some videos can become unavailable over time while the respective curated annotations stay available for download but do not have matching videos.

To address this problem, we propose a new setup, \textit{text-video retrieval with uncurated \& unpaired data}, assuming the availability of text queries only and without related videos during training (Figure~\ref{fig:teaser}). The setting is motivated by the fact that it can be considered easier to collect or generate text data, e.g., by producing topic-specific text queries rather than providing a video to match a specific context. To allow the training of a text-video retrieval system based on the given text, we assume to have access to an uncurated video collection as the only source of available videos.  

As different domains and datasets contain diverse styles of textual descriptions of videos, we propose a novel method, \textit{In-Style}, to transfer the caption style of given text queries to uncurated web videos, which can be from a deviating distribution compared to the given text queries. To transfer the style of the text queries, we leverage large image-language models~\cite{li2022blip, radford2021learning} by creating pseudo pairs that correspond to the given text queries and videos from the uncurated collection by matching them in the shared embedding space~\cite{radford2021learning}. Thus, we identify a subset of videos that have more similarity to the text queries than the rest of the videos. We then adopt an image-to-text captioning model (a captioner) to mimic the style of our text queries by training with these pseudo pairs. The stylized captioner is now capable of producing relevant video descriptions in the desired style; therefore, we re-annotate the web videos with the captioner to obtain aligned paired data; we call them generated pairs. Finally, we show that generated pairs help to adapt models pre-trained on large-scale web data~\cite{li2022blip, shvetsova2022everything} to the desired single or multiple styles of given text queries. 

We evaluate our model on text-video retrieval over 5 benchmark datasets. Specifically, we demonstrate the advantages of the In-Style method on the new task of uncurated \& unpaired text-video retrieval
with image-language~\cite{li2022blip} and video-language~\cite{shvetsova2022everything} pre-trained backbones. We show the generalization of the proposed approach by training a single model for multiple datasets at once leading to an improved state-of-the-art zero-shot text-video retrieval performance.  

We summarize our contributions in the following: 
(i) we introduce a new task of \textit{text-video retrieval with uncurated \& unpaired data} where during training, only text queries are available, whereas for the standard text-video retrieval task, paired text-video data is used; 
(ii) we propose a novel method, In-Style, to transfer the style of text queries in an unsupervised way, showing that style is an important component for language-based retrieval tasks; therefore, we repurpose large pre-trained image-language models to generate pseudo-captions of the same style for uncurated web videos; 
(iii) we demonstrate the advantages of our In-Style method for the new task over 5 different datasets with individual models for each dataset as well as one generalized model and we achieve state-of-the-art performance on zero-shot text-video retrieval.

\section{Related Work}

\myparagraph{Text-Video Retrieval.}
Text-video retrieval methods usually focus on learning modules that are able to capture relations between features 
from text and video modalities~\cite{yu2018joint, liu2019use, gabeur2020multi, chen2020fine, croitoru2021teachtext, dzabraev2021mdmmt, wang2021t2vlad}. Currently, many approaches leverage pre-training on large-scale video-text~\cite{bain2021frozen, miech2019howto100m, miech2020end} or image-text~\cite{lei2021less, li2022blip} datasets with a further adaptation of the backbone to individually downstream datasets. In this context, ClipBERT~\cite{lei2021less} proposed sparse sampling instead of using dense full-length videos that allow lightweight training. However, foundation models~\cite{bommasani2021opportunities} such as CLIP~\cite{radford2021learning}, combine the success of transformer architectures~\cite{dosovitskiyimage} using a contrastive objective~\cite{oord2018representation} and being trained on large collections of text-image pairs from the web, providing a strong zero-shot~\cite{luo2022clip4clip, portillo2021straightforward} baseline on downstream tasks that outperforms many previous methods. Therefore, more recent approaches focus on adapting text-image CLIP pre-trained models for text-video retrieval~\cite{gorti2022x, bogolin2022cross, fang2021clip2video, gao2021clip2tv, liu2022ts2}. X-pool~\cite{gorti2022x} introduces cross-modal attention to reason between text and frames of a video, TS2-Net~\cite{liu2022ts2} proposes dynamic adjustments over temporal and spatial token dimensions, which allows fine-tuning spatial model on video data without architecture changes. Another way to leverage foundation models is to enhance training data~\cite{wu2022cap4video, zhao2022lavila}. Cap4Video~\cite{wu2022cap4video} generates auxiliary captions for available curated training videos by using ZeroCap~\cite{tewel2022zerocap} that optimizes GPT-2~\cite{radford2019language} text generation using a CLIP-based loss~\cite{radford2021learning}. LaViLa~\cite{zhao2022lavila} proposes to generate additional narrations for a dense coverage of long videos from the Ego4D dataset~\cite{ego4degocentric, grauman2022ego4d} by fine-tuning a pre-trained large language model~\cite{radford2019language} on existing annotated text-video paired data. In contrast, we propose to exclude pre-annotated text-video paired data from the training and, relying on text descriptions only, generate text-video pairs leveraging uncurated web videos while transferring the style of original captions.  

\myparagraph{Large-scale Multimodal Pre-training.}
Representation learning~\cite{radford2021learning, li2022blip, chen2021mocov3, he2022masked, zhao2022lavila, caron2021emerging, yu2022coca} aims to obtain general representations that improve performance on downstream tasks such as retrieval~\cite{luo2022clip4clip, portillo2021straightforward, yu2022coca, li2022blip}, classification~\cite{chen2021mocov3, zhao2022lavila, caron2021emerging}, 
segmentation~\cite{caron2021emerging}, question-answering~\cite{li2022blip, yu2022coca} and 
captioning~\cite{li2022blip, yu2022coca}. 
While some methods rely only on one modality such as images~\cite{caron2021emerging, he2022masked} or text~\cite{radford2019language}, 
there is also increasing interest in multi-modal representations~\cite{radford2021learning, li2020unicoder, su2019vl, li2020hero, luo2020univl, shvetsova2022everything} which require multi-modal aligned pairs.  However, the acquisition of human-annotated paired data is expensive; therefore, noisy web data~\cite{radford2021learning, miech2019howto100m} allows for significant scaling of such datasets. Many methods successfully utilize web image-text pairs~\cite{radford2021learning, jia2021align, yu2022coca}, whereas uncurated video-text pairs are not only harder to collect but are also more prone to misalignments. Therefore, efforts are made to align ASR (automatic speech recognition) with video frames via contrastive learning~\cite{miech2019howto100m, miech2020end, xu2021videoclip, zellers2021merlot} or in an unsupervised way~\cite{han2022temporal}. To overcome those issues, we propose to generate synthetic video descriptions with the desired caption style and train models on those captions instead of raw ASR text. 

For contrastive-based vision-language representation learning methods, the dual-encoder architecture is a common choice as it features two parallel branches for two modalities, which are contrasted against each other to learn a joint embedding space~\cite{radford2021learning, su2019vl, li2020hero, luo2020univl}. Recently, BLIP~\cite{li2022blip} and CoCa~\cite{yu2022coca} propose a unified multi-task contrastive-generative framework that combines contrastive and captioning objectives. These methods rely on both curated image-text and uncurated web image datasets, with BLIP additionally iteratively applying the generation and filtering of synthetic captions. Compared to those works, we adopt pre-trained image-language models for uncurated \& unpaired text-video retrieval by transferring the caption style directly on uncurated videos without any aligned data during training.

\begin{figure*}[]
\begin{center}
\includegraphics[scale=0.28]{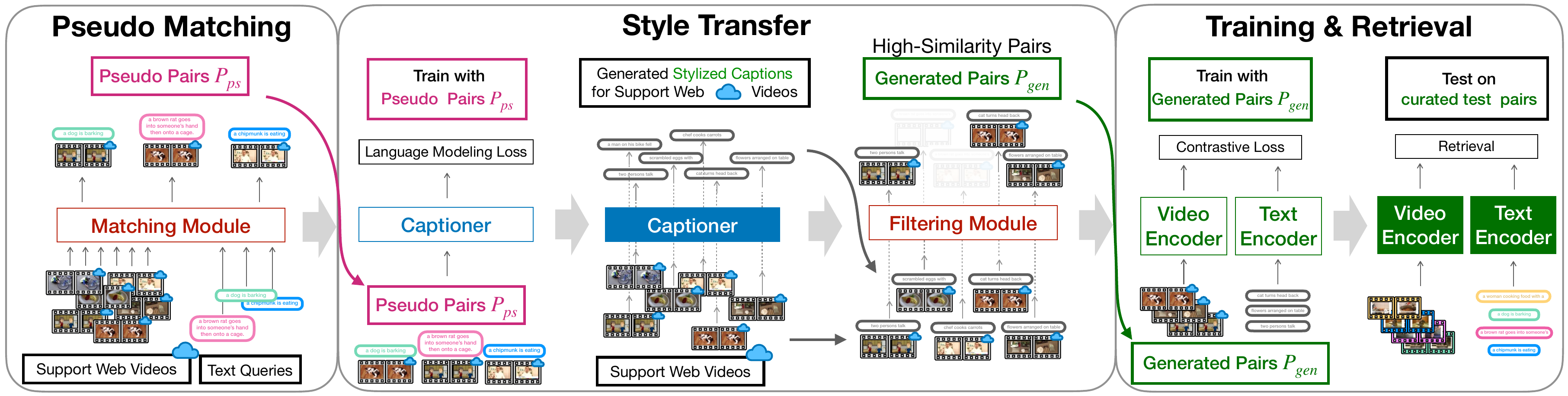}
\end{center}
\vspace{-1.4em}
\caption{ \small{ \textbf{The proposed In-Style method.} First, in the pseudo matching step, pseudo pairs $P_{ps}$, which consist of text queries and the most related web videos from the support set, are created. Style Transfer: the captioner is tuned with the obtained pseudo pairs $P_{ps}$ to adapt to the style of the given text queries. Next, new stylized captions are generated for all videos in the support set and then filtered to avoid noisy captions; the resulting set of the generated pairs $P_{gen}$ contains web videos and aligned captions of the desired style. 
To complete the retrieval task, we adapt the dual video-text encoder model with the generated pairs $P_{gen}$ and evaluate it on the curated paired test set.  }
}
\vspace{-4mm}
\label{fig:multi_stage}
\end{figure*}

\section{Uncurated \& Unpaired Text-Video Retrieval}
\label{subsec:unsupervised_text_video_retrieval}

In this section, we introduce the proposed uncurated \& unpaired text-video retrieval training setup. Typically, models for text-video retrieval are trained on \textit{paired} text-video data. Given a set of pairs of captions $t_i$ and corresponding videos $v_i$: ${\{(t_i, v_i)\}} \in D$, where $D$ is a data distribution, the goal is to learn a similarity function $s(t_i, v_j)$ that calculates the similarity between the caption $t_i$ and the video $v_j$.  The training can be done from scratch, but typically pre-trained image-language~\cite{radford2021learning, li2022blip} or video-language models~\cite{luo2020univl, miech2019howto100m} are fine-tuned on the target paired text-video data and then evaluated on the test set from the same distribution $D$~\cite{liu2022ts2, fang2021clip2video}. If the evaluation is performed on multiple datasets, the model is usually fine-tuned for each dataset individually. 

In contrast, we propose \textit{a text-video retrieval with uncurated \& unpaired data}, where only target text queries are available during training without any videos. More precisely, given a set of text descriptions $\{t_i\}$ from data distribution $D$, we aim to learn useful information about the similarity $s(t_i, v_j)$ in $D$ relying only on the textual descriptions. We further assume that a large set of freely accessible web videos $V' = \{v'_j\} \in D'$ without any paired text is available to support the training (such as videos of the HowTo100M dataset~\cite{miech2019howto100m}). We note that the data distribution $D'$ in the support video dataset can deviate from the distribution $D$.

Finally, to avoid training different models individually for each target dataset, we further consider learning a \textit{generalized} model that maintains the performance of individual models over a set of $K$ datasets of different caption styles and coming from different data distributions $D_1,..., D_K$.

\section{In-Style Method}

To address the task of uncurated \& unpaired text-video retrieval, we aim to transfer the style of the text queries (the only available curated information) to an uncurated web video dataset. To this end, we rely on web-scale pre-trained image-language models as a supervisory signal and leverage them as a matching module and pre-trained captioning model that we adapt throughout the training process. The steps of the proposed In-Style method are shown in \cref{fig:multi_stage}. The first step is \textit{Pseudo Matching}, described in \cref{sec:pseudo_text_video_pairs}, which matches the given text queries to the most relevant videos from the set of all uncurated web videos. The following \textit{Style Transfer} step (\cref{subsec:style_presrvation}) adapts the pre-trained captioning model (the captioner) to the target text style by training it on the previously obtained pseudo pairs. The captioner is then used to generate new style-adapted captions for all available web videos, which are then filtered to avoid too noisy pairs; we refer to the resulting filtered web videos with style-adapted video descriptions as generated pairs.  Finally, we adapt a pre-trained vision-language model for the task of text-video retrieval on the generated pairs (\cref{subsec:retrieval}). Moreover, in \cref{subsec:multidataset_training}, we propose the training of a generalized model on multiple styles of text queries at the same time and introduce a new contrastive objective, In-Style, that improves training on more than one text style at once.

\subsection{Pseudo Matching}
\label{sec:pseudo_text_video_pairs}

First, we obtain pseudo video-text pairs, with each pair containing one of the available text queries and the most relevant uncurated video from the web collection. For pseudo matching, we leverage image-language models such as CLIP~\cite{radford2021learning} or BLIP~\cite{li2022blip} that excel in zero-shot retrieval performance~\cite{luo2022clip4clip}. Such models usually follow a dual-encoder architecture: encoders $f_t$ and $f_v$ projects text $t$ and image $x$ into a common multimodal embedding space. The similarity of text and image is computed as a cosine similarity in this common space: $sim(t, x) = \frac{f_t(t)^\top f_v(x)}{\left\lVert f_t(t)\right\rVert \left\lVert f_v(x)\right\rVert}$. We use this metric to match the available text queries to the closest video. 

Since available videos can vary in overall duration (for example, five or more minutes) and cover a lot of different actions, we divide all videos into non-overlapping clips of $s$-seconds. We 
denote $V' = \{v'_j\}$ as a set of all such video clips. Then, we calculate a multimodal representation for each video clip $v'_j$ as an average representation of $m$ uniformly sampled frames of a video (see supplement). Using precomputed embeddings, we connect every caption $t_i$ with a video $v'$ with maximum similarity from available set of videos $V'$, such as: 
\begin{equation}
    v_i^p = \underset{v'_j  \in V'}{\arg\max}~\text{sim}(t_i, v'_j).
\end{equation}
To increase the diversity of matched videos, we do not allow multiple captions to match the same video clip; therefore, when video clip $v_i^p$ is matched, we exclude it from $V'$. Thus, we obtain a set of pseudo text-video pairs $P_{ps} = \{(t_i, v_i^p)\}$. In \cref{sec:experiments}, we show that this step allows us to introduce a weak supervision that may not find the exact match but provides a basis for further style transfer.

\subsection{Style Transfer}
\label{subsec:style_presrvation}

We aim to transfer the style of the given text queries to other unrelated web videos by generating new captions with the desired style. Inspired by the ability of language models conditioned on visual input~\cite{li2022blip} to generate plausible descriptions for diverse visual inputs, we propose to adapt the pre-trained image captioner $g$ using the obtained set of noisy pseudo text-video pairs $P_{ps}$. By doing this, we adapt the captioner to both the style of the captions as well as the style of the web videos. This allows us to generate new stylized captions $P_{gen}$  for the full support set of videos $V'$ using this captioner.

\myparagraph{Captioner.}
More specifically, we follow the BLIP~\cite{li2022blip} captioner architecture, which we extend for video captioning by conditioning the model not only on a single image but on a number of video frames. To this end, we apply the image encoder on each frame individually and inject a joint set of visual tokens into the text decoder model, which produces text in an autoregressive manner. We provide further details in the supplement. 
To train the captioner $g$ on the pseudo text-video pairs $P_{ps}$, we utilize the common language model loss
that optimizes cross-entropy loss between ground truth and predicted probabilities of the next token given a correct set of previous tokens in the sentence. Following BLIP, we also use label smoothing with parameter 0.1 while calculating cross-entropy. 

\myparagraph{Stylization of Captions.} For each video $v'_i \in V'$, we generate a caption $t^g_i = g(v'_i)$ with a captioner $g$ trained on pseudo pairs by using a nucleus sampling~\cite{holtzman2019curious}. Nucleus sampling was shown to generate more diverse and detailed captions than a beam search~\cite{vijayakumar2016diverse,li2022blip}. 

\myparagraph{Filtering.} As the captioner $g$ is adapted on pseudo pairs and shifts the model closer to a vocabulary of given text queries $\mathcal{D}$, some of the generated captions $t^g_i$ might be noisy and not descriptive for the web videos. Therefore, we further filter the generated pairs based on a similarity score $s(t^g_i, v'_i)$ utilizing the large pre-trained image-language dual encoders the same way as it was used for creating pseudo text-video pairs (Section~\ref{sec:pseudo_text_video_pairs}). Leaving only pairs with similarity higher a threshold  $s(t^g_j, v'_j) > th$, we obtain a paired set of web videos and stylized related captions $P_{gen} = \{(t^g_j, v'_j)\}$. 
In \cref{subsec:ablation}, we show that even a noisy set of pseudo pairs is enough to adapt a captioner for generating captions in a desired text style and that stylized captions combined with the following filtering provide a strong learning signal to boost the performance of retrieval in target distribution $D$.

\subsection{Training and Retrieval}
\label{subsec:retrieval}
\myparagraph{Single-Style Training.} To allow for text-video retrieval based on the stylized captions and the paired video data, we train a dual-encoder architecture~\cite{li2022blip} on the set of generated pairs $P_{gen}$ with the contrastive loss~\cite{oord2018representation}. 
We show that $P_{gen}$ provides better supervision than $P_{ps}$ or even a combination $P_{gen}$ + $P_{ps}$. Practically, we consider several pre-trained models: the image-text model BLIP~\cite{li2022blip}, which we adapted for video as described in Section~\ref{sec:pseudo_text_video_pairs}, as well as video-text model EAO~\cite{shvetsova2022everything}, which is pre-trained on the HowTo100M dataset with ASR-video pairs, which serve as noisy supervision. Following previous works, we use symmetric contrastive loss, which brings together text $t^g_i$ and video $v_i$ from a text-video pair $(t^g_i, v_i) \in P_{gen}$ (a positive pair) in shared video-text embedding space, and contrasting them on video and text from different pairs (negatives), that are pushed apart:
{\small
\begin{equation}
\begin{split}
L &=-\frac{1}{2B}\sum_{i=1}^B{(\log\frac{\exp(\frac{s(t^g_i, v'_i)}{\tau})}{\sum\limits_{j=1}^B{\exp(\frac{s(t^g_i, v'_j)}{\tau})}}} + \log\frac{\exp(\frac{s(v'_i,t^g_i)}{\tau})}{\sum\limits_{j=1}^B{\exp(\frac{s(v'_i, t^g_j)}{\tau})}})
\end{split}
\end{equation}}
where $\tau$ denotes a temperature parameter and $B$ is a number of pairs. 

For the fine-tuning of the BLIP model, we follow the original setup and utilize the extension of contrastive training with a momentum encoder and a queue that keeps more negatives as well as soft labels. To fine-tune the EAO model, we follow the respective setup without a momentum encoder or soft labels.

\begin{figure}[!t]
\begin{center}
\includegraphics[scale=0.29]{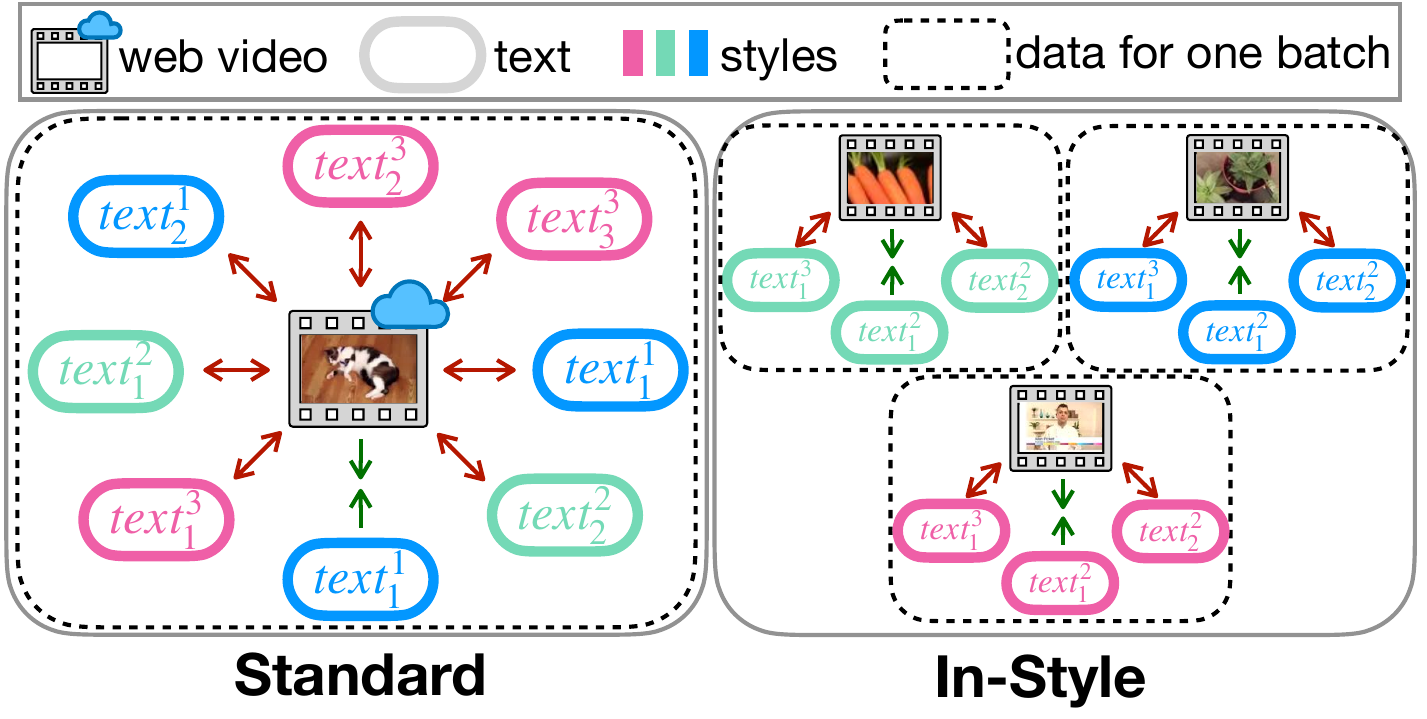}
\end{center}
\vspace{-1.7em}
\caption{ \small{\textbf{Multi-dataset training.} \textbf{Left}: Standard contrastive training with multiple datasets. \textbf{Right}: Our In-Style training procedure. Each batch consists of text queries that belong only to the same style. Note that we use only web videos from the support set; therefore, all videos are from the same distribution.
}
}
\vspace{-1.4em}
\label{fig:multi-style}
\end{figure}

\myparagraph{Multi-Style Training.}
\label{subsec:multidataset_training}
Finally, we consider training a generalized model on multiple sources of text queries coming from different data distributions $D_1, ..., D_K$.  Let's denote $P^1_{gen}, ..., P^N_{gen}$ set of generated pairs for the captions from $D_1, ..., D_N$ respectively. Here, different sources can have various styles that might highlight different aspects of videos in their captions (Table~\ref{tab:text_styles_short}). As an example, captions in the YouCook2 dataset~\cite{zhou2018towards} are more ``action''-oriented, e.g., ``combine macaroni sauce and cheese'' or ``stir in crushed tomatos'', while captions of the LSMDC dataset~\cite{rohrbach2015long} are third-person descriptions, e.g., ``Someone gazes at the beautiful animal'' or ``Someone chews the sweet'. 
In standard training~\cite{li2022blip}, all different styles with their matching videos would be present in contrastive loss together, which can lead to a mixture of different visual topics and text styles, which are easy to separate and which might include only a few hard negatives per sample. To avoid this possibly noisy setting, we propose to modify the training procedure and select video-caption pairs with captions from the same data source for contrastive loss. 
Formally, during training, we iterate over generated sets of pair $P^1_{gen}, ..., P^N_{gen}$ sampling a minibatch $\{(t^g_i, v'_i)\}_{i=1}^B$ from a single set $P_{gen} \in \{P^1_{gen}, ..., P^N_{gen}$\} and calculating loss $L(\{(t^g_i, v'_i)\}_{i=1}^B)$ performing one optimization step with a minibatch (Figure~\ref{fig:multi-style}). 
We note that for BLIP training, we keep separate queues for each set $P_{gen}$.

We show in ~\cref{sec:exp-multi-style-training} that this setting can be beneficial for learning a generalized model. %
Our intuition is that text queries with the same style provide stronger negatives for the model, allowing the model to concentrate on the content of the captions rather than a style.

\section{Experimental Evaluation}
\label{sec:experiments}

\begin{table*}
    \setlength{\tabcolsep}{3pt}
    \resizebox{1\linewidth}{!}{
    
    \begin{tabular}{@{}llc|cccc|cccc|cccc|cccc|cccc|cccc@{}}
    	\toprule
                 Pre-trained  & \multirow{2}{*}{Method} & \multirow{2}{*}{Supervision} & \multicolumn{4}{c}{MSR-VTT} & \multicolumn{4}{c}{YouCook2} & \multicolumn{4}{c}{DiDeMo} & \multicolumn{4}{c}{MSVD} & \multicolumn{4}{c}{LSMDC} & \multicolumn{4}{c}{Mean} \\
                Model & & & R1 & R5 & R10 & MR & R1 & R5 & R10 & MR & R1 & R5 & R10 & MR & R1 & R5 & R10 & MR & R1 & R5 & R10 & MR & R1 & R5 & R10 & MR \\
                \midrule 
                   \multirow{3}{*}{BLIP~\cite{li2022blip}} & Zero-shot & none & 34.1 & 60.2 & 70.6 & 3 & 6.0 & 16.2 & 23.1 & 70 & 28.2 & 52.0 & 62.7 & 5 & 38.8 & 64.8 & 74.0 & 2 & 14.5 & 29.3 & 36.4 & 32.5 & 24.3 & 44.5 & 53.4 & 22.5 \\ 
                   
                 & In-Style (ours) & only text & \textbf{36.2} & \textbf{61.8} & \textbf{71.9} & \textbf{3} & \textbf{8.6} & \textbf{21.6} & \textbf{30.0} & \textbf{37} & \textbf{32.1} & \textbf{61.9} & \textbf{71.2} & \textbf{3} & \textbf{44.8} & \textbf{72.5} & \textbf{81.2} & \textbf{2} & \textbf{16.1} & \textbf{33.6} & \textbf{39.7} & \textbf{25} & \textbf{27.6} & \textbf{50.3} & \textbf{58.8} & \textbf{14} \\ 
                  & \textcolor{gray}{GT fine-tuning} & \textcolor{gray}{T-V pairs} & 
                  \textcolor{gray}{42.9} & \textcolor{gray}{69.7} & \textcolor{gray}{78.9} & \textcolor{gray}{2} & \textcolor{gray}{12.6} & \textcolor{gray}{32.0} & \textcolor{gray}{43.6} & \textcolor{gray}{15} & \textcolor{gray}{40.2} & \textcolor{gray}{70.6} & \textcolor{gray}{79.3} & \textcolor{gray}{2} & \textcolor{gray}{48.1} & \textcolor{gray}{76.6} & \textcolor{gray}{85.0} & \textcolor{gray}{2} & \textcolor{gray}{23.8} & \textcolor{gray}{41.1} & \textcolor{gray}{50.9} & \textcolor{gray}{10} & \textcolor{gray}{33.5} & \textcolor{gray}{58.0} & \textcolor{gray}{67.5} & \textcolor{gray}{6.2} \\
                  \midrule 
                   \multirow{3}{*}{EAO~\cite{shvetsova2022everything}} & Zero-shot & none & 9.9 & 24.0 & 32.6 & 28 & 19.8 & 42.9 & 55.1 & 8 & 6.6 & 19.0 & 26.8 & 42 & 18.0 & 40.4 & 52.3 & 9 & 3.6 & 8.5 & 13.0 & 177 & 11.6 & 27.0 & 36.0 & 52.8 \\ 
                 
                 & In-Style (ours) & only text & \textbf{16.4} & \textbf{35.8} & \textbf{48.9} & \textbf{10} & \textbf{20.3} & \textbf{46.4} & \textbf{58.8} & \textbf{7} & \textbf{13.2} & \textbf{31.6} & \textbf{44} & \textbf{15} & \textbf{23.4} & \textbf{50} & \textbf{62.4} & \textbf{5} & \textbf{4.9} & \textbf{12.3} & \textbf{16.7} & \textbf{94} & \textbf{15.64} & \textbf{35.22} & \textbf{46.16} & \textbf{26.2} \\
                 
                    & \textcolor{gray}{GT fine-tuning} & \textcolor{gray}{T-V pairs} & 
                    \textcolor{gray}{22.8} & \textcolor{gray}{47.8} & \textcolor{gray}{60.3} & \textcolor{gray}{6} & \textcolor{gray}{26.7} & \textcolor{gray}{55.9} & \textcolor{gray}{68.6} & \textcolor{gray}{4} & \textcolor{gray}{19.2} & \textcolor{gray}{43.1} & \textcolor{gray}{54.4} & \textcolor{gray}{8} & \textcolor{gray}{25.1} & \textcolor{gray}{53.6} & \textcolor{gray}{65.7} & \textcolor{gray}{5} & \textcolor{gray}{8.9} & \textcolor{gray}{21.2} & \textcolor{gray}{29.4} & \textcolor{gray}{40} & \textcolor{gray}{20.5} & \textcolor{gray}{44.3} & \textcolor{gray}{55.7} & \textcolor{gray}{12.6} \\

         	\arrayrulecolor{black}\bottomrule
    \end{tabular}
    }
    \vspace{-0.23cm}
    \caption{\small{
    \textbf{Text-video retrieval with style transfer.}  Comparison between the upper bound, where the retrieval model trained with ground truth aligned text-video pairs (T-V pairs), zero-shot respective models (no style transfer or tuning), and our In-Style method, where we follow our new setting of \textit{uncurated \& unpaired text-video retrieval}} for style transfer based only on input text queries.
    \label{tab:main_results}
    }
    \vspace{-0.25cm}
\end{table*} 
\begin{table*}
    \setlength{\tabcolsep}{3pt}
    \resizebox{1\linewidth}{!}{
    
    \begin{tabular}{@{}l|cccc|cccc|cccc|cccc|cccc|cccc@{}}
    	\toprule
                Training Dataset & \multicolumn{4}{c}{MSR-VTT} & \multicolumn{4}{c}{YouCook2} & \multicolumn{4}{c}{DiDeMo} & \multicolumn{4}{c}{MSVD} & \multicolumn{4}{c}{LSMDC} & \multicolumn{4}{c}{Mean} \\
                & R1 & R5 & R10 & MR & R1 & R5 & R10 & MR & R1 & R5 & R10 & MR & R1 & R5 & R10 & MR & R1 & R5 & R10 & MR & R1 & R5 & R10 & MR \\
                \midrule 
                 MSR-VTT & \cellcolor{almond}36.2 & \cellcolor{almond}61.8 & \cellcolor{almond}71.9 & \cellcolor{almond}3 & 7.6 & 18.8 & 25.9 & 62 & 29.0 & 54.5 & 65.4 & 4 & 43.3 & 70.7 & 79.9 & 2 & 15.2 & 28.5 & 35.3 & 31 & 26.3 & 46.9 & 55.7 & 20.4 \\
                 
                 YouCook2 & 31.5 & 55.5 & 64.4 & 4 & \cellcolor{almond}8.6 & \cellcolor{almond}21.6 & \cellcolor{almond}30.0 & \cellcolor{almond}\textbf{37} & 25.1 & 53.9 & 65.2 & 4 & 41.1 & 67.3 & 76.8 & 2 & 14.2 & 28.8 & 36.9 & 30 & 24.1 & 45.4 & 54.7 & 15.4 \\
                 
                 Didemo & 34.0 & 58.5 & 68.9 & 3 & 6.8 & 17.2 & 24.5 & 69 & \cellcolor{almond}32.1 & \cellcolor{almond}61.9 & \cellcolor{almond}\textbf{71.2} & \cellcolor{almond}\textbf{3} & 43.7 & 71.6 & 80.5 & 2 & 16.6 & 30.5 & 38.4 & 28 & 26.6 & 47.9 & 56.7 & 21 \\
                 
                 MSVD & 36.0 & 59.4 & 69.5 & 3 & 6.4 & 16.4 & 23.6 & 70 & 27.0 & 54.9 & 65.0 & 4 & \cellcolor{almond}\textbf{44.8} & \cellcolor{almond}72.5 & \cellcolor{almond}81.2 & \cellcolor{almond}\textbf{2} & 14.5 & 27.4 & 34.8 & 32 & 25.7 & 46.1 & 54.8 & 22.2 \\ 
                 
                 LSMDC & 33.9 & 60.3 & 69.9 & 3 & 7.1 & 18.1 & 25.6 & 68 & 31.7 & 59.9 & 69.1 & 3 & 44.6 & 71.7 & 80.0 & 2 & \cellcolor{almond}16.1 & \cellcolor{almond}\textbf{33.6} & \cellcolor{almond}39.7 & \cellcolor{almond}\textbf{25} & 26.6 & 48.7 & 56.8 & 20.2 \\
                 Target dataset (mean over diagonal) & - & - & - & - & - & - & - & - & - & - & - & - & - & - & - & - & - & - & - & -  & \cellcolor{almond} 27.5 & \cellcolor{almond}\textbf{50.3} & \cellcolor{almond}58.8 & \cellcolor{almond}\textbf{14} \\ 
                 \midrule
                 All five datasets -- standard training & 36.4 & 62.1 & 71.8 & 3 & \textbf{8.7} & 21.4 & 29.4 & 44 & 31.4 & \textbf{62.5} & 71.2 & 3 & 44.7 & 72.9 & 81.5 & 2 & 16.3 & 31.9 & 39.5 & 25 & 27.5 & 50.2 & 58.7 & 15.4 \\
                 \textbf{All five datasets -- In-Style (ours)} & \textbf{36.7} & \textbf{61.9} & \textbf{72.3} & \textbf{3} & 8.5 & \textbf{21.8} & \textbf{30.4} & 38.5 & \textbf{32.6} & 61.8 & \textbf{71.2} & \textbf{3} & 44.7 & \textbf{73.1} & \textbf{82.0} & \textbf{2} & \textbf{16.6} & 32.2 & \textbf{39.8} & 26 & \textbf{27.8} & 50.2 & \textbf{59.1} & 14.5 \\

         	\arrayrulecolor{black}\bottomrule
    \end{tabular}
    }
    \vspace{-0.23cm}
    \caption{
    \small{\textbf{Generalization performance of different models over all datasets.} Mean denotes an average of R1, R5, R10, MR over 5 datasets, correspondingly. 
    \textbf{Top}: the proposed In-Style method with the input text queries only from one \colorbox{almond}{respective} training dataset. \textbf{Bottom:} training with 5 different text query styles. Comparison between standard multi-dataset training and proposed In-Style procedure.  
    \label{tab:main_generalization}
    }
    }
    \vspace{-0.4cm}
\end{table*} 
\begin{table}[t]
    \small
    \setlength{\tabcolsep}{3pt}
    \centering
    \resizebox{1.0\linewidth}{!}{
    
    \begin{tabular}{@{}c|m{6.5cm}@{}}
    	\toprule
                Dataset  & Examples\\ 
                \midrule
                   & 1) The peoples are sharing their view on this car of \\
                  MSR-VTT & different models \\
                 ($\sim$43 symbols & 2) Someone is showing the ingredients for a dish \\
                 in a text) & they are going to make \\
                 & 3) A man is playing an instrument \\
                \midrule

YouCook2 & 1) Combine macaroni sauce and cheese \\
 ($\sim$39 symbols& 2) Grate and cube potatoes \\
 in a text) & 3) Stir in crushed tomatos \\
\midrule

 & 1) A dog runs down a hill and stop behind a shrub. Dog sniffs and chews at patch of grass on rock. the \\
 DiDeMo & dog approaches, then begins to sniff the cluster of \\
 ($\sim$147 symbols  & plants first time hand is seen petting dog. \\
  
   in a text) & 2) Only big screen is visible the camera first pans \\
   & to the large screen. The view shifts from the basketball court to the fans in the seats across the stadium. Camera goes to the bigscreens the dancers are shown on the jumbotraun. \\
  
  & 3) A bus stops. The bus stops at the end of the driveway. A kid is coming out of a school bus. School bus doors open. \\

 \midrule

 MSVD & 1) The cats are fighting \\
 ($\sim$31 symbols & 2) The lady sliced a vegetable \\ 
 in a text) & 3) A man is eating a pizza \\ 
\midrule

  & 1) SOMEONE goes to the kitchen, wets a towel, \\ 
 LSMDC &  comes back to the bed, kneels it, places the towel  \\
 ($\sim$46 symbols & on SOMEONE's brow. \\
  in a text) & 2) He slaps SOMEONE again. \\
  & 3) SOMEONE moves off through the crowd. \\

         	\arrayrulecolor{black}\bottomrule
    \end{tabular}
    }
    \caption{Three random examples of text descriptions in different datasets. With the dataset name, we also report the median length of a text in the dataset. 
    \label{tab:text_styles_short}
    }
    \vspace{-0.25cm}
\end{table} 
\begin{table*}
    \setlength{\tabcolsep}{3pt}
    \resizebox{1\linewidth}{!}{
    
    \begin{tabular}{@{}lll|cccc|cccc|cccc|cccc|cccc@{}}
    	\toprule
                \multirow{2}{*}{Method} & \multirow{2}{*}{Image-Text Datasets} & \multirow{2}{*}{Video-Text Datasets} & \multicolumn{4}{c}{MSR-VTT} & \multicolumn{4}{c}{YouCook2} & \multicolumn{4}{c}{DiDeMo} & \multicolumn{4}{c}{MSVD} & \multicolumn{4}{c}{LSMDC} \\
                & & &  R1 & R5 & R10 & MR & R1 & R5 & R10 & MR & R1 & R5 & R10 & MR & R1 & R5 & R10 & MR & R1 & R5 & R10 & MR \\
                \midrule 
                HowTo100M~\cite{miech2019howto100m} & - & HowTo100M & 7.5& 21.2& 29.6 &38 & 6.1&17.3&24.8&46& - & - & - & - & - & - & - & - & - & - & - & -\\
                SupportSet~\cite{patrick2020support} & - & HowTo100M & 8.7 & 23.0 & 31.1 & 31 & - & - & - & - & - & - & - & - & 8.9 & 26.0 & 37.9 & 18 & - & - & - & - \\
                VATT~\cite{akbari2021vatt} &  & HowTo100M+AS & - & - & 29.7 & 49 & - & - & 45.5 & 13 & - & - & - & - & - & - & - & - & - & - & - & - \\
                EAO$^\S$~\cite{shvetsova2022everything} & - & HowTo100M & 9.9 & 24.0 & 32.6 & 28 & \textbf{19.8} & \textbf{42.9} & \textbf{55.1} & \textbf{8} & 6.6 & 19.0 & 26.8 & 42 & 18.0 & 40.4 & 52.3 & 9 & 3.6 & 8.5 & 13.0 & 177 \\
                Nagrani et al.~\cite{nagrani2022learning} & - & VideoCC3M & 19.4 & 39.5 & 50.3 & - & - & - & - & - & - & - & - & - & - & - & - & - & - & - & - & -\\
                Frozen in Time~\cite{bain2021frozen} & CC+COCO & WebVid-2M & 24.7 & 46.9 & 57.2 & 7 & - & - & - & - & 21.1 & 46.0 & 56.2 & 7 & - & - & - & - & - & - & - & -\\
                CLIP-straight~\cite{portillo2021straightforward} & WIT & - & 31.2 & 53.7 & 64.2 & 4 & - & - & - & - & - & - & - & - & 37.0 & 64.1 & 73.8 & \textbf{2} & 11.3 & 22.7 & 29.2 & 56.5 \\
                CLIP4CLIP~\cite{luo2022clip4clip}  & WIT & HowTo100M & 32.0 & 57.0 & 66.9 & 4 & - & - & - & - & - & - & - & - & 38.5 & 66.9 & 76.8 & \textbf{2} & 15.1 & 28.5 & 36.4 & 28\\
                Nagrani et al.~\cite{nagrani2022learning} & WIT & VideoCC3M & 33.7 & 57.9 & 67.9 & - & - & - & - & - & - & - & - & - & - & - & - & - & - & - & - & -\\
                BLIP$^{||}$~\cite{li2022blip} &  CC+COCO+3more$^*$ & - & 33.3 & 57.3 & 67.5 & 3.5 & 5.8 & 15.0 & 21.9 & 76 & 24.6 & 50.4 & 59.7 & 5.3 & 37.0 & 63.3 & 72.6 & 3 & 15.2 & 28.2 & 35.9 & 35\\
                \textbf{In-Style (ours)} (CLIP) & WIT & HowTo100M$^\dagger$+VATEX$^\ddagger$  & {35.0} & {59.6} & {70.4} & \textbf{3} & {5.1} & {14.0} & {20.3} & {103} & {26.6} & {50.5} & {62.6} & {5} & {38.6} & 66.3 & {77.9} & 3 & 16.0 & \textbf{31.6} & 38.5 & \textbf{26.5} \\       
                \textbf{In-Style (ours)} (BLIP) & CC+COCO+3more$^*$ & HowTo100M$^\dagger$+VATEX$^\ddagger$  & \textbf{36.0} & \textbf{61.9} & \textbf{71.5} & \textbf{3} & 6.8 & 16.7 & 24.5 & 63 & \textbf{29.4} & \textbf{59.2} & \textbf{68.6} & \textbf{3.5} & \textbf{44.9} & \textbf{72.7} & \textbf{81.1} & \textbf{2} & 16.4 & 30.1 & 38.7 & 28\\
                \textbf{In-Style (ours)} (BLIP) & CC+COCO+3more$^*$ & HowTo100M$^\dagger$+WikiHow  & 34.2 & 59.6 & 69.0 & \textbf{3} & 7.3 & 19.2 & 27.1 & 46 & 29.7 & 56.2 & 67.4 & 4 & 42.8 & 70.2 & 79.1 & \textbf{2} & \textbf{17.0} & 30.8 & \textbf{39.6} & 27 \\
                \textbf{In-Style (ours)} (BLIP) & CC+COCO+3more$^*$ & HowTo100M$^\dagger$+Food.com  & 32.8 & 54.9 & 65.8 & 4 & 7.2 & 19.8 & 27.9 & 47 & 25.7 & 52.8 & 63.1 & 5 & 39.5 & 64.9 & 74.9 & \textbf{2}  & 14.5 & 28.9 & 37.2 & 30.5 \\
                
                \midrule
                \textcolor{gray}{\textbf{In-Style (ours)} (BLIP)} & \textcolor{gray}{CC+COCO+3more$^*$} & \textcolor{gray}{HowTo100M$^\dagger$+Target$^\ddagger$ } & \textcolor{gray}{36.2} & \textcolor{gray}{61.8} & \textcolor{gray}{71.9} & \textcolor{gray}{3} & \textcolor{gray}{8.6} & \textcolor{gray}{21.6} & \textcolor{gray}{30.0} & \textcolor{gray}{37} & \textcolor{gray}{32.1} & \textcolor{gray}{61.9} & \textcolor{gray}{71.2} & \textcolor{gray}{3} & \textcolor{gray}{44.8} & \textcolor{gray}{72.5} & \textcolor{gray}{81.2} & \textcolor{gray}{2} & \textcolor{gray}{16.1} & \textcolor{gray}{33.6} & \textcolor{gray}{39.7} & \textcolor{gray}{25} \\
                
                \textcolor{gray}{\textbf{In-Style (ours)} (EAO)} & - & \textcolor{gray}{HowTo100M+Target$^\ddagger$ }  & \textcolor{gray}{16.4} & \textcolor{gray}{35.8} & \textcolor{gray}{48.9} & \textcolor{gray}{10} & \textcolor{gray}{20.3} & \textcolor{gray}{46.4} & \textcolor{gray}{58.8} & \textcolor{gray}{7} & \textcolor{gray}{13.2} & \textcolor{gray}{31.6} & \textcolor{gray}{44.0} & \textcolor{gray}{15} & \textcolor{gray}{23.4} & \textcolor{gray}{50.0} & \textcolor{gray}{62.4} & \textcolor{gray}{5} & \textcolor{gray}{4.9} & \textcolor{gray}{12.3} & \textcolor{gray}{16.7} & \textcolor{gray}{94}  \\
                \bottomrule
    \end{tabular}
    }
    \vspace{-0.2cm}
    \caption{\small{
    \textbf{Zero-shot comparison with other methods.}
    \textbf{Top}: zero-shot retrieval with methods pre-trained on video-language or/and images-language web or/and curated datasets which exclude target datasets during training. For our In-Style method, the VATEX dataset is used as a source of text queries.  \textbf{Bottom}: uncurated \& unpaired text-video retrieval with text queries from the respective target datasets for comparison purposes. Note that this setting is not zero-shot. $\dagger$~denotes that only videos were used (without paired text) and $\ddagger$~--~only text (without videos). $^{\S}$For EAO, performance with S3D backbone is reported.  $^{||}$For BLIP, the performance of dual encoder architecture is reported (not image-grounded text encoder).  $^*$CC~\cite{changpinyo2021conceptual}+COCO~\cite{lin2014microsoft}+VG~\cite{krishna2017visual}+SBU~\cite{ordonez2011im2text} +LAION~\cite{schuhmann2021laion}}. AS denotes AudioSet~\cite{gemmeke2017audio}.
    \label{tab:sota}
    }
    \vspace{-0.1cm}
\end{table*} 
\begin{table*}
    \setlength{\tabcolsep}{3pt}
    \resizebox{1\linewidth}{!}{
    
    \begin{tabular}{@{}l|cccc|cccc|cccc|cccc|cccc|cccc@{}}
    	\toprule
                 Training Data & \multicolumn{4}{c}{MSR-VTT} & \multicolumn{4}{c}{YouCook2} & \multicolumn{4}{c}{DiDeMo} & \multicolumn{4}{c}{MSVD} & \multicolumn{4}{c}{LSMDC} & \multicolumn{4}{c}{Average} \\
                & R1 & R5 & R10 & MR & R1 & R5 & R10 & MR & R1 & R5 & R10 & MR & R1 & R5 & R10 & MR & R1 & R5 & R10 & MR & R1 & R5 & R10 & MR \\
                \midrule 
                 --- (zero-shot) & 34.1 & 60.2 & 70.6 & 3 & 6.0 & 16.2 & 23.1 & 70 & 28.2 & 52.0 & 62.7 & 5 & 38.8 & 64.8 & 74.0 & 2 & 14.5 & 29.3 & 36.4 & 32.5 & 24.3 & 44.5 & 53.3 & 22.5  \\ 
                 Pseudo pairs $P_{ps}$  & 35.0 & 61.4 & 70.9 & 3 & 7.5 & 19.6 & 28.9 & 43 & \textbf{33.1} & 59.8 & 71.2 & \textbf{3} & 44.3 & 72.4 & 81.0 & 2 & 16.8 & 32.7 & \textbf{40.4} & \textbf{25} & 27.3 & 49.2 & 58.4 & 15.2 \\ 
                 Generated pairs $P_{gen}$  & \textbf{36.2} & \textbf{61.8} & \textbf{71.9} & \textbf{3} & 8.6 & 21.6 & \textbf{30.0} & \textbf{37} & 32.1 & \textbf{61.9} & \textbf{71.2} & \textbf{3} & \textbf{44.8} & \textbf{72.5} & \textbf{81.2} & \textbf{2} & 16.1 & \textbf{33.6} & 39.7 & \textbf{25} & \textbf{27.6} & \textbf{50.3} & \textbf{58.8} & \textbf{14.0} \\
                 
                 Combined $P_{ps}$ + $P_{gen}$ & 36.0 & 61.3 & 71.5 & 3 & \textbf{8.9} & \textbf{21.8} & 29.8 & \textbf{37} & 32.6 & 61.8 & 70.2 & 3 & 44.4 & 72.2 & 80.8 & 2 & \textbf{17.1} & 32.4 & \textbf{40.4} & 26 & 27.8 & 49.9 & 58.5 & 14.2 \\

         	\arrayrulecolor{black}\bottomrule
    \end{tabular}
    }
    \vspace{-0.25cm}
    \caption{\small{
    \textbf{Different types of training pairs for text-video retrieval step. } We evaluate text-video retrieval with pseudo pairs $P_{ps}$ only, with generated pairs $P_{gen}$ only, and the combination of both $P_{ps}$ + $P_{gen}$. 
    }
    \label{tab:style_preservation}
    }
    \vspace{-0.3cm}
\end{table*}

\begin{table*}
    \setlength{\tabcolsep}{3pt}
    \resizebox{1\linewidth}{!}{
    
    \begin{tabular}{@{}l|cccc|cccc|cccc|cccc|cccc|cccc@{}}
    	\toprule
                Training data & \multicolumn{4}{c}{MSR-VTT} & \multicolumn{4}{c}{YouCook2} & \multicolumn{4}{c}{DiDeMo} & \multicolumn{4}{c}{MSVD} & \multicolumn{4}{c}{LSMDC} & \multicolumn{4}{c}{Average} \\
                & R1 & R5 & R10 & MR & R1 & R5 & R10 & MR & R1 & R5 & R10 & MR & R1 & R5 & R10 & MR & R1 & R5 & R10 & MR & R1 & R5 & R10 & MR \\
                \midrule 
                 --- (zero-shot) & 34.1 & 60.2 & 70.6 & \textbf{3} & 6.0 & 16.2 & 23.1 & 70 & 28.2 & 52.0 & 62.7 & 5 & 38.8 & 64.8 & 74.0 & \textbf{2} & 14.5 & 29.3 & 36.4 & 32 & 24.3 & 44.5 & 53.3 & 22.5 \\ 
                 $P_{gen}$ with zero-shot captioner & \textbf{36.3} & 61.6 & 71.8 & \textbf{3} & 7.1 & 18.4 & 25.6 & 65 & 28.7 & 56.3 & 65.0 & 4 & 43.8 & 71.2 & 80.1 &\textbf{2} & 16.0 & 29.2 & 37.7 & 30 & 26.3 & 47.3 & 56.1 & 20.8 \\ 
                 In-Style $P_{gen}$ (non-target)  & 36.0 & 61.9 & 71.5 & \textbf{3} & 6.8 & 16.7 & 24.5 & 63 & 29.4 & 59.2 & 68.6 & 3.5 & 44.9 & \textbf{72.7} & 81.1 & \textbf{2} & \textbf{16.4} & 30.1 & 38.7 & 28 & 26.7 & 48.1 & 56.9 & 19.9\\ 
                 In-Style $P_{gen}$ (target) & 36.2 & \textbf{61.8} & \textbf{71.9} & \textbf{3} & \textbf{8.6} & \textbf{21.6} & \textbf{30.0} & \textbf{37} & \textbf{32.1} & \textbf{61.9} & \textbf{71.2} & \textbf{3} & \textbf{44.8} & 72.5 & \textbf{81.2} & \textbf{2} & 16.1 & \textbf{33.6} & \textbf{39.7} & \textbf{25} & \textbf{27.6} & \textbf{50.3} & \textbf{58.8} & \textbf{14}  \\

         	\arrayrulecolor{black}\bottomrule
    \end{tabular}
    }
    \caption{ \small{\textbf{Source of generated pairs $\pmb{P_{gen}}$ for text-video retrieval.} Comparison between zero-shot BLIP (no adaption of retrieval model), zero-shot BLIP captioner, and adapted BLIP captioner with our In-Style method with either text queries from VATEX (non-target) or text queries from the target datasets. 
    }
    \label{tab:caption_source}
    }
    \vspace{-0.3cm}
\end{table*}

We evaluate the proposed uncurated \& unpaired text-video retrieval approach on five popular benchmark datasets: MSR-VTT~\cite{xu2016msr}, YouCook2~\cite{zhou2018towards}, MSVD~\cite{chen2011collecting}, LSMDC~\cite{rohrbach2015long}, and DiDeMo~\cite{anne2017localizing}. All datasets cover different styles of captions and videos, which include YouTube and Flickr videos on various topics and video clips from movies. As a source of support videos, we use the large-scale web dataset HowTo100M~\cite{miech2019howto100m}. We additionally test our model with text queries from the VATEX dataset~\cite{wang2019vatex} as well as with third-party text queries (not video captions), specifically with the recipe steps from Food.com dataset~\cite{foodcom} and task descriptions from WikiHow dataset~\cite{wikihow}. 

\subsection{Dataset Details}

\noindent\textbf{MSR-VTT}~\cite{xu2016msr} contains in total 10k videos on various topics and 200K captions. More precisely, every 20 captions describe the same video in different words. We use split 9K+1K~\cite{gabeur2020multi} in evaluation, resulting in 180K captions for training and 1K text-video pairs for testing.

\noindent\textbf{YouCook2}~\cite{zhou2018towards} is a dataset of 13.5K cooking instructional video clips, where each clip is annotated with a short cooking recipe step. Following~\cite{miech2019howto100m,shvetsova2022everything}, we use a 10K+3.5K training-testing split, leveraging 10K captions for training.

\noindent\textbf{MSVD}~\cite{chen2011collecting} contains 2K video snippets, where each is associated with approximately 40 sentences. The standard split consists of 1200 videos for training, 100 for validation, and 670 for testing. The training set contains 48K captions.

\noindent\textbf{LSMDC}~\cite{rohrbach2015long} is a collection of 202 movies sliced into  118K movie clips with one description per clip and with about 100K clips used for training, and 7408 and 1000 text-video pairs used for validation and testing, respectively. 

\noindent\textbf{DiDeMo}~\cite{anne2017localizing} is a fine-grained text-video dataset. 10K Flickr videos are paired with multiple detailed sentences (40K sentences in total). During training, we use single sentences (33K captions), whereas for evaluation on the test set, we follow~\cite{bain2021frozen} and concatenate all the descriptions for video into one paragraph, acting as a video-paragraph retrieval task (we do not use ground truth time-stamp annotations). 

\noindent\textbf{VATEX}~\cite{wang2019vatex}  dataset contains 35K video clips with multiple annotated captions for a video, covering 600 different human activities. The training set contains 260K captions.

\noindent\textbf{Food.com}~\cite{foodcom} is a text dataset that contains more than 230K recipe texts with over 2.2M recipe steps crawled from websites. We use recipe steps as text queries in our training.

\noindent\textbf{WikiHow}~\cite{wikihow} is a large-scale text dataset using the online WikiHow knowledge base. The dataset contains more than 230K articles covering a variety of topics/tasks and descriptions of steps to solve these tasks. We use only headline steps as text quires, which gives us 1.7M captions.

\noindent\textbf{HowTo100M}~\cite{miech2019howto100m} is a dataset of instructional videos that cover a large variety of topics. The dataset consists of more than 1M videos that were collected by querying on YouTube 23,000 different ``how to'' tasks. In our setup,  we use 8-second non-overlapping clips from a 100K random subset of videos (no more than 15 clips per video) as a support video dataset, resulting in $\sim$1.4M video clips.

\subsection{Implementation Details} 

\paragraph{Model.} We leverage the pre-trained dual-encoder CLIP (ViT-B/32) model~\cite{radford2021learning} in the matching module and the filtering module. Captioner weights are initialized with BLIP (ViT-B/16)  captioner~\cite{li2022blip}, which is pre-trained on five different image-text datasets, including LAION~\cite{schuhmann2021laion} with 129M images. For retrieval, we consider two architectures: dual encoder image-text initialized with BLIP (ViT-B/16) and dual encoder video-text architecture initialized from EAO~\cite{shvetsova2022everything} pre-trained on HowTo100M with noisy ASR narrations. We follow~\cite{shvetsova2022everything} and use a model with a S3D~\cite{xie2018rethinking} feature extractor and weights that were pre-trained with a video-text-audio triplet, but only utilize the video-text encoder and report all results without audio. 

\paragraph{Training.} For training, we uniformly sample $m=8$ frames per video with a resolution of $224\times224$ augmented with RandAugment~\cite{cubuk2020randaugment}. For captioner training and BLIP-architecture retrieval model training, we use AdamW optimizer~\cite{loshchilov2017decoupled} with a weight decay of 0.05, a batch size of 128, and a learning rate $1.0e-05$ for captioner and $1.0e-06$ for retrieval. For the EAO model following~\cite{shvetsova2022everything}, we use Adam optimizer~\cite{kingma2014adam} without weight decay. 
More training details can be found in the supplement. 

\paragraph{Evaluation.} For testing, we use $m=64$ frames for the fine-grained DiDeMo dataset, and $m=12$ for all others, following~\cite{luo2022clip4clip}. For text-video retrieval, we report standard recall metrics for R1, R5, R10, and the median rank (MR).

\subsection{Text Query Style} 
We consider text style as a set of attributes and properties of the text shared across a text corpus. Such properties might be the usage of stop words, sentence construction, sentiment, text length, etc. To highlight those differences, we show three text examples from the different datasets in Table~\ref{tab:text_styles_short}. The respective word clouds for these datasets with and without stop words can be found in~\cref{fig:word_clouds_wo_stop_words} in the supplement. It shows that the sentence structure and most frequent words change across datasets. For example, the YouCook2 test queries always start with an action verb, while in other datasets, the subject+verb+object structure is mostly used. While in the MSR-VTT dataset, frequent words are third-person nouns like ``man'', ``woman'', ``person'', ``people'', the DiDeMo uses more words about camera position like ``camera'', ``left'', ``right'', ``screen'', ``view'', and the LSMDC mostly describes a subject as ``someone''. While the MSR-VTT and the MSVD datasets might look similar, Table~\ref{tab:text_styles_short} shows that sentences in the MSR-VTT are 1.5 times longer than in the MSVD on average. We consider such properties as style properties of the text. 
 
\subsection{Uncurated \& Unpaired Text-Video Retrieval}

\paragraph{Single Dataset Training.} First, we demonstrate the efficiency of the proposed style transfer method in uncurated \& unpaired text-video retrieval on five different downstream datasets in Table~\ref{tab:main_results}. We present results for the image-text pre-trained BLIP~\cite{li2022blip} model as well as for the video-text pre-trained EAO~\cite{shvetsova2022everything} model. We consider three evaluation scenarios: 1) zero-shot performance; 2) the performance of our style transfer method in the text-video retrieval task with uncurated \& unpaired data where only text queries are available during training; 3) training with the ground truth aligned text-video pairs, which can be considered as an upper bound for our task. It shows that the proposed In-Style method significantly outperforms zero-shot performance even without using any aligned training samples from the target distribution. This supports the hypothesis that the style of the text queries is an important component of text-video retrieval. Moreover, we observe that the gap between training with ground truth aligned pairs and the style transfer can be remarkably small, especially on the MSVD dataset, indicating the benefits with respect to a potential annotation cost reduction in the proposed setup.

\paragraph{Multi-Dataset Training.} 
\label{sec:exp-multi-style-training}

Second, we evaluate the proposed multi-dataset training procedure with the In-Style method in Table~\ref{tab:main_generalization}. Here, a minibatch is compiled from a single text source as shown in Figure~\ref{fig:multi-style}. This is favorable compared to the standard training, where data points in a minibatch are randomly sampled from all data sources together.  It shows that the proposed procedure leads to improved retrieval performance compared to individually trained models and better generalization across all datasets compared to the standard multi-dataset training. We attribute the performance increase compared to standard multi-dataset training to the fact that considering the captions of only the same style in contrastive loss provides a model with a cleaner learning signal with stronger text negative counterparts. As an example, ``add sliced cucumber'' in YouCook2 style would be a stronger negative in comparison to a correct ``add sliced tomato'' query than a ``a person in a video puts sliced cucumber in a salad'' in MSR-VTT style. More discussions of generalization can be found in the supplement.

\subsection{Comparison with SOTA}

We further compare the proposed method with zero-shot retrieval baselines in Table~\ref{tab:sota}. We report the performance of BLIP and CLIP backbones trained with text queries from the VATEX dataset, thus text queries do not follow the distribution of any of the test datasets. The closest counterpart to our model is Nagrani et al. method~\cite{nagrani2022learning}, which utilizes the pre-trained image-text CLIP backbone, which is further trained with the VideoCC3M dataset~\cite{nagrani2022learning} -- a video-text dataset collected by automatic transferring image captions from text-image CC3M dataset~\cite{sharma2018conceptual}. The conceptual difference between \cite{nagrani2022learning} and our method is that \cite{nagrani2022learning} proposes to transfer \textit{image} captions from the image-caption dataset by pairing images to videos, while the proposed In-Style method adapts the model to the \textit{video} captions. While noting that a direct comparison to different state-of-the-art methods is limited due to different pre-training datasets, it can be observed that the proposed In-Style method achieves the best results on four out of five datasets, underperforming only in YouCook2, which might benefit from HowTo100M pre-training. We additionally validate the statement that text queries can be used without any corresponding videos by using texts from WikiHow~\cite{wikihow} and Food.com~\cite{foodcom} datasets that contain descriptions of different actions/steps to solve tasks or cook meals. In Table~\ref{tab:sota}, we show that style transfer from both datasets especially benefits YouCook2 retrieval performance that we attribute to the similarity in text styles (see the supplement). However, style transfer from the WikiHow dataset, which is more diverse and covers a larger variety of topics,  also improves the performance over the baselines on the DiDemo, MSVD, and LSMDC datasets.

\subsection{Efficiency of Style Transfer}

\begin{figure*}[]
\begin{center}
\includegraphics[scale=0.19]{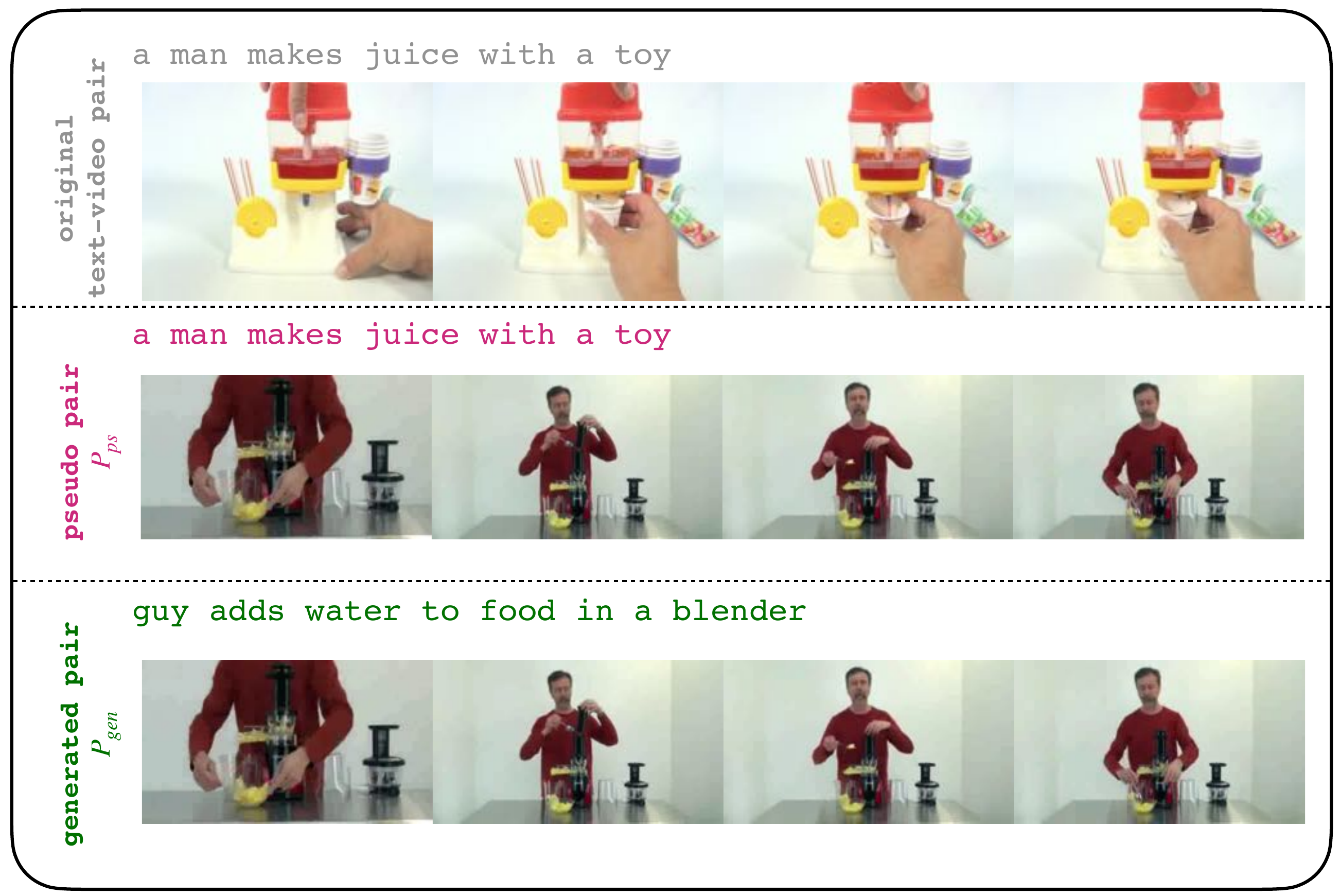}
\includegraphics[scale=0.19]{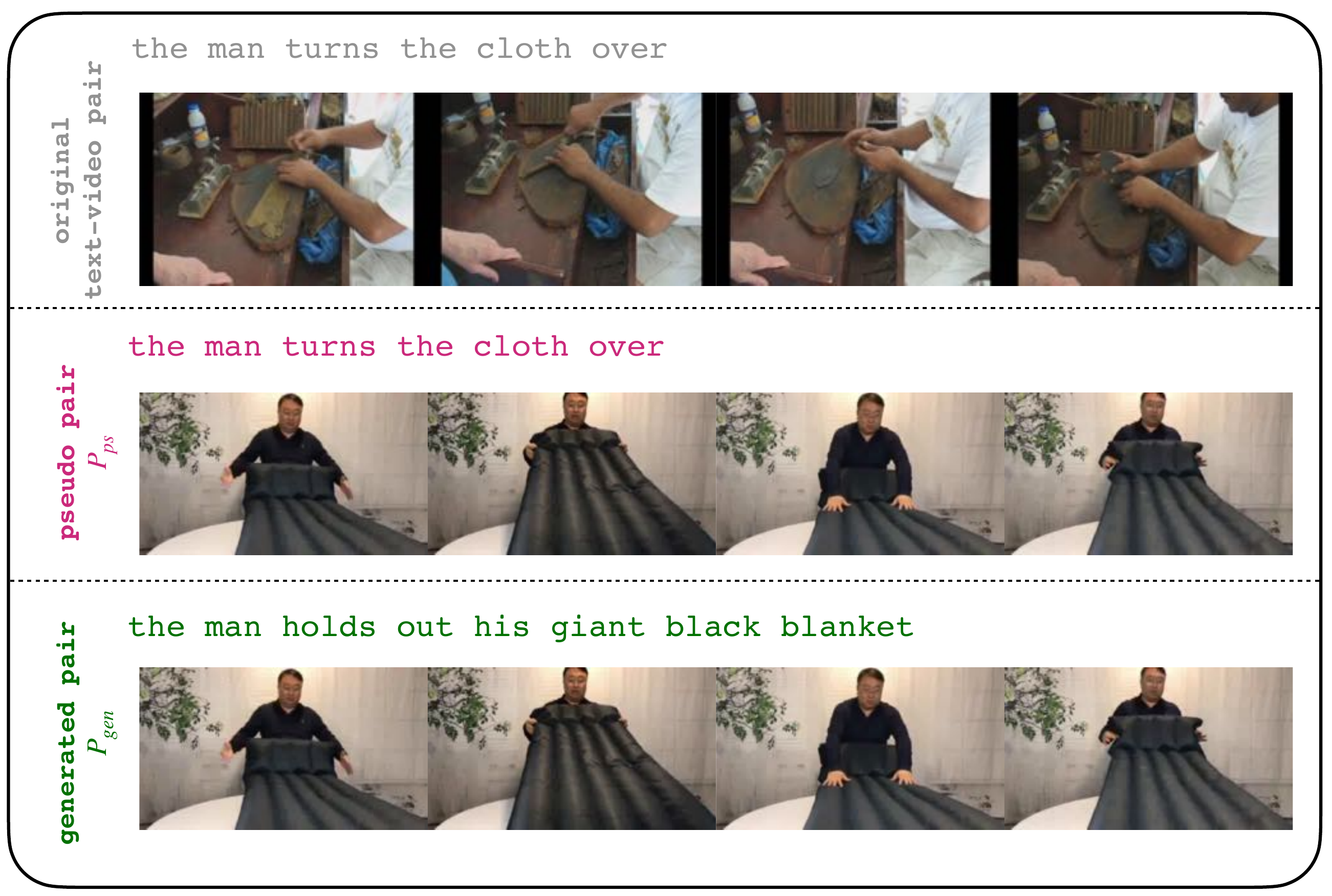}
\end{center}
\vspace{-1.1em}
\caption{ \small{\textbf{Qualitative evaluation of $P_{ps}$ and   $P_{gen}$ on the MSR-VTT (left) and DiDeMo (right) datasets.} First, a text query is matched with one of the videos (a pseudo pair $P_{ps}$), and then, after the style transfer step, for each video, a new caption is generated in the same style but with updated content (a generated pair $P_{gen}$).}
}
\label{fig:qualitative}
\end{figure*}

\paragraph{Training Pairs.} 
In \cref{tab:style_preservation}, we compare the performance of the models trained either with pseudo pairs $P_{ps}$ or with generated pairs $P_{gen}$, or with a combination of them $P_{ps} + P_{gen}$. All setups boost the performance of text-video retrieval by a large margin compared to zero-shot text-video retrieval. The generated pairs $P_{gen}$ achieve better performance than pseudo pairs $P_{ps}$ on all datasets except LSMDC, whereas a combination of $P_{ps} + P_{gen}$ does not improve performance on average. We note that the number of pairs in $P_{gen}$ is significantly larger than in $P_{ps}$ (\cref{tab:number_pairs}) for all datasets except LSMDC (a dataset of movies, which might contain a larger domain shift to YouTube videos compared to other datasets). We assume that in this case, $P_{gen}$ contains better-aligned pairs since each generated text description is conditioned on the corresponding video, while in $P_{ps}$, a fixed set of descriptions is matched (see examples in \cref{fig:qualitative}) explaining the performance drop with $P_{ps} + P_{gen}$.

\paragraph{Style Transfer.}  In Table~\ref{tab:caption_source}, we consider how much the text style transfer in the generated pairs $P_{gen}$ influences the retrieval performance. For this, we considered three sets of $P_{gen}$ for the training retrieval model: 1) $P_{gen}$ generated with zero-shot BLIP captioner; 2) In-Style $P_{gen}$ generated with captioner trained on $P_{ps}$ with text queries from a different non-target dataset (we used the VATEX dataset); 3) In-Style $P_{gen}$ with a captioner trained on $P_{ps}$ with text queries from the target dataset. We observe that training the model with generated text-video pairs (from uncurated web videos from the HowTo100M dataset) by a zero-shot image-pretrained captioner already improves the performance in all video retrieval datasets. We attribute this to the content and style adaptation of the image-language model to the specific appearances in the videos. However, such models tend to generate ``static'' descriptions that do not involve actions. Thus, text queries from non-target video datasets, namely the VATEX dataset, improve the retrieval performance further. Yet, we notice that YouCook2 does not benefit from the VATEX text queries as from the zero-shot generated captions. Finally, using training text queries from the target dataset excels on the considered benchmarks.

\subsection{Ablation Study}
\label{subsec:ablation}
\noindent\textbf{Matching Method.} To obtain generated pairs, we train the captioner with pseudo pairs that were created by a matching module. In Table~\ref{tab:matching}, we consider two options for the matching module: image-text pre-trained dual encoders from BLIP~\cite{li2022blip} and CLIP~\cite{radford2021learning}, as well as the ``Random'' option where text queries are simply matched with the random videos.  We report the text-video retrieval performance of our final model using the given option of the matching module. We observe that the matching module based on CLIP leads to better performance. We attribute that to the robustness of CLIP to the noisy web data as it was trained on large-scale web image-text pairs, whereas BLIP utilizes additional filtering to reduce the noise during training. 

\begin{table}
    \vspace{-0.2cm}
    \setlength{\tabcolsep}{3pt}
    
    \begin{subtable}[t!]{0.48\linewidth}
    \resizebox{0.9\linewidth}{!}{
    \begin{tabular}[t!]{@{}l|cccc@{}}
    	\toprule
                Filt. Thr.  & R1 & R5 & R10 & MR \\ 
                \midrule
                0.26 & 43.9 & 71.8 & 80.8 & \textbf{2}\\
                0.27 & 44.2 & 72.2 & 80.9 & \textbf{2}\\
                0.28 & 44.8 & \textbf{72.5} & \textbf{81.2} & \textbf{2} \\
                0.29 & 45.0 & 72.3 & 80.9 & \textbf{2}  \\
                0.30 &  \textbf{45.1} & 72 & 80.8 & \textbf{2} \\
         	\arrayrulecolor{black}\bottomrule
    \end{tabular}
    }%
    \caption{\small{Filtering threshold \label{tab:filt-threshold}}}
    \end{subtable}%
    \begin{subtable}[t!]{0.48\linewidth}
    \setlength{\tabcolsep}{3pt}
    \resizebox{0.9\linewidth}{!}{
    \begin{tabular}{@{}l|cc@{}}
    	\toprule
                Dataset  &  \#Pseudo & \#Generated  \\ 
                &  Pairs & Pairs \\
                \midrule
                 MSR-VTT & 180k & 495k \\
                 YouCook & 10k & 168k\\
                 Didemo & 33k & 280k \\
                 MSVD & 48k & 379k \\ 
                 LSMDC & 101k & 144k \\
         	\arrayrulecolor{black}\bottomrule 
    \end{tabular}
    }
    \caption{ \small{Number of $P_{ps}$ and $P_{gen}$ }\label{tab:number_pairs}}
    \end{subtable} \\
     
    \begin{subtable}[t!]{0.48\linewidth}
    \setlength{\tabcolsep}{3pt}
    \resizebox{0.9\linewidth}{!}{
    \begin{tabular}[t!]{@{}l|cccc@{}}
    	\toprule
                Matching  & R1 & R5 & R10 & MR \\ 
                \midrule
                Random & 39.1 & 66.3 & 75.8 & \textbf{2} \\
                BLIP & 44.1 & 71.4 & 80.0 & \textbf{2}  \\
                CLIP & \textbf{44.8} & \textbf{72.5} & \textbf{81.2} & \textbf{2} \\
         	\arrayrulecolor{black}\bottomrule
    \end{tabular}
    }
    \caption{\small{Matching method \label{tab:matching}}}
    \end{subtable} 
    \begin{subtable}[t!]{0.48\linewidth}
    \setlength{\tabcolsep}{3pt}
    \resizebox{1.0\linewidth}{!}{
    
    \begin{tabular}[t]{@{}l|cccc@{}}
    	\toprule
                Training pairs  & B@4 & ROUGE & CIDEr \\ 
                \midrule
                 -- (zero-shot) & 0.305 & 0.519 & 0.610 \\
                Pseudo pairs & 0.559 & 0.628 & 1.059 \\
                \arrayrulecolor{black! 50}\bottomrule
                GT pairs & 0.659 & 0.680 & 1.296 \\
         	\arrayrulecolor{black}\bottomrule 
    \end{tabular}
    }
    \caption{\small{Captioning performance \label{tab:captioning}}
    }
    \end{subtable} \\ 
    \vspace{-0.25cm}
    \caption{Ablations of our In-Style method on the MSVD.
    \label{tab:ablations}
    }
    \vspace{-0.5cm}
\end{table}

\noindent\textbf{Filtering Threshold.} In Table~\ref{tab:filt-threshold}, we consider the effect of filtering on the quality of the generated pairs $P_{gen}$. We find threshold $th=0.28$ works the best, indicating that filtering is an important step for our style transfer framework.

\noindent\textbf{Captioning Perforformance} Finally, we evaluate the captioning performance of the captioner trained with pseudo pairs $P_{ps}$ with the standard NLP metrics BLEU@4, ROUGE and CIDEr. Table~\ref{tab:captioning} demonstrates that the captioner trained with pseudo pairs almost doubles the zero-shot captioner performance, significantly reducing the gap to the training with ground truth supervision.

\section{Conclusion}

In this work, we address a new task of \textit{text-video retrieval with uncurated \& unpaired data}, where during training only text queries are available. Motivated by the fact that different domains imply diverse styles of video descriptions, we introduced the In-Style method that preserves the style of the given input queries and transfers it to the support set of unrelated web videos, creating aligned text-video pairs with the style of input. Utilization of obtained text-video pairs as supervision leads to a significant performance boost in text-video retrieval. Moreover, we show the performance generalization of a single model that we train with multiple styles simultaneously, proposing a training procedure for multi-dataset training. %
We evaluate the proposed model over multiple datasets and show the advantages of the In-Style method on the task of uncurated \& unpaired text-video retrieval and achieve new state-of-the-art results for zero-shot text-video retrieval.

\section*{Acknowledgements}
We would like to thank Stephan Alaniz for his invaluable help in this work. 
Nina Shvetsova is supported by German Federal Ministry of Education and Research (BMBF) project STCL - 01IS22067.

{\small
\bibliographystyle{ieee_fullname}
\bibliography{egbib}
}

\clearpage

\addcontentsline{toc}{section}{Appendix} 
\appendix
\noindent{\Large\bf Supplementary Material}\\[1em]
In the supplementary material, we first elaborate on some discussions in Section~\ref{sec:add_discussion}; further, we provide In-Style Method details in Section~\ref{sec:method_details} and implementation details in Section~\ref{sec:imp_details}; and finally, we provide more qualitative evaluations in Section~\ref{sec:qualitative} and discuss limitations in Section~\ref{sec:limitations}.

\section{Additional Discussions}
\label{sec:add_discussion}

\begin{table*}
    \setlength{\tabcolsep}{3pt}
    \centering
    \resizebox{0.97\linewidth}{!}{
    
    \begin{tabular}{@{}c|m{17cm}@{}}
    	\toprule
                Dataset  & Examples\\ 
                \midrule
                  & 1) A bulldozer removes dirt \\
                MSR-VTT & 2) An infomercial with a pharmeceutical company talking about an epilepsy drug pending approval from the FDA \\
                 ($\sim$43 symbols & 3) Extreme violence scenes with people fighting with each other \\
                in a text) & 4) A woman is in front of a whiteboard talking about the numbers written on it \\
                & 5) A man is playing an instrument \\
                \midrule

& 1) Add some herb sprinkle and stir the meat \\
YouCook2 & 2) Bake the pizza on the grill \\
($\sim$39 symbols & 3) Pour butter into the wok \\
in a text) & 4) Peel an onion and chop into pieces \\
 & 5) Add the tomato paste crushed tomatoes tomato puree and beef stock to the pan\\
\midrule

  \multirow{6}{*}{DiDeMo}  & 1) First time we see the dancers go down on one leg the men hit the ground with their sticks. They first start crouching and hitting the ground with the sticks \\
  
\multirow{7}{*}{in a text)} & 2) When the man puts his head down the guitar player is looking up. The guitarist is looking straight up. A man plays the guitar while looking up. The guitarist is looking straight up as he plays. \\
  
($\sim$147 symbols  & 3) Red phone booth is visible a red phone booth is in the scene. A person walks in the middle of the camera. A red phone booth can be seen a red telephone booth is on the sidewalk. \\

   & 4) The camera moves back to the left to the tree. White square exits frame left the camera pans back the way it came. Square area lines with stones comes into view the fence comes into view. \\

& 5) Fog moves in toward the ice skater. a woman spins around several times very fast. A woman pirouettes as she comes near the camera. Woman spins more than 5 times in a row. \\
 \midrule

 & 1) A lady is pouring raw strawberry juice into a bowl \\
MSVD & 2) A man is slicing the crust into a potato \\ 
 ($\sim$31 symbols& 3) A man lifts three sunflowers \\ 
  in a text) & 4) A man is putting a pan into an oven \\
 & 5) A boy rides around in circles on a tricycle \\
\midrule

 & 1) The dish is covered in saffron and spices. \\
LSMDC & 2) He slaps SOMEONE again.  \\
($\sim$46 symbols& 3) He and SOMEONE join forces to grab the cube, which's connected with several more wire. \\
 in a text) & 4) In the race, a rider falls. \\
 & 5) SOMEONE dashes to a clothes closet and ducks inside.
The cup spins across the floor. \\
\midrule

 & 1) Someone is demonstrating how to paint a metal sheet on a window \\
VATEX & 2) A person is cooking scallops in the pan over a fire place and then begins to pour them in some water \\
($\sim$71 symbols& 3) Two women make a video tutorial on how to bake cookies \\
 in a text) & 4) A person is throwing garbage into the trash can and talking \\
 & 5) A woman is outside and preparing an edible meal by inserting herbs into it, then placing them on the ground \\
 \midrule

  & 1) In a blender or food processor, puree the first 3 ingredients until smooth \\
Food.com & 2) Combine juice, remaining 2 tablespoons sugar and lemon juice \\
($\sim$54 symbols& 3) Place chicken on wire rack \\
 in a text) & 4) Then toss in the artichokes and serve immediately \\
 & 5) Cut beef into 3 - inch pieces \\
  \midrule

   & 1) Do lunges in the park \\
WikiHow & 2) Cut a large milk jug \\
($\sim$41 symbols& 3) Buy your favorite knit kit \\
 in a text) & 4) Trim the lining of your sweater \\
 & 5) If your child is in a fight, put his hands on the sand \\

         	\arrayrulecolor{black}\bottomrule
    \end{tabular}
    }
    \caption{Five random examples of text descriptions in different datasets. With the dataset name, we also report the median length of a text in the dataset. 
    \label{tab:text_styles}
    }
    \vspace{-0.25cm}
\end{table*}

\begin{figure*}[t]
\centering
\begin{subfigure}{0.33\linewidth}
\centering
\includegraphics[width=1\linewidth]{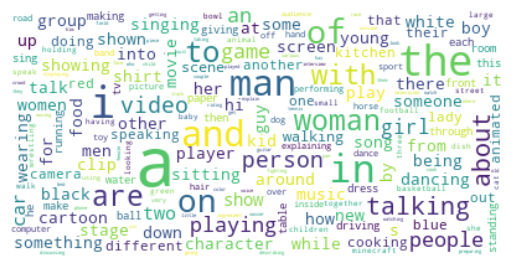}
\caption{\small{MSR-VTT} }
\end{subfigure}
\begin{subfigure}{0.33\linewidth}
\centering
\includegraphics[width=1\linewidth]{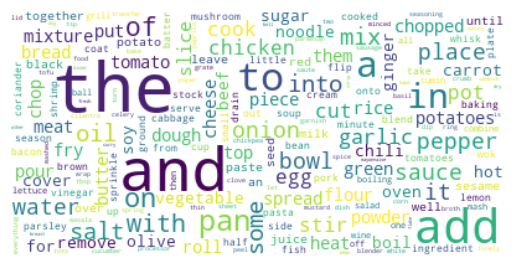}
\caption{\small{YouCook2} }
\end{subfigure}
\begin{subfigure}{0.33\linewidth}
\centering
\includegraphics[width=1\linewidth]{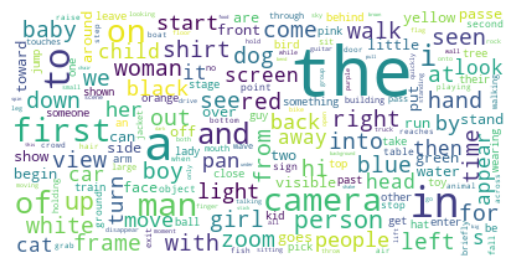}
\caption{\small{DiDeMo} }
\end{subfigure}
\begin{subfigure}{0.33\linewidth}
\centering
\includegraphics[width=1\linewidth]{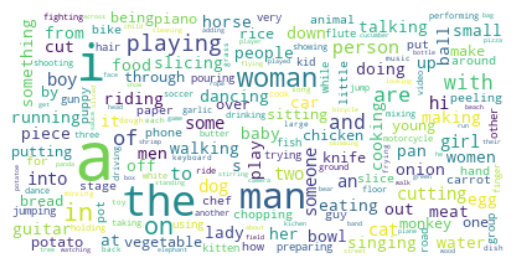}
\caption{\small{MSVD} }
\end{subfigure}
\begin{subfigure}{0.33\linewidth}
\centering
\includegraphics[width=1\linewidth]{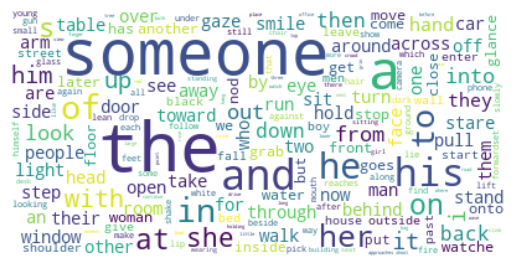}
\caption{\small{LSMDC} }
\end{subfigure}

\caption{ Word clouds for text queries from different datasets (stop words are \textbf{included}). \label{fig:word_clouds}} 
\end{figure*}

\begin{figure*}[t]
\centering
\begin{subfigure}{0.33\linewidth}
\centering
\includegraphics[width=1\linewidth]{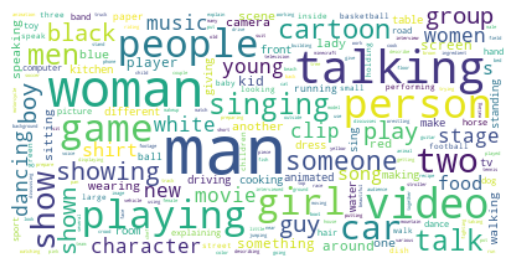}
\caption{\small{MSR-VTT} }
\end{subfigure}
\begin{subfigure}{0.33\linewidth}
\centering
\includegraphics[width=1\linewidth]{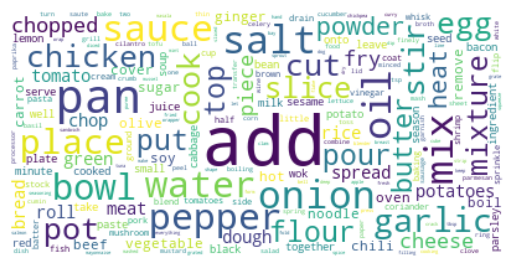}
\caption{\small{YouCook2} }
\end{subfigure}
\begin{subfigure}{0.33\linewidth}
\centering
\includegraphics[width=1\linewidth]{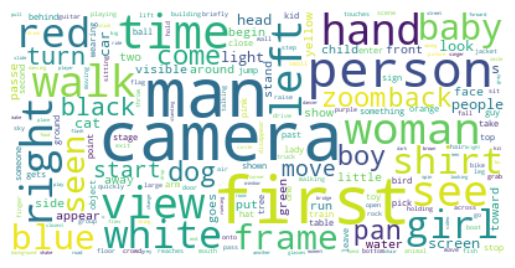}
\caption{\small{DiDeMo} }
\end{subfigure}
\begin{subfigure}{0.33\linewidth}
\centering
\includegraphics[width=1\linewidth]{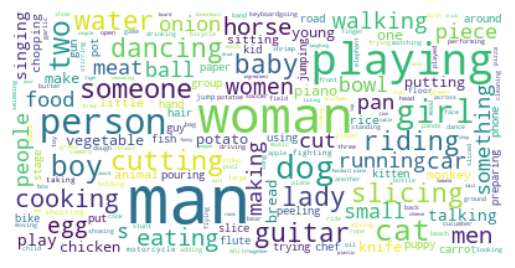}
\caption{\small{MSVD} }
\end{subfigure}
\begin{subfigure}{0.33\linewidth}
\centering
\includegraphics[width=1\linewidth]{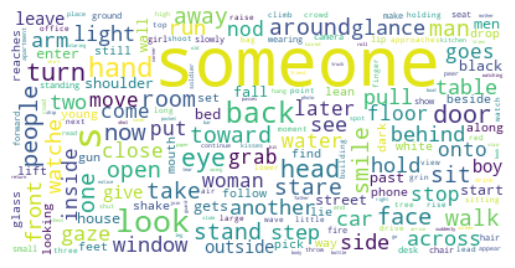}
\caption{\small{LSMDC} }
\end{subfigure}

\caption{ Word clouds for text queries from different datasets (stop words are \textbf{excluded}). \label{fig:word_clouds_wo_stop_words}} 
\end{figure*}

\paragraph{Text Query Style.} Figure~\ref{fig:word_clouds} and Figure~\ref{fig:word_clouds_wo_stop_words} show the respective word clouds for the five datasets with and without stop words. In Table~\ref{tab:text_styles}, we show five different text examples from the different datasets considered in this paper to further highlight the differences in the text styles.

\paragraph{Model Generalization.}

\begin{table*}
    \setlength{\tabcolsep}{3pt}
    \centering

    \begin{subtable}[t!]{0.48\linewidth}
    \centering
    \resizebox{1\linewidth}{!}{
    \begin{tabular}{@{}l|ccccc@{}}
    	\toprule
                Dataset  & MSR-VTT & YouCook2 & DiDeMo & MSVD & LSMDC \\ 
                \midrule
                 MSR-VTT & 495k & 111k & 160k & 239k & 87k \\
                 YouCook & 111k & 168k & 65k & 97k & 40k \\
                 Didemo  & 160k & 65k & 280k & 135k & 57k \\
                 MSVD    & 239k & 97k & 135k & 379k & 75k \\
                 LSMDC   & 87k & 40k & 57k & 75k & 144k \\
         	\arrayrulecolor{black}\bottomrule 
    \end{tabular}
    }
    \caption{Number of shared video clips
    \label{tab:shared_videos_number}
    }
    \end{subtable}
    \hspace{\fill}
    \begin{subtable}[t!]{0.48\linewidth}
    \centering
    \resizebox{1\linewidth}{!}{
    \begin{tabular}{@{}l|ccccc@{}}
    	\toprule
                Dataset  & MSR-VTT & YouCook2 & DiDeMo & MSVD & LSMDC \\ 
                \midrule
                MSR-VTT & 1 & 0.23 & 0.32 & 0.48 & 0.18 \\
                 YouCook2 & 0.66 & 1 & 0.39 & 0.58 & 0.24 \\
                 DiDeMo & 0.57 & 0.23 & 1 & 0.48 & 0.2 \\
                 MSVD & 0.63 & 0.26 & 0.36 & 1 & 0.2 \\
                 LSMDC & 0.61 & 0.28 & 0.4 & 0.52 & 1 \\
         	\arrayrulecolor{black}\bottomrule 
    \end{tabular}
    }%
    \caption{Ratio of shared video clips per dataset
    \label{tab:shared_videos_ratio}
    }
    \end{subtable}%

    \caption{Number/Ratio of shared video clips in the datasets' generated pairs $P_{gen}$.
    \label{tab:shared_videos}
    }
\end{table*}

In the following, we discuss the generalization performance of the model trained on different text styles (Table~\ref{tab:main_generalization} in the main paper.) When we consider models trained only with one text style (text queries that are only from one dataset) in the top half of Table~\ref{tab:main_generalization}, it shows that the mean retrieval performance is higher for the queries with the style of MSR-VTT, DiDeMo, or LSMDC datasets. Interestingly, MSVD-style text queries, which are similar to MSR-VTT queries in terms of sentence structure (Table~\ref{tab:text_styles}) and usage of stop words (Figure~\ref{fig:word_clouds} and Figure~\ref{fig:word_clouds_wo_stop_words}), show similar MSR-VTT retrieval performance compared to  DiDeMo and LSMDC-style text queries, and moreover, lower performance on the DiDeMo, and LSMDC datasets.  We hypothesize that longer and more descriptive text is beneficial for model generalization (the MSVD dataset contains the shortest text descriptions of all datasets). 

In Table~\ref{tab:shared_videos_number}, we also report the number of shared video clips in generated pairs $P_{gen}$ with different datasets' text styles. Interestingly, $P_{gen}$ based on the LSMDC text queries has the smallest number of pairs in general. Moreover, even though it has low overlap in videos with $P_{gen}$ from all other datasets (0.18--0.24, see in Table~\ref{tab:shared_videos_ratio}), LSMDC's $P_{gen}$ shows one of the highest generalization to the MSVD, DiDeMo, and MSR-VTT datasets. Analyzing examples in Table~\ref{tab:text_styles}, we suggest that LSMDC captions are more concise and descriptive in terms of object-verb-subject details (who does what in a video). For example, in the 3rd MSR-VTT example, there is only a general description that people are fighting without specification of exact people and their actions, and in the 5th example, the caption only says that a man is playing an instrument, without specification of an instrument. At the same time, in LSMDC texts, the object-action-subject description is more detailed, such as in the first example, all consecutive actions are specified, or in the 2nd example, the exact hitting action (``he slaps'') is specified. Therefore, we hypothesize that a text style with more concise descriptions is better for model generalization. However, as Table 2 demonstrates, the best mean performance over all datasets is achieved while training with different target text styles. 

\section{In-Style Method Details}
\label{sec:method_details}

\begin{figure*}[t]
\centering
\begin{subfigure}{0.49\linewidth}
\centering
\includegraphics[width=0.92\linewidth]{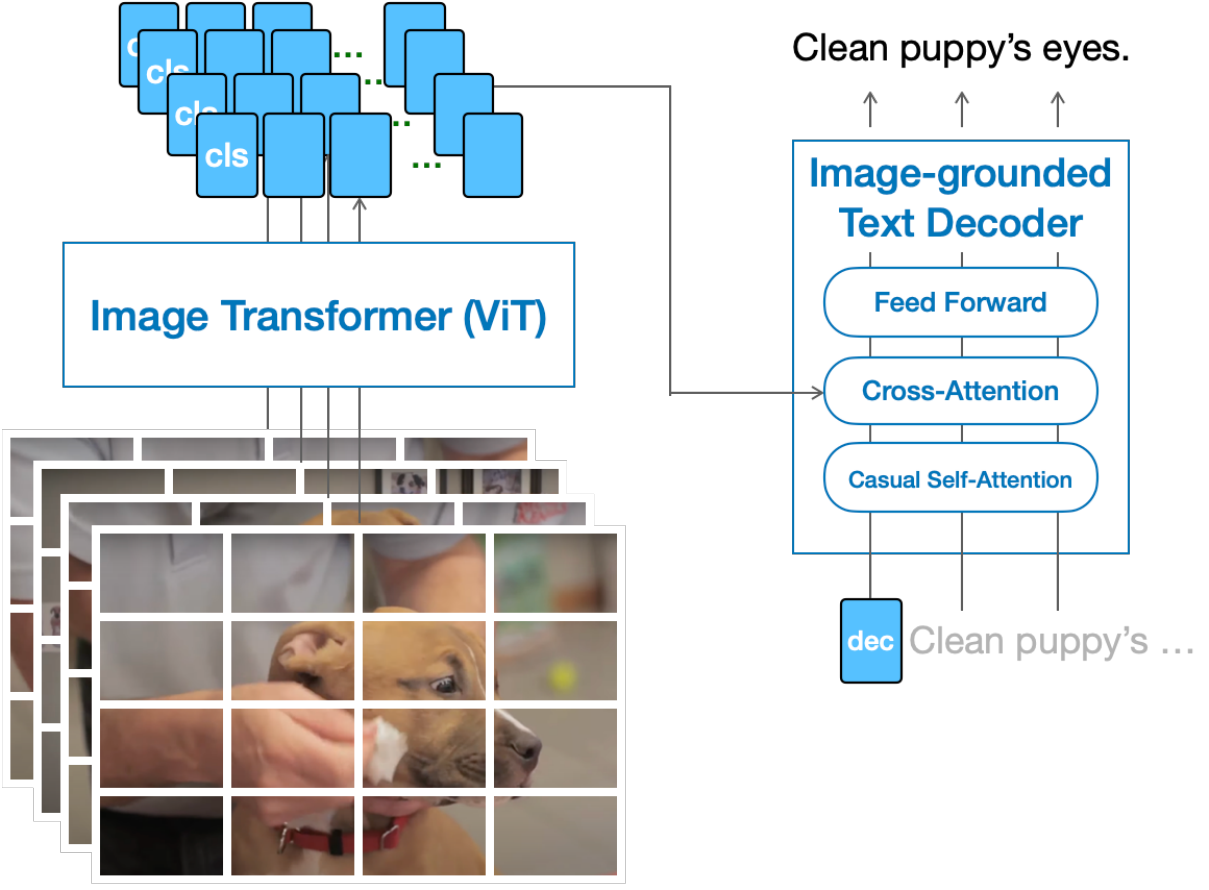}
\caption{\small{Captioner} }
\label{fig:captioner}
\end{subfigure}
\hspace{\fill}
\begin{subfigure}{0.49\linewidth}
\centering
\includegraphics[width=0.92\linewidth]{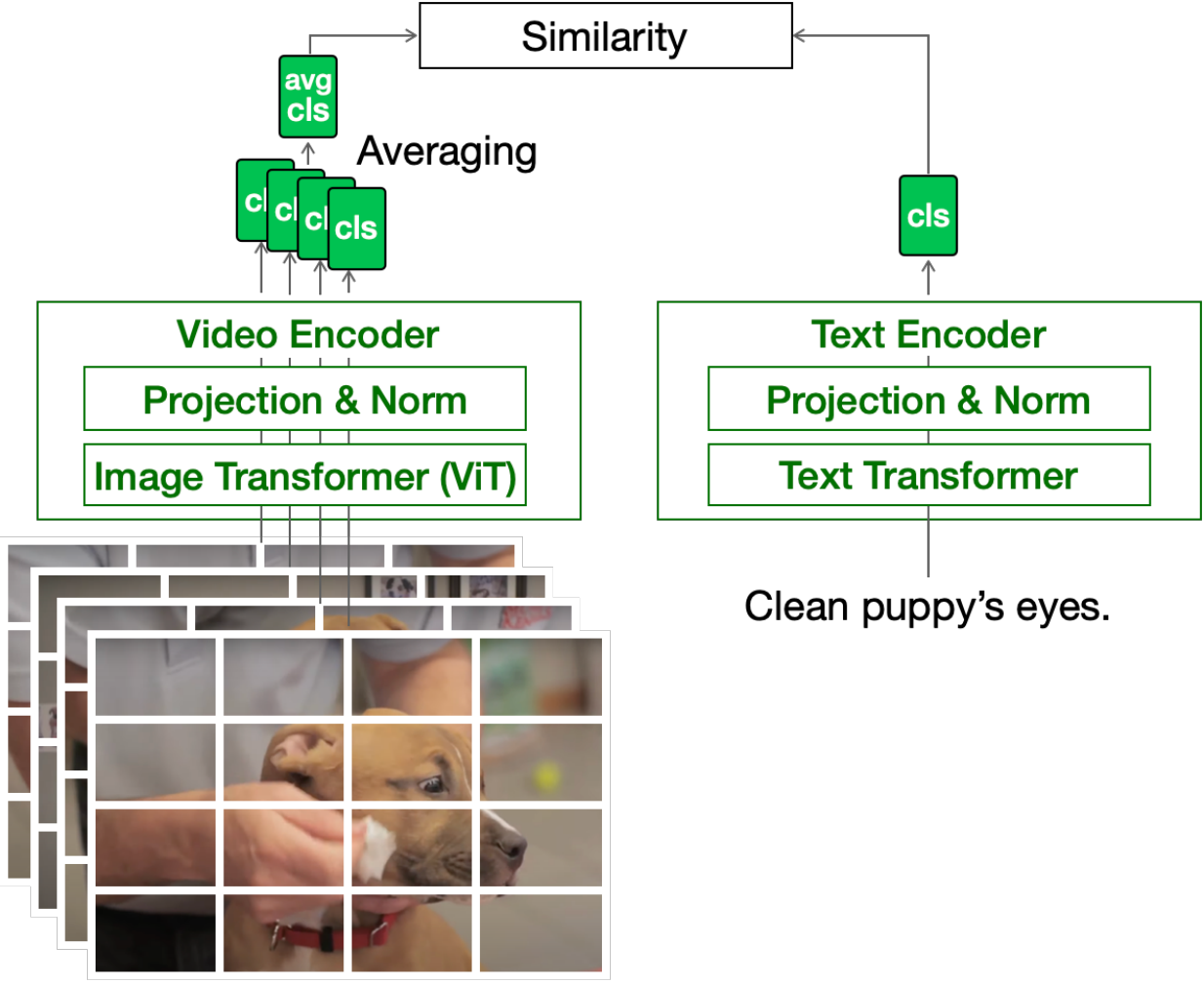}
\caption{ \small{Video-Text Dual Encoder} }
\label{fig:dual_encoder}
\end{subfigure}

\caption{Schematic visualization of (a) the video captioner architecture; and (b) the video-text dual encoder model. }
\end{figure*}

\paragraph{Captioner.} For the captioner, we follow BLIP~\cite{li2022blip} image captioner architecture, which we extend to video captioning as shown in Figure~\ref{fig:captioner}. Namely, we encode  $m$ uniformly sampled frames (we use $m=8$ for training and $m=12$ for inference) from a video by the image transformer to obtain frame-wise tokens. Then, we feed this set of encoded tokens from all the frames into the cross-attention of an image-grounded text decoder. Therefore, the predicted text is conditioned on multiple video frames at once. The image-grounded text decoder predicts the next text token given an input of previous text tokens (where the ``dec'' token is concatenated to the beginning of the input sequence and denotes the start of the output). During the inference, the text tokens are generated one by one in an autoregressive manner. 

\paragraph{Video-Text Dual Encoder.}  In the pseudo matching, filtering,  and retrieval steps of our In-Style method, 
we use the video-text dual encoder model. We initialize the dual encoder model from image-text pre-trained BLIP~\cite{li2022blip} or CLIP~\cite{radford2021learning} models, which we extend to video-text models. Specifically, we obtain a video embedding by averaging image representations of $m$ uniformly sampled frames, as shown in Figure~\ref{fig:dual_encoder}. By default, eight frames are used $m=8$ during training and $m=12$ during evaluation. But $m$ is increased to 64 during evaluation on the fine-graned DiDeMo dataset as in~\cite{luo2022clip4clip, wu2022cap4video}. 
To obtain image representation, we feed image patches into ViT transformer~\cite{dosovitskiyimage}, and the ``cls'' output token is later projected by a linear projection into common embedding space and further normalized. The text representation is obtained by projecting and normalizing an output ``cls'' token in the case of BLIP and an output ``eot'' token in the case of CLIP.

\section{Implementation Details}
\label{sec:imp_details}

\paragraph{EOA Model Details.} In case of the EOA~\cite{shvetsova2022everything} architecture, we use the model variant with S3D visual backbone with a frozen S3D backbone pre-trained by Miech et al. ~\cite{miech2020end} in the visual branch and a GoogleNews pre-trained Word2vec model~\cite{mikolov2013efficient} in the text branch. These backbones are fixed and not trained. As initialization, we use EOA~\cite{shvetsova2022everything} weights pre-trained with audio modality, but we discard the audio branch to consider text-video-only retrieval.

\paragraph{EOA Training Details.}  We follow~\cite{shvetsova2022everything} to train the EOA model. We compute one S3D feature per second~\cite{miech2020end} without any data augmentation; namely, we compute one feature per 16 frames, sampling frames with 16 fps and 224x224 resolution. We train with Adam~\cite{kingma2014adam} optimizer, no weight decay, and a batch size of 128. We use a temperature of 0.05 in the loss function. 

\paragraph{BLIP Evaluation Details.} We would like to highlight that while evaluating a model with a BLIP backbone, we use dual encoder architecture (unimodal encoders), not a cross-attention architecture (named an image-grounded text encoder in~\cite{li2022blip}), which performance on MSR-VTT dataset~\cite{xu2016msr} was reported in the original paper~\cite{li2022blip}. Dual encoder models independently encode videos and texts into common embedding space, allowing for fast retrieval among thousands of videos by pre-computing video embeddings and calculating the similarity between text and video embeddings with a dot product (cosine similarity)~\cite{miech2021thinking}. Cross-attention architectures compute similarity by propagating video and text together in the model with cross-attention layers, attending all words and all spatial-temporal video patches to each other. Cross-attention architecture significantly boosts retrieval performance compared to dual encoder models~\cite{miech2021thinking}; however, it demands enormous computational overhead in the inference phase, requiring propagating all videos paired with a given text query to compute similarity. In the original paper~\cite{li2022blip}, BLIP performance on the MSR-VTT dataset was reported with the cross-attention model used to rerank 128 closest videos found by the dual encoder model. Since the majority of the methods~\cite{radford2021learning,shvetsova2022everything,patrick2020support,miech2019howto100m,luo2022clip4clip,akbari2021vatt} leverage dual encoder architecture for video retrieval due to computational benefits, for comparison purpose we also base our method on dual encoder models.

\section{Qualitative Results}
\label{sec:qualitative}

\begin{figure*}[!t]
\begin{center}
\includegraphics[scale=0.19]{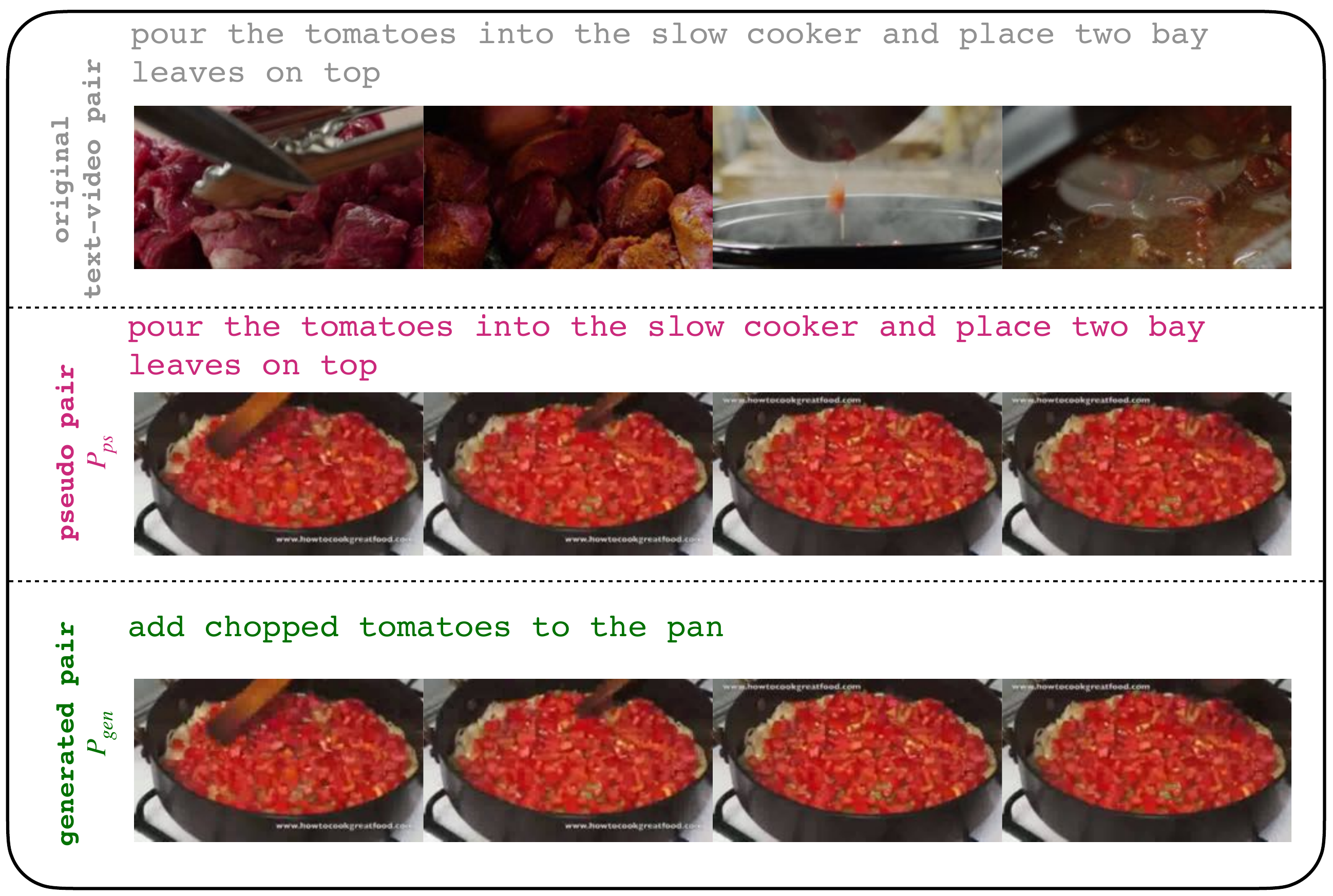}
\includegraphics[scale=0.19]{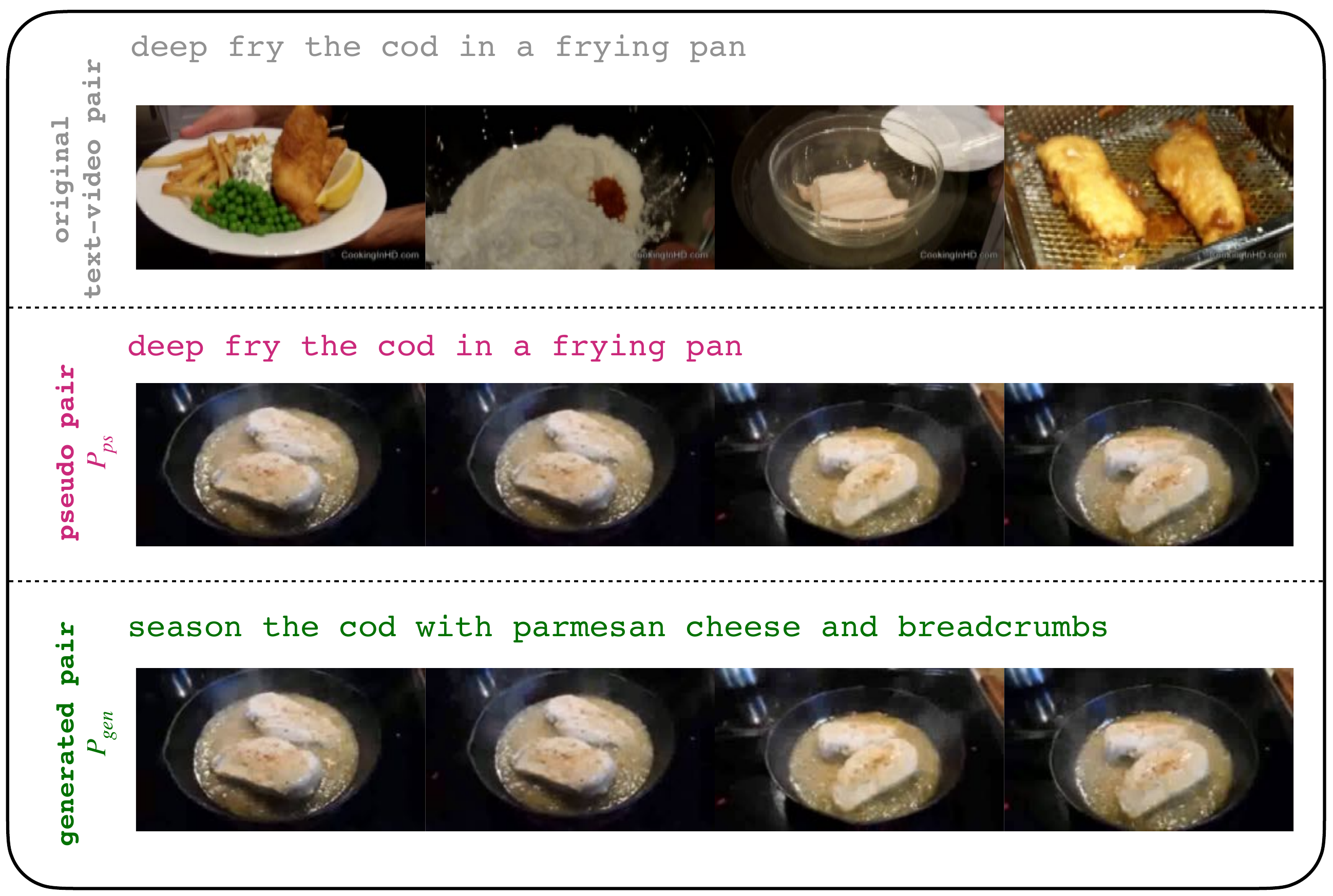}
\includegraphics[scale=0.19]{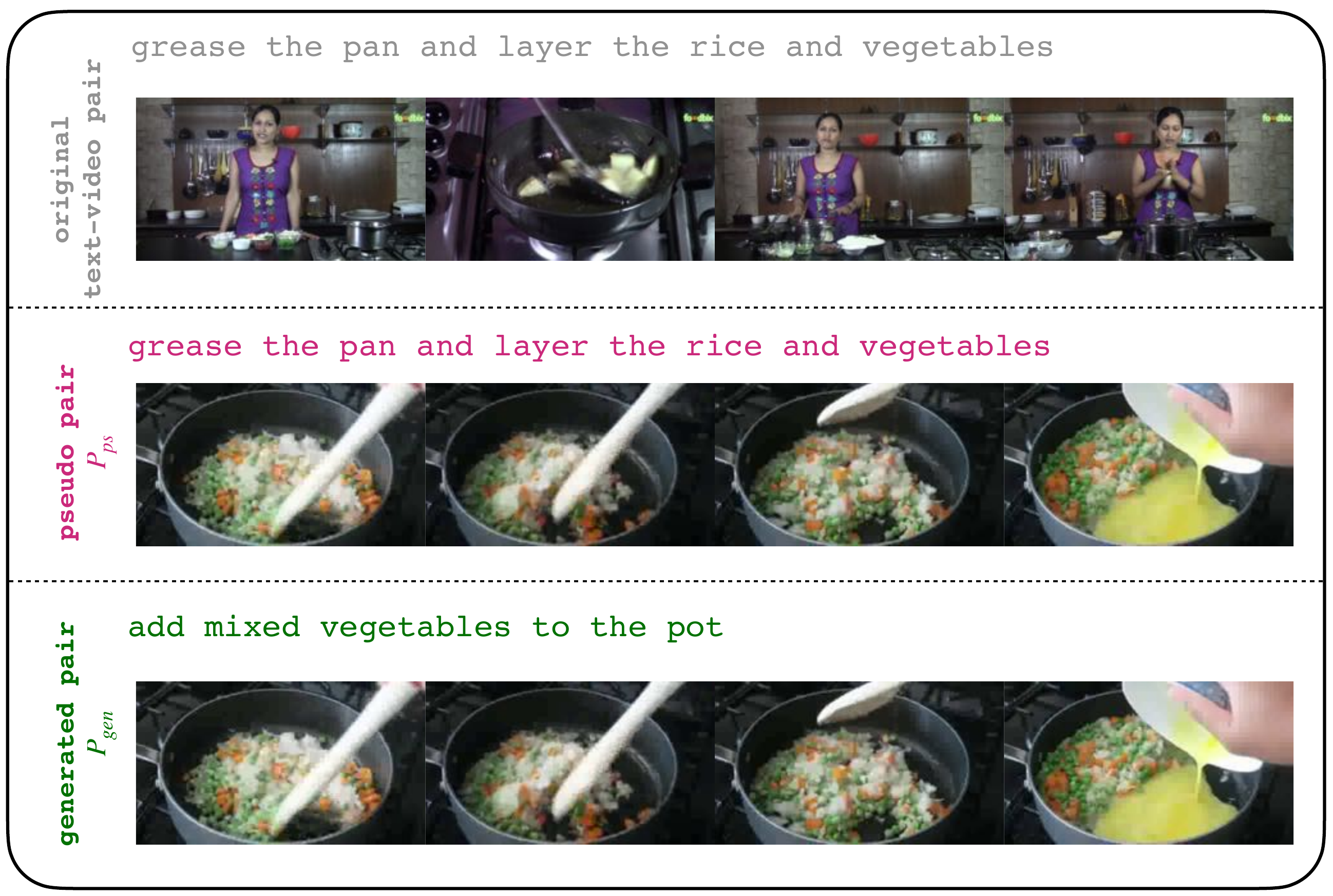}
\includegraphics[scale=0.19]{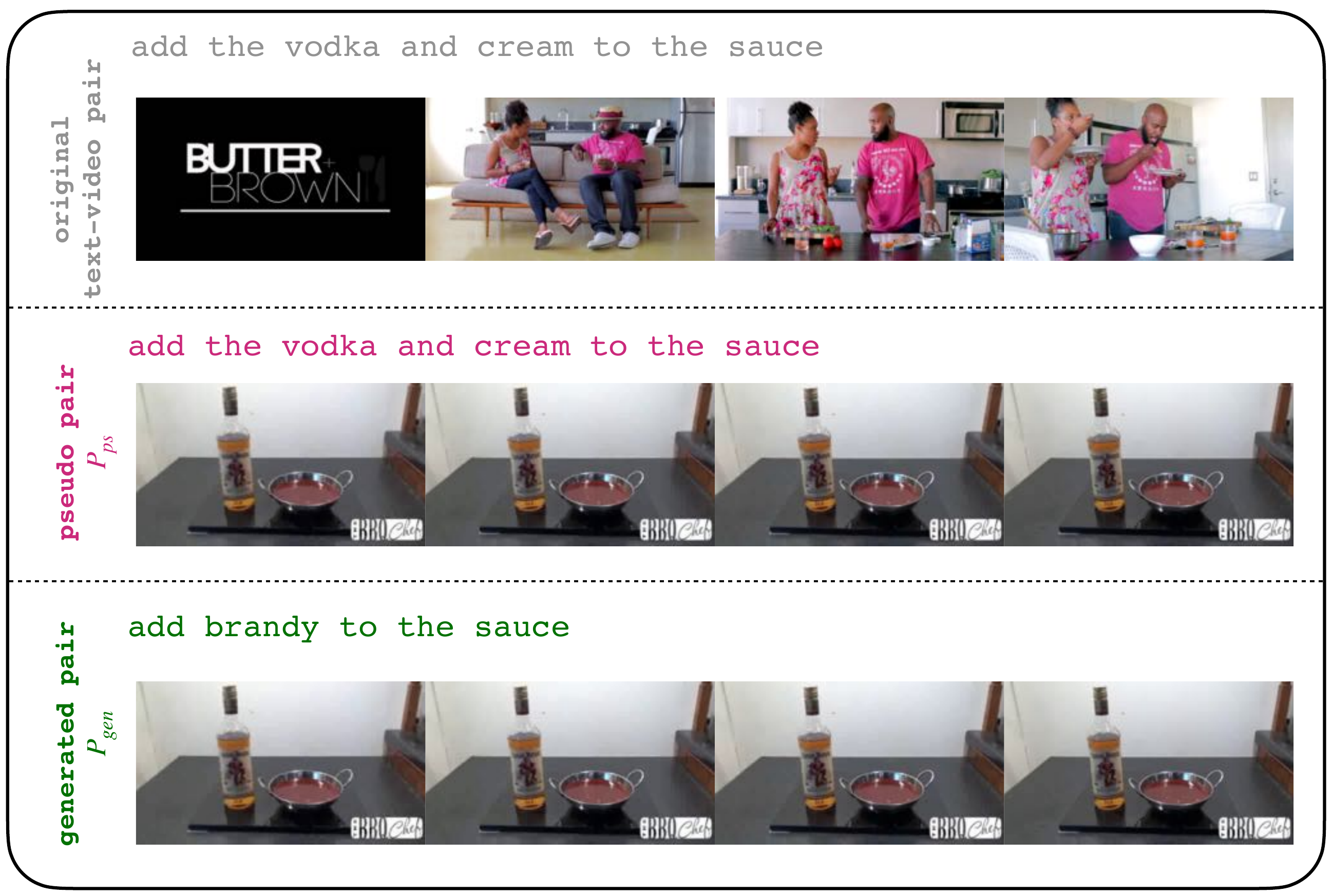}
\includegraphics[scale=0.19]{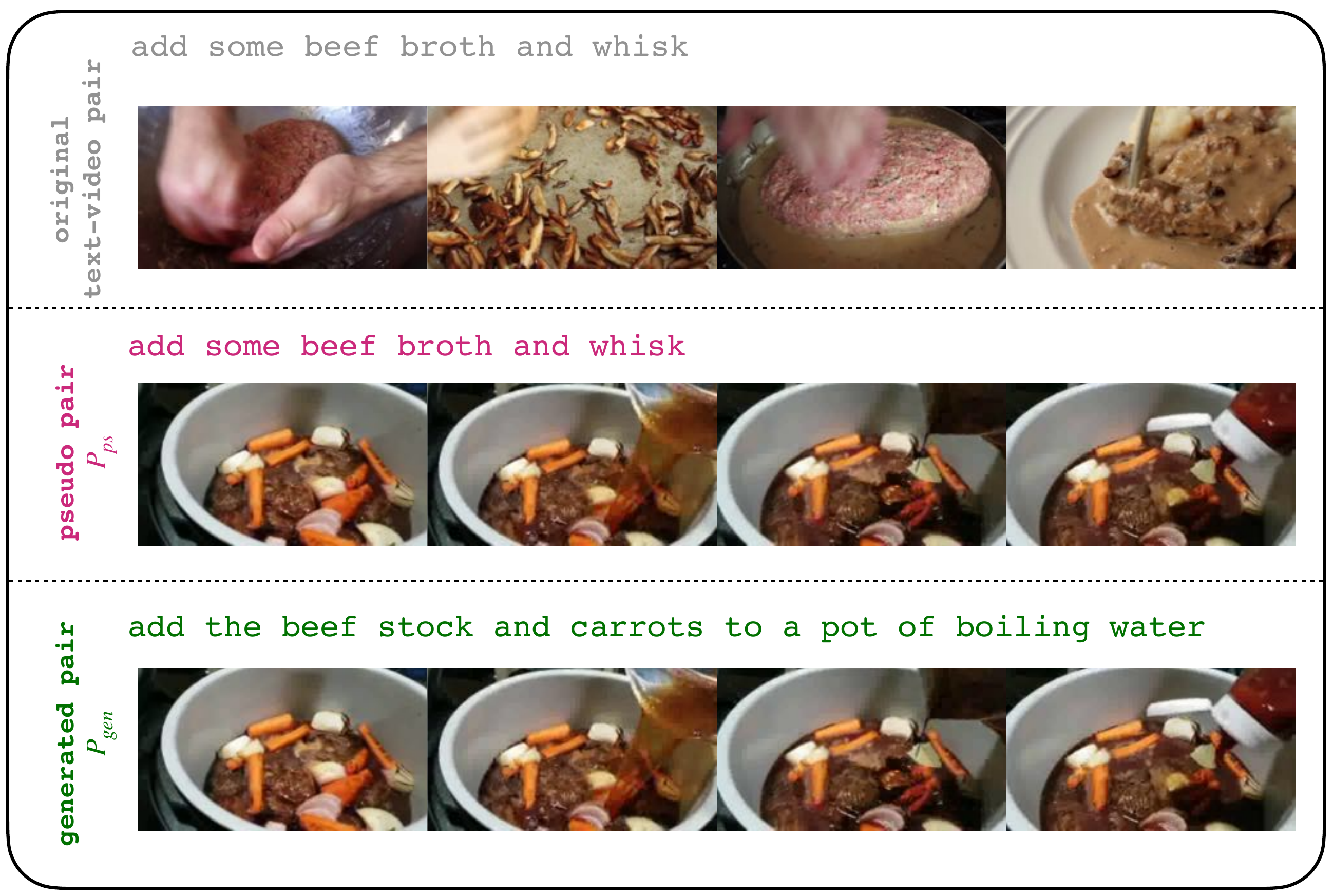}
\includegraphics[scale=0.19]{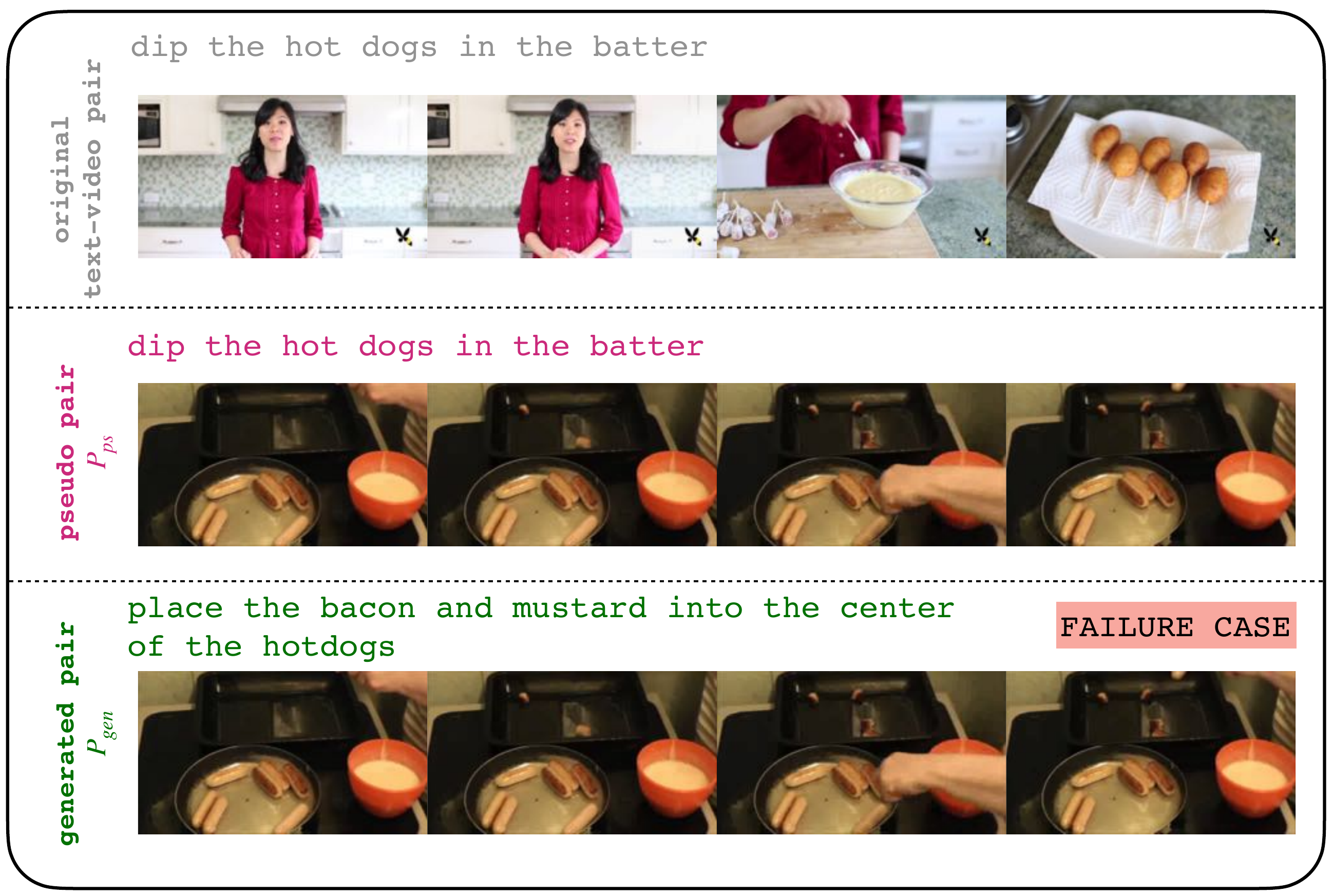}
\end{center}
\vspace{-1.1em}
\caption{ \small{\textbf{Qualitative evaluation of $P_{ps}$ and   $P_{gen}$ on the YouCook2.} First, a text query is matched with one of the videos (a pseudo pair $P_{ps}$), and then, after the style transfer step, for each video, a new caption is generated in the same style but with updated content (a generated pair $P_{gen}$)}
}
\label{fig:youcook_qual}
\end{figure*}

\begin{figure*}[!t]
\begin{center}
\includegraphics[scale=0.19]{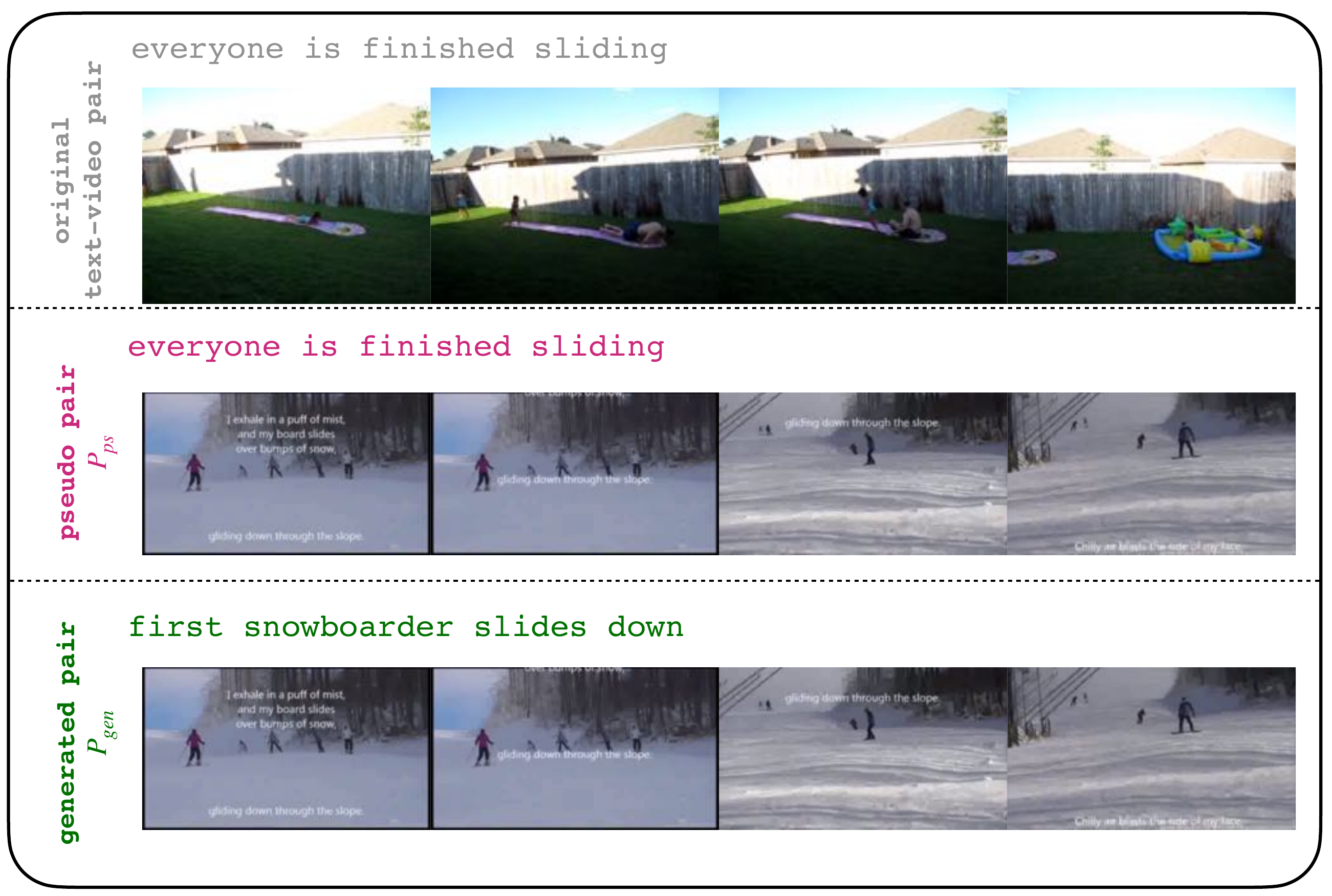}
\includegraphics[scale=0.19]{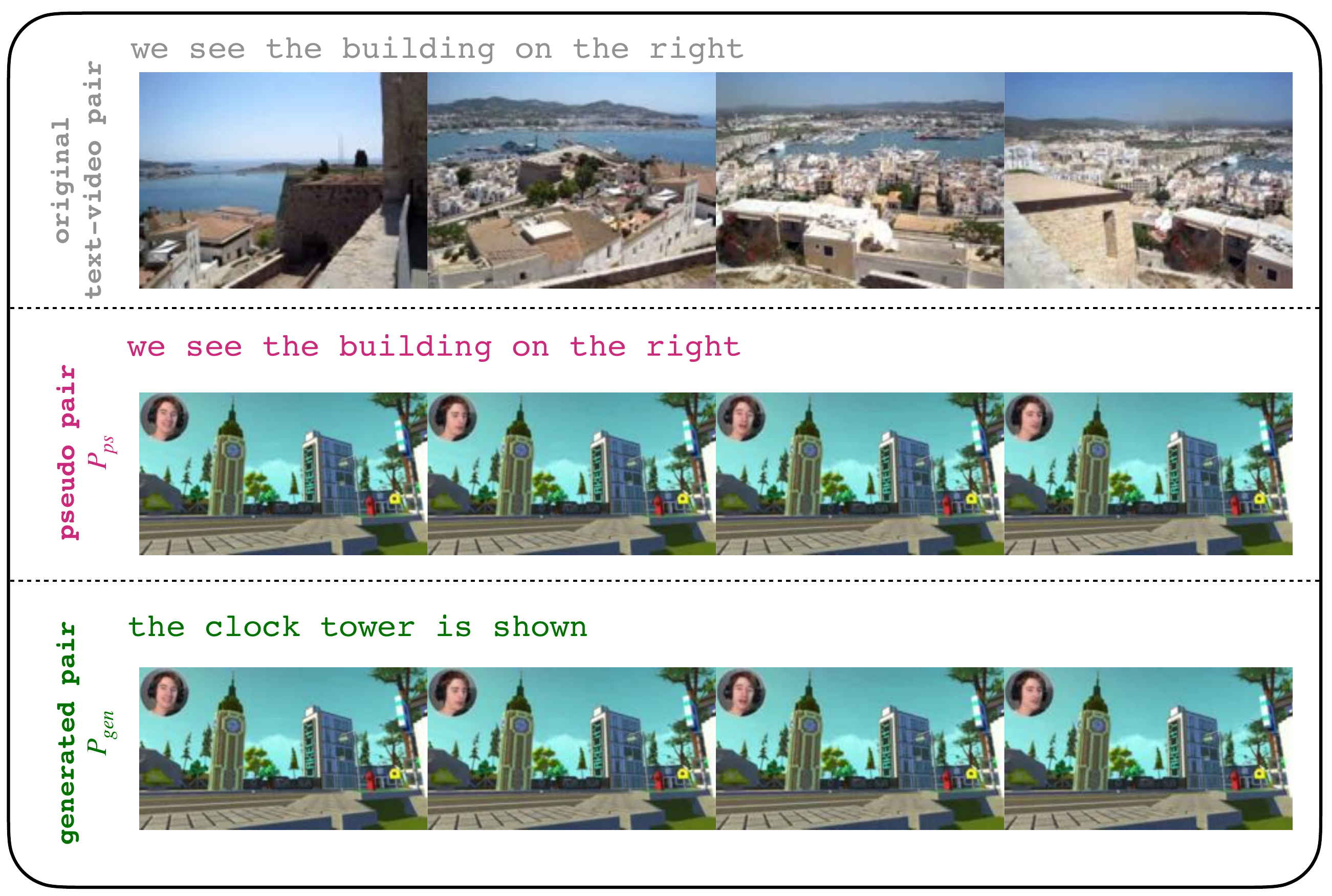}
\includegraphics[scale=0.19]{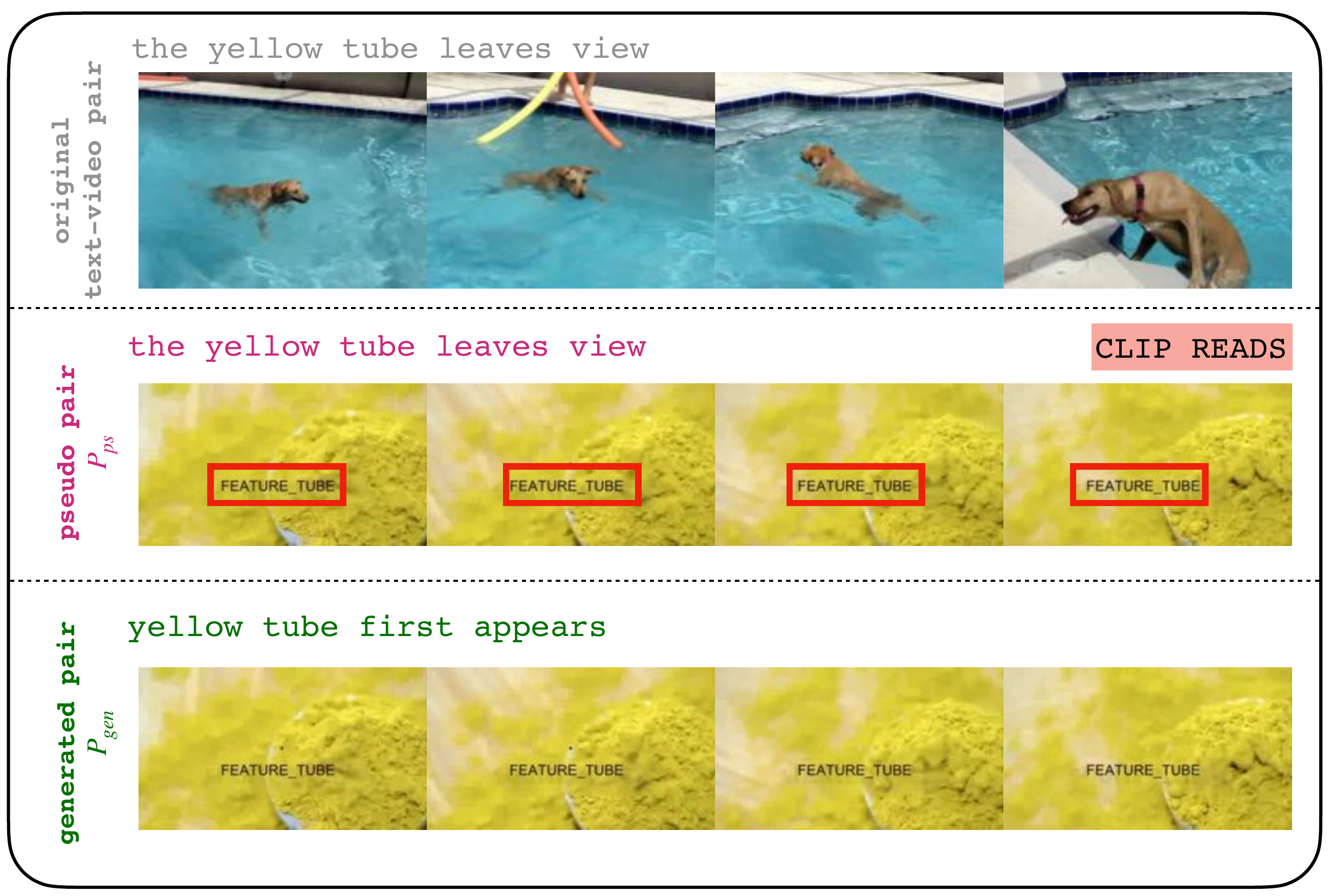}
\includegraphics[scale=0.19]{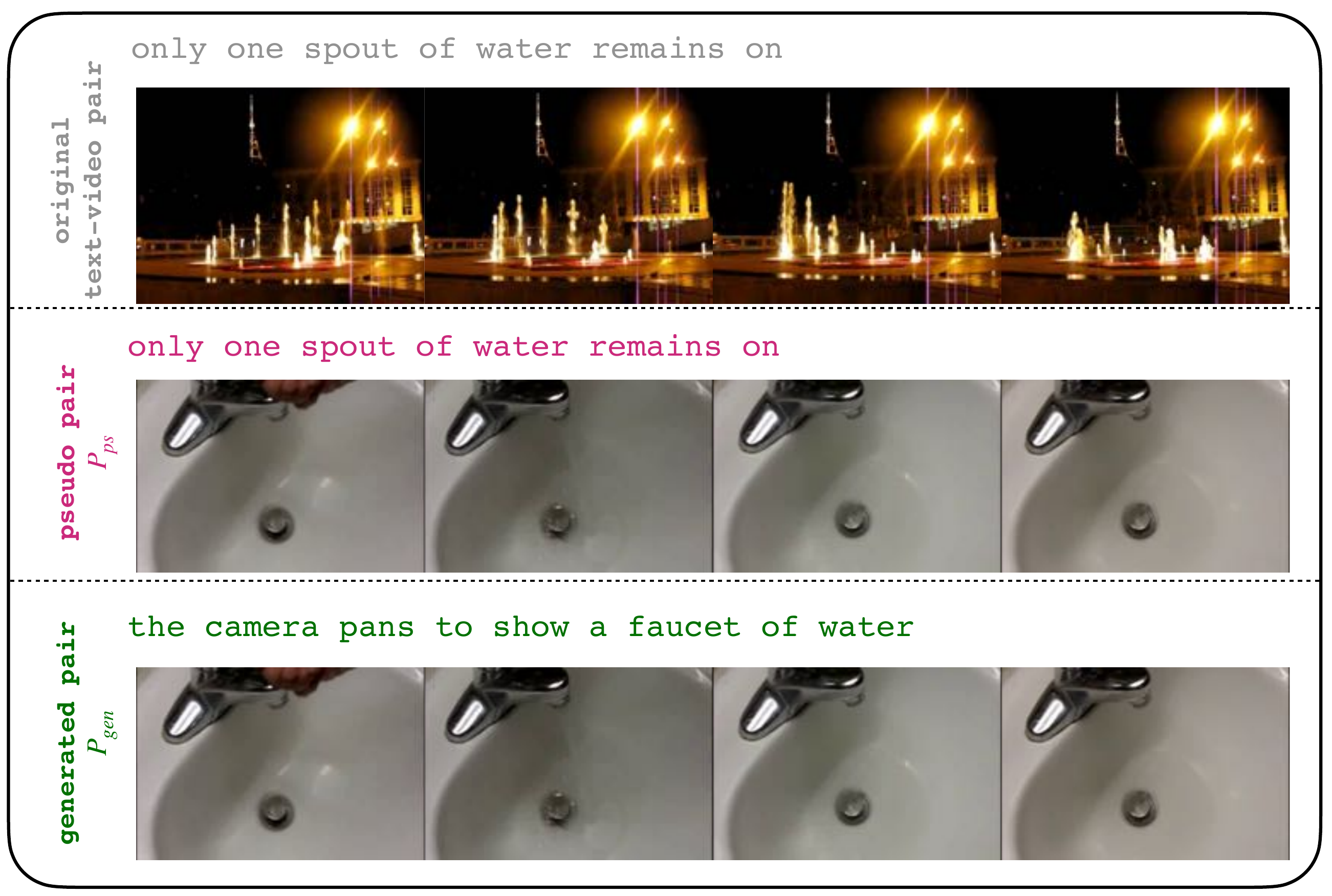}
\includegraphics[scale=0.19]{figs/didemo6.pdf}
\includegraphics[scale=0.19]{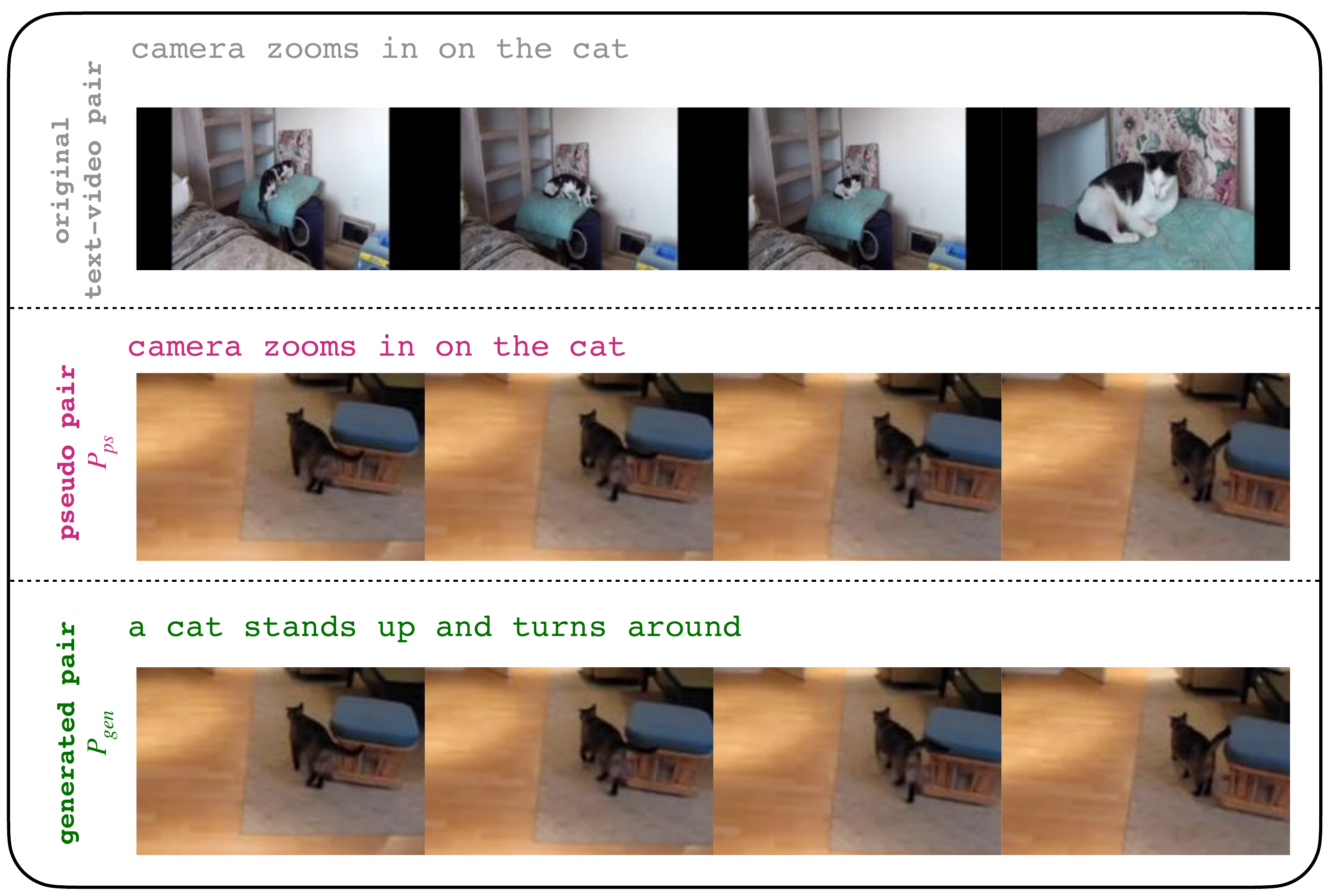}
\end{center}
\vspace{-1.1em}
\caption{ \small{\textbf{Qualitative evaluation of $P_{ps}$ and   $P_{gen}$ on the DiDeMo.} First, a text query is matched with one of the videos (a pseudo pair $P_{ps}$), and then, after the style transfer step, for each video, a new caption is generated in the same style but with updated content (a generated pair $P_{gen}$)}
}
\label{fig:didemo_qual}
\end{figure*}

\begin{figure*}[!t]
\begin{center}
\includegraphics[scale=0.19]{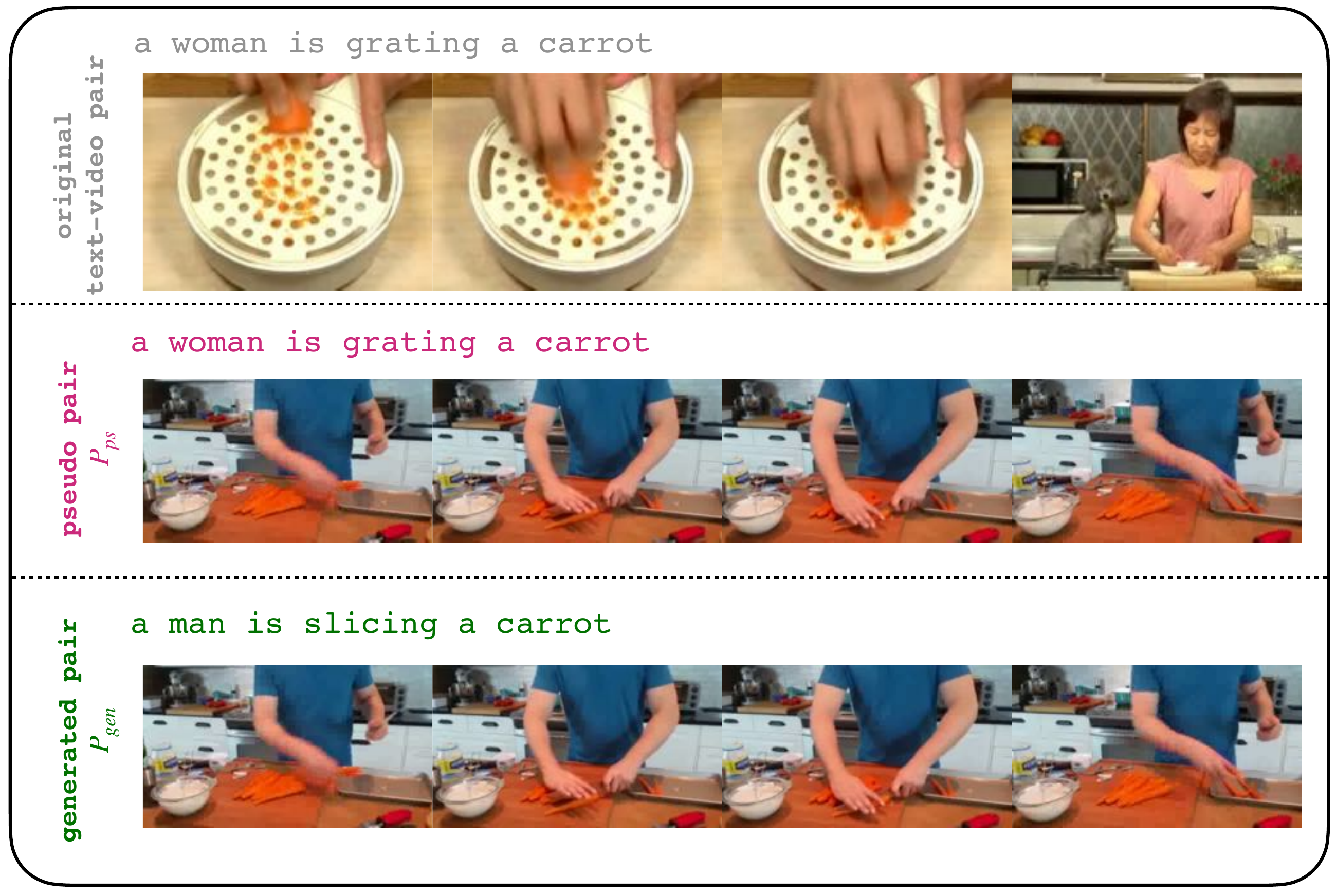}
\includegraphics[scale=0.19]{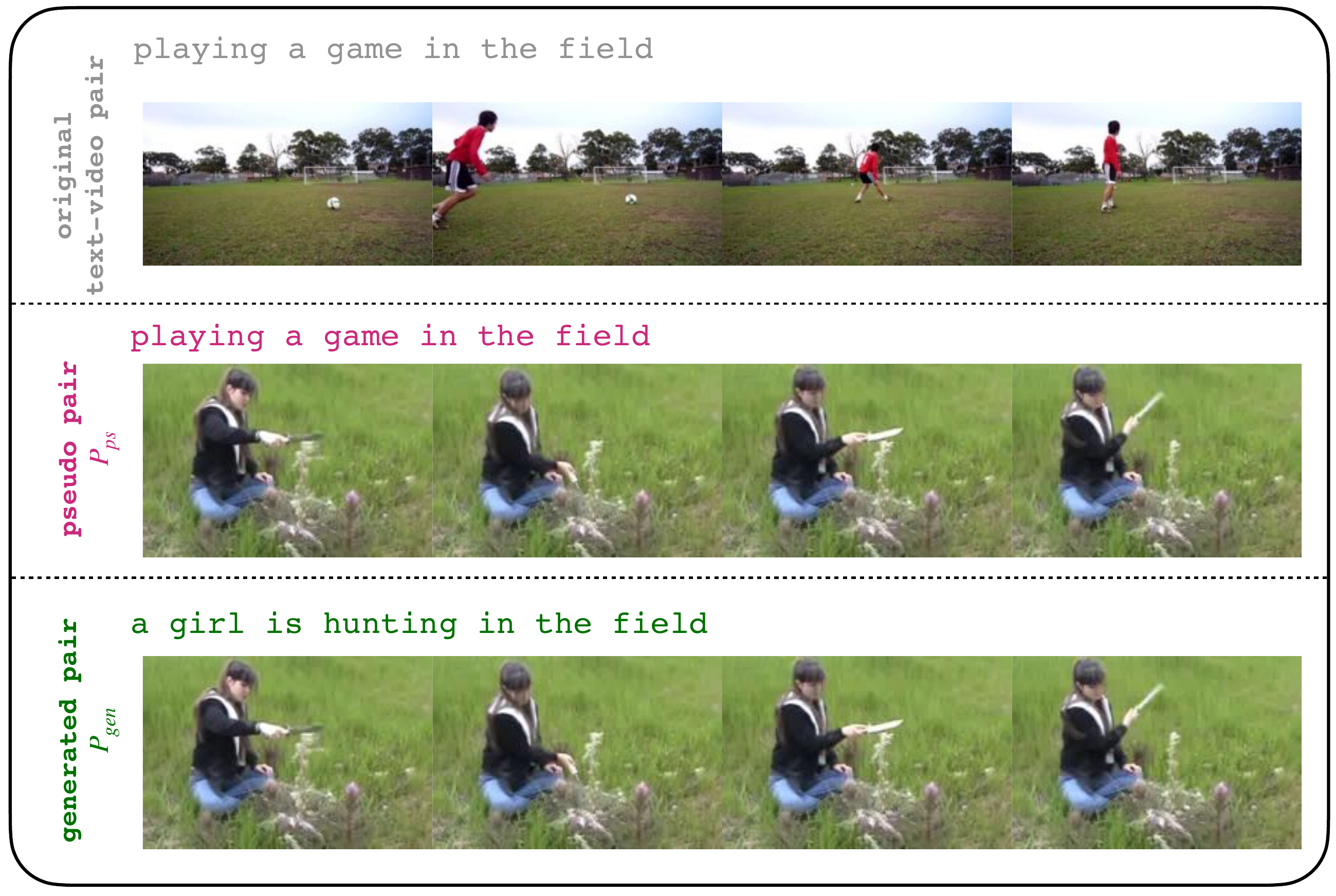}
\includegraphics[scale=0.19]{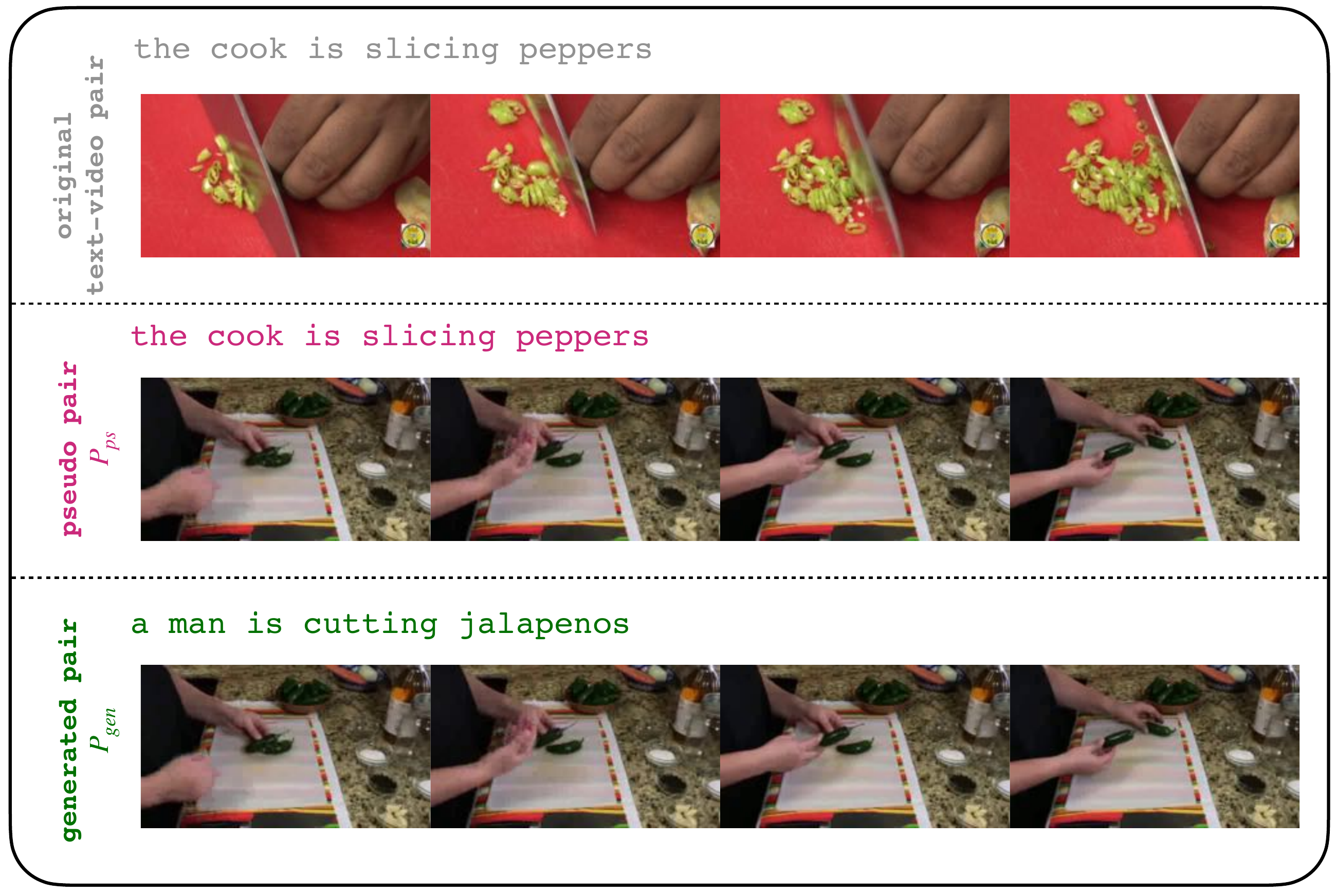}
\includegraphics[scale=0.19]{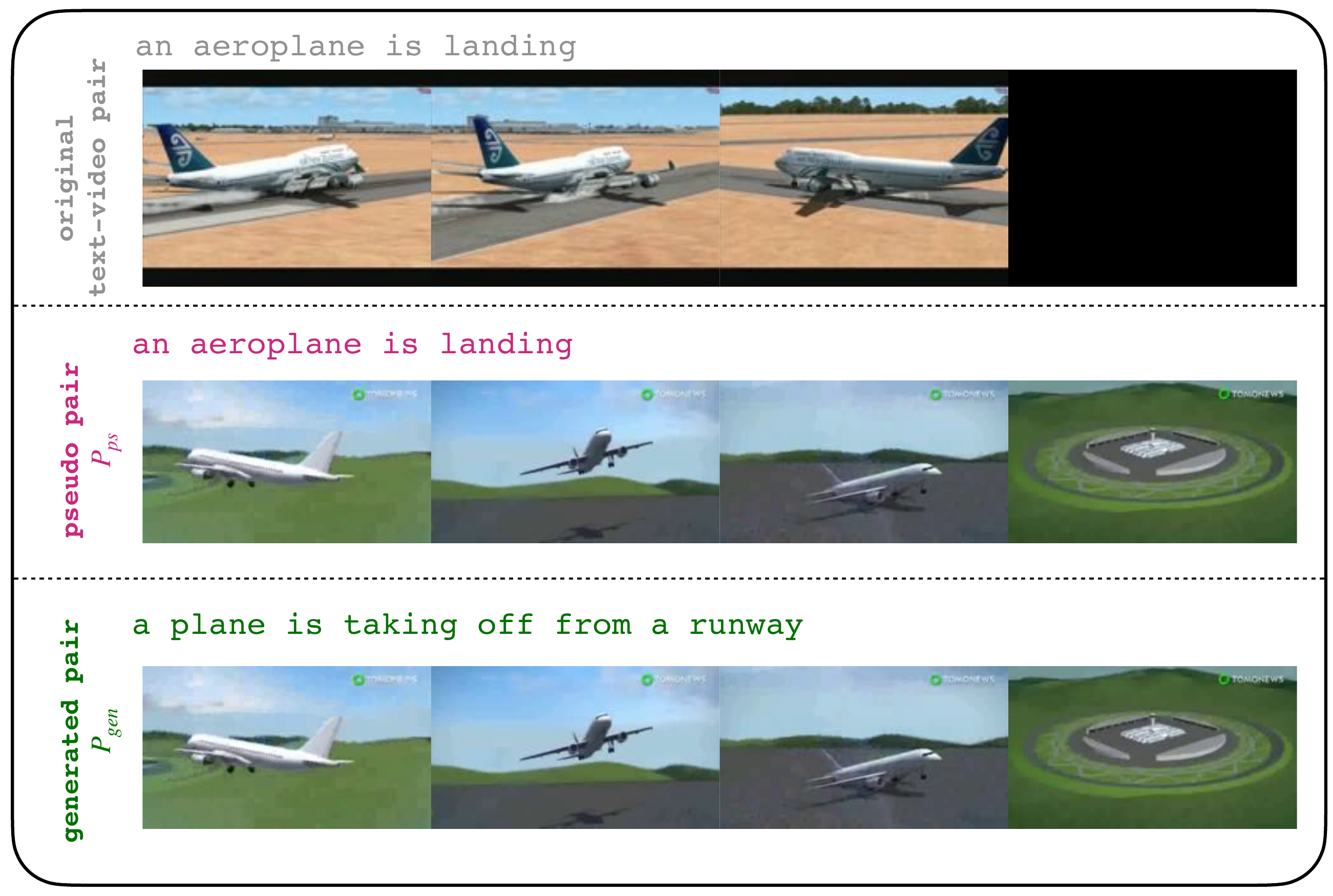}
\includegraphics[scale=0.19]{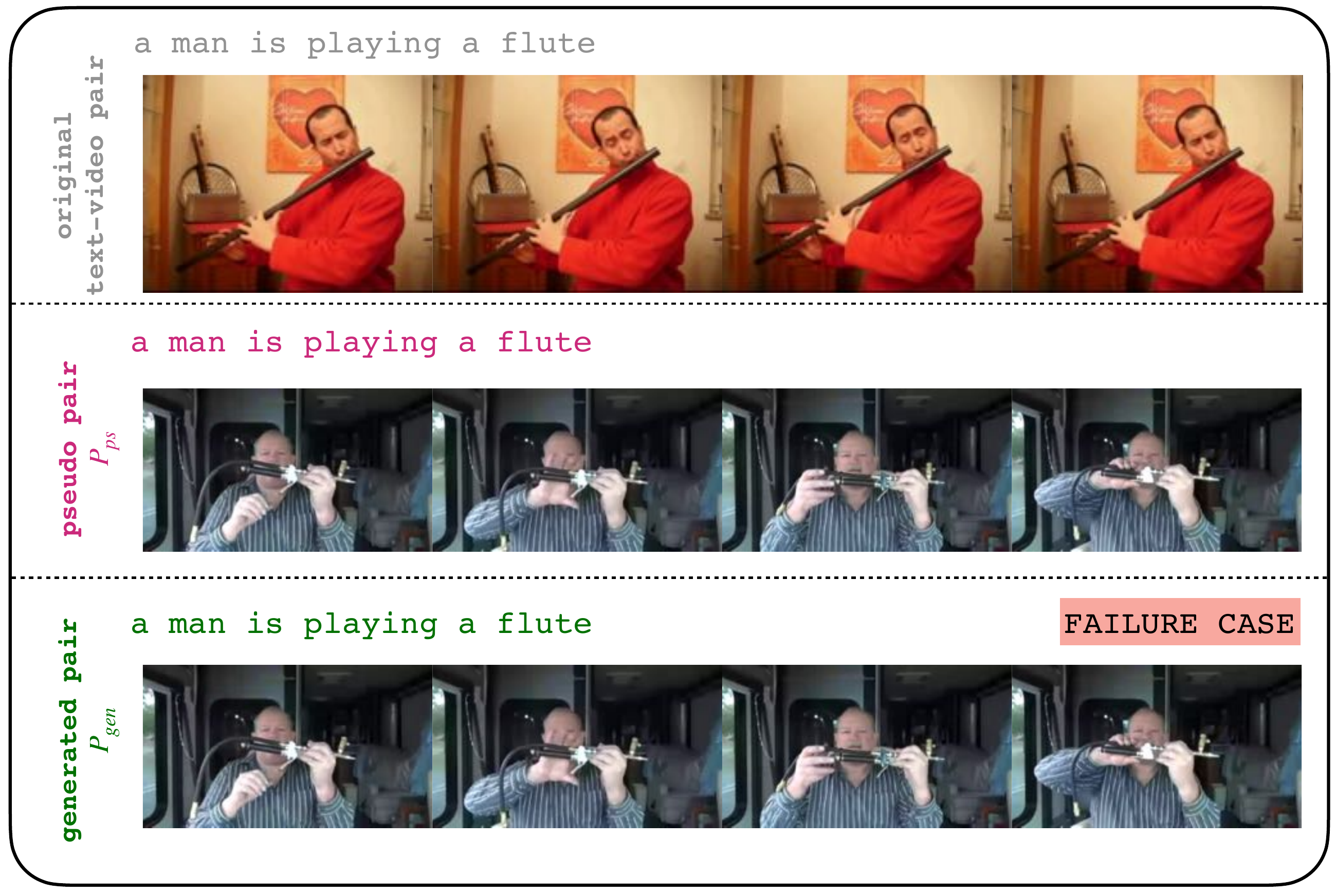}
\includegraphics[scale=0.19]{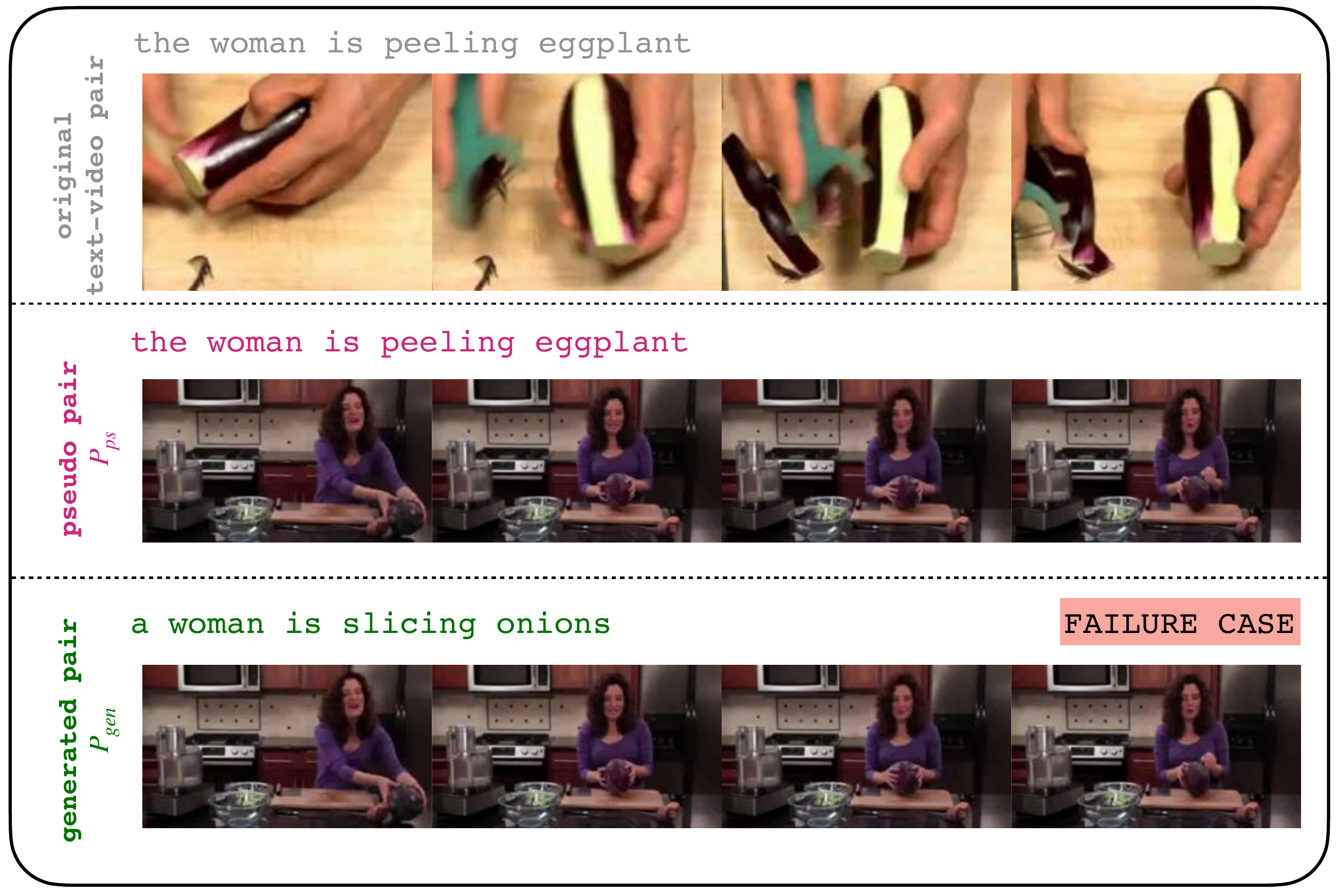}
\end{center}
\vspace{-1.1em}
\caption{ \small{\textbf{Qualitative evaluation of $P_{ps}$ and   $P_{gen}$ on the MSVD.} First, a text query is matched with one of the videos (a pseudo pair $P_{ps}$), and then, after the style transfer step, for each video, a new caption is generated in the same style but with updated content (a generated pair $P_{gen}$)}
}
\label{fig:msvd_qual}
\end{figure*}

\begin{figure*}[!t]
\begin{center}
\includegraphics[scale=0.19]{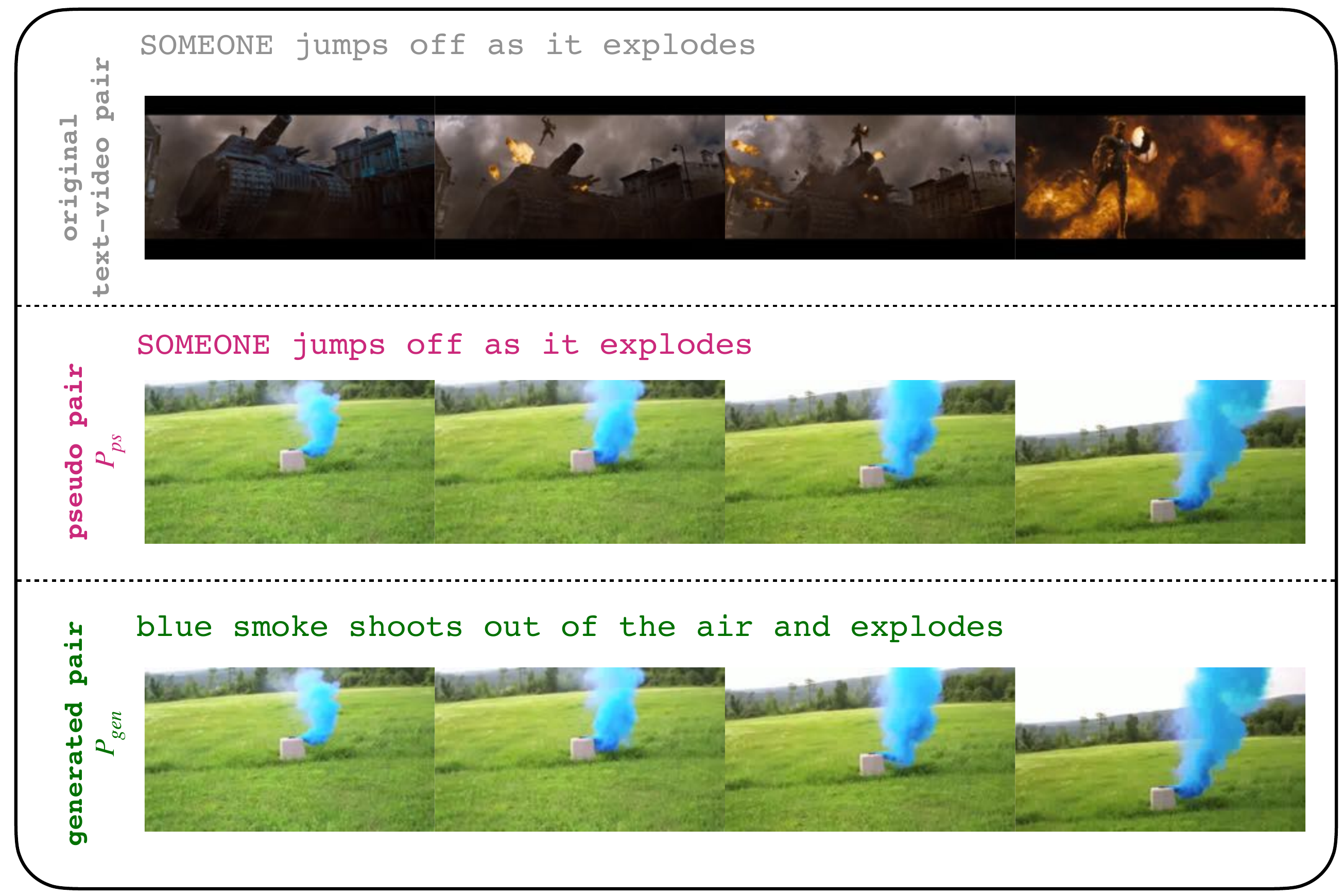}
\includegraphics[scale=0.19]{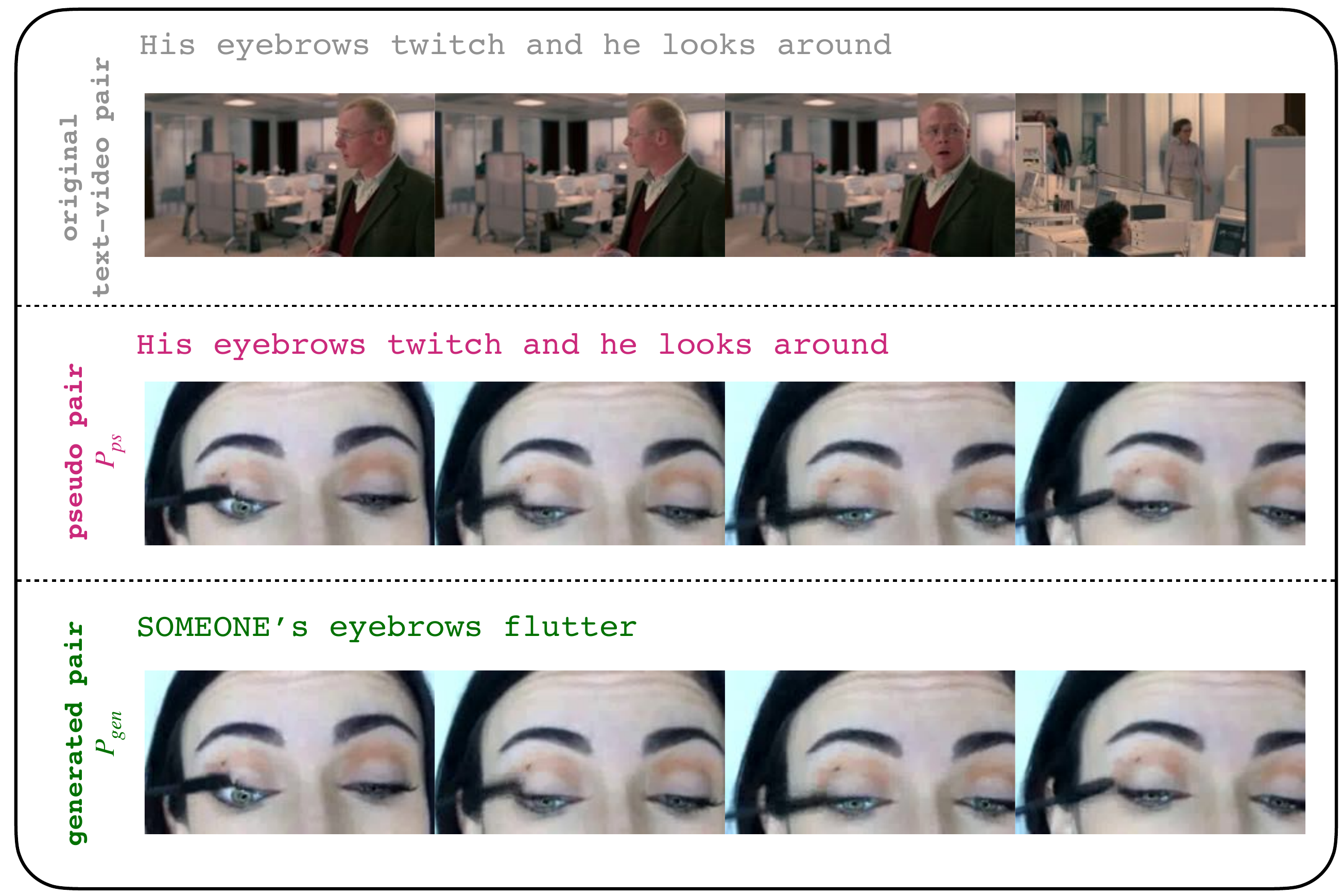}
\includegraphics[scale=0.19]{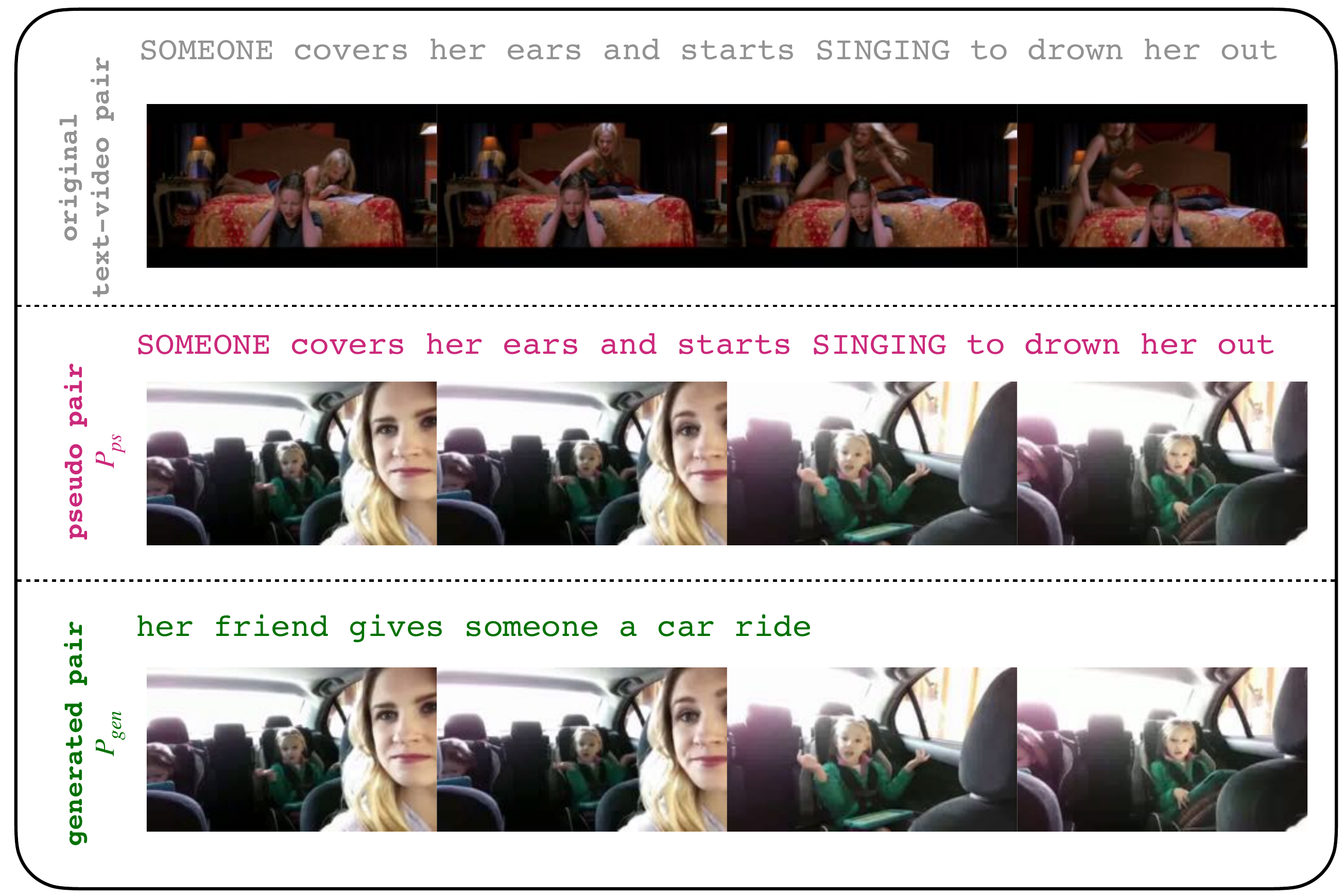}
\includegraphics[scale=0.19]{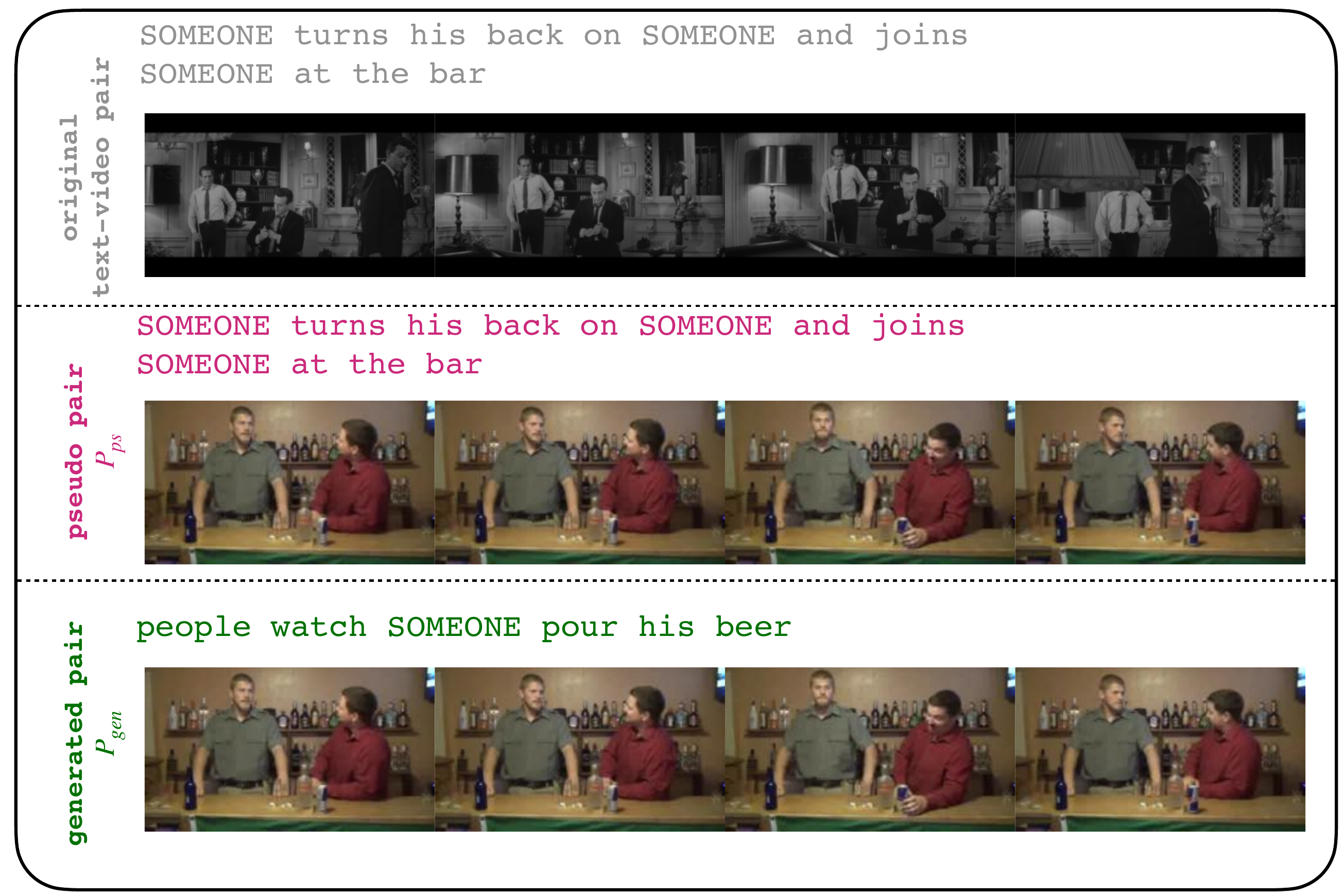}
\includegraphics[scale=0.19]{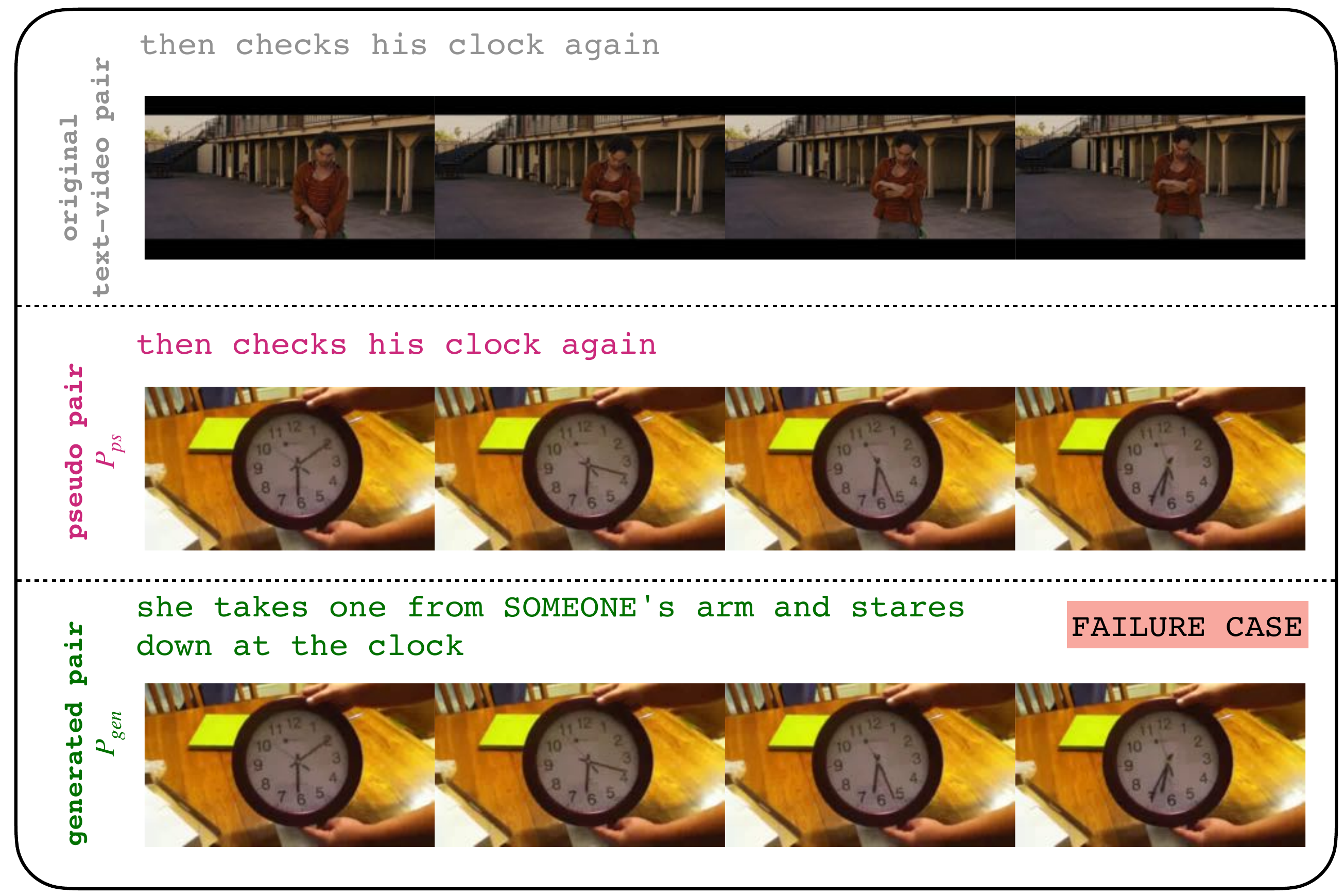}
\includegraphics[scale=0.19]{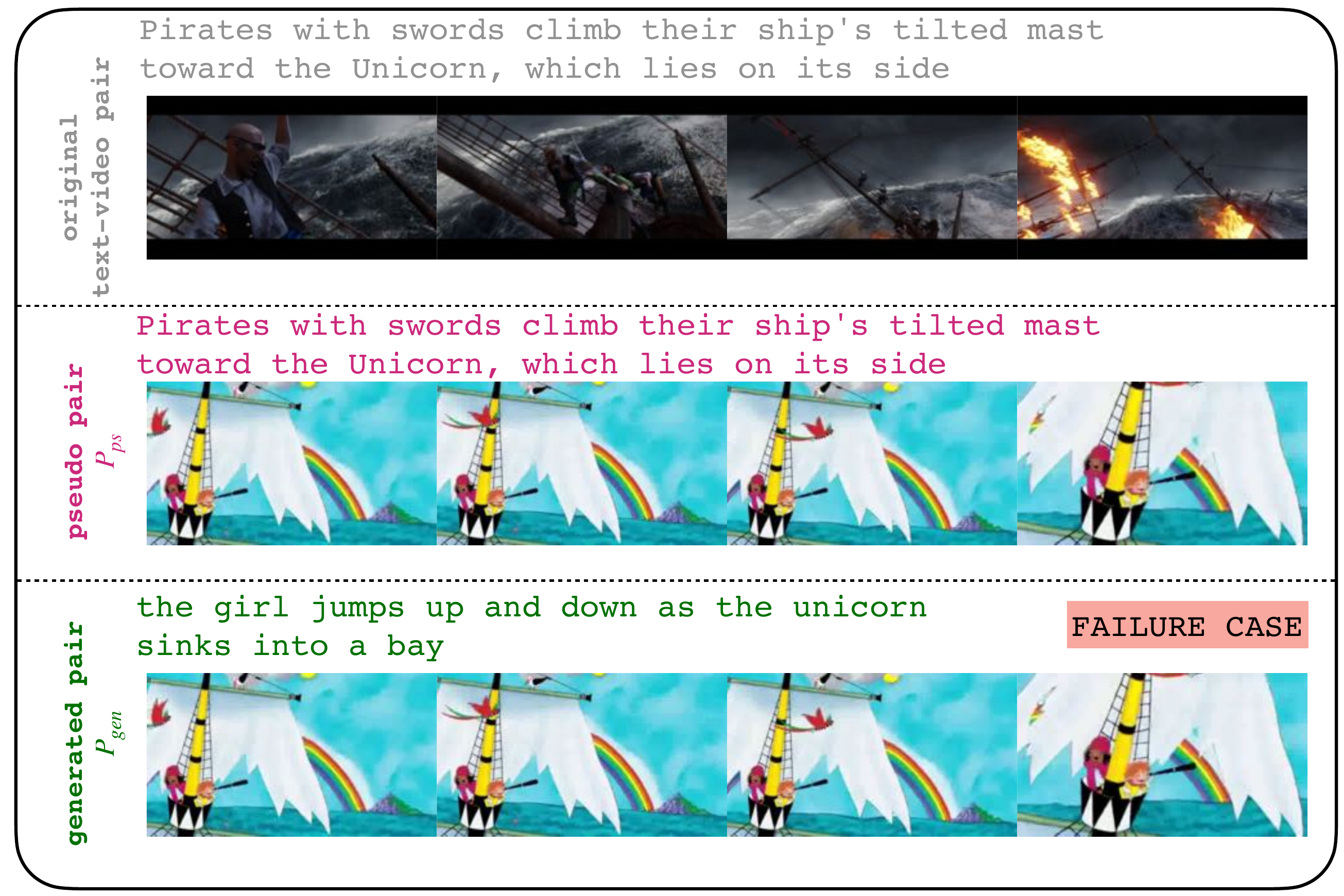}
\end{center}
\vspace{-1.1em}
\caption{ \small{\textbf{Qualitative evaluation of $P_{ps}$ and   $P_{gen}$ on the LSMDC.} First, a text query is matched with one of the videos (a pseudo pair $P_{ps}$), and then, after the style transfer step, for each video, a new caption is generated in the same style but with updated content (a generated pair $P_{gen}$)}
}
\label{fig:lsmdc_qual}
\end{figure*}

\paragraph{Pseudo pairs $P_{ps}$ and generated pairs $P_{gen}$.}
We provide additional qualitative results for pseudo pairs $P_{ps}$ and generated pairs $P_{gen}$ with the MSR-VTT dataset in Figure~\ref{fig:msrvtt_qual}, the YouCook2 in Figure~\ref{fig:youcook_qual}, the DiDeMo in Figure~\ref{fig:didemo_qual}, the MSVD  in Figure~\ref{fig:msvd_qual}, and the LSMDC dataset in Figure~\ref{fig:lsmdc_qual}. We observe on various datasets that generated captions capture content better than in initially obtained pseudo pairs after the matching step. It confirms our discussion of Table~\ref{tab:style_preservation} (in the main paper), that generated pairs $P_{gen}$ on average provide better improvement than pseudo pairs $P_{ps}$.

\begin{figure*}[!t]
\begin{center}
\includegraphics[scale=0.19]{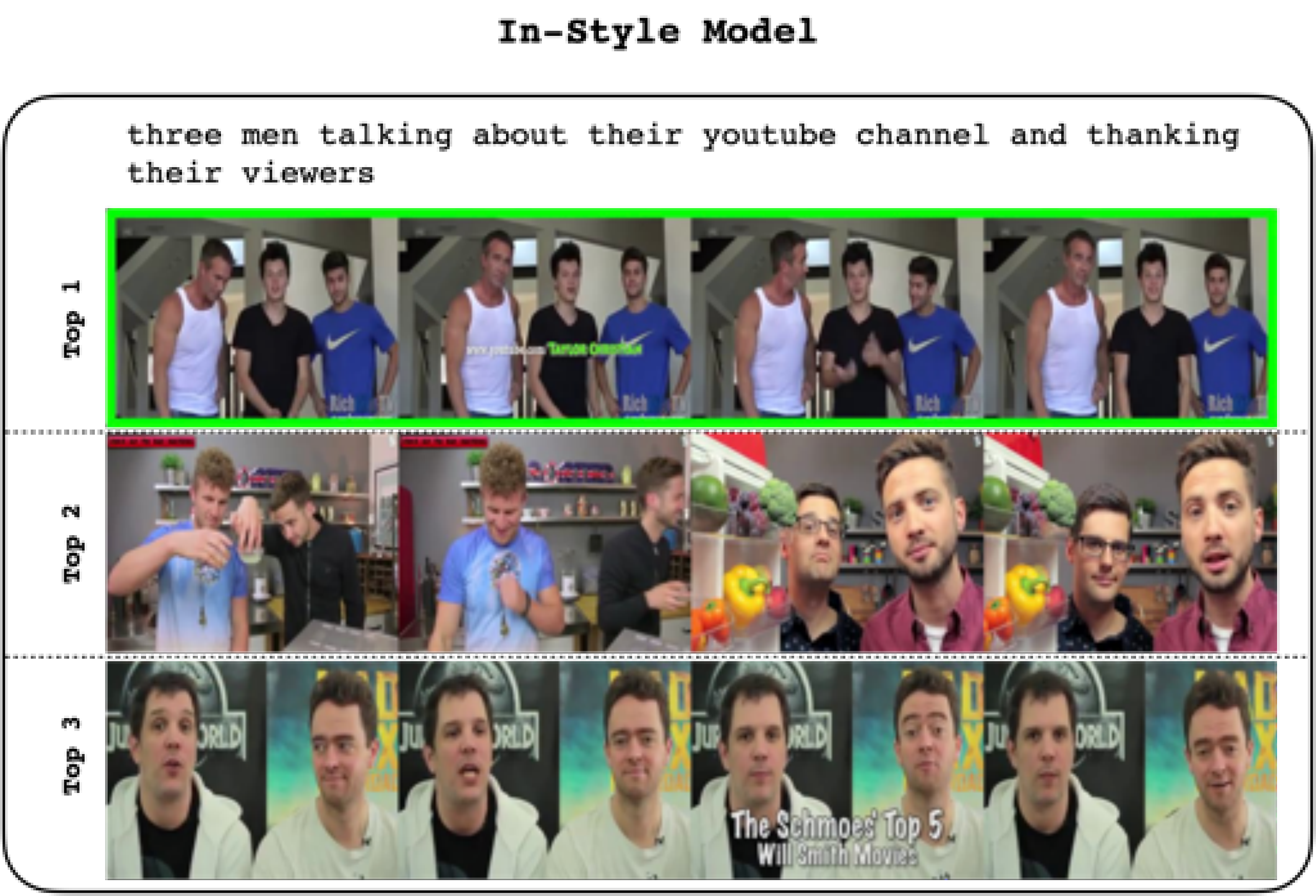}
\vspace{2mm}
\includegraphics[scale=0.19]{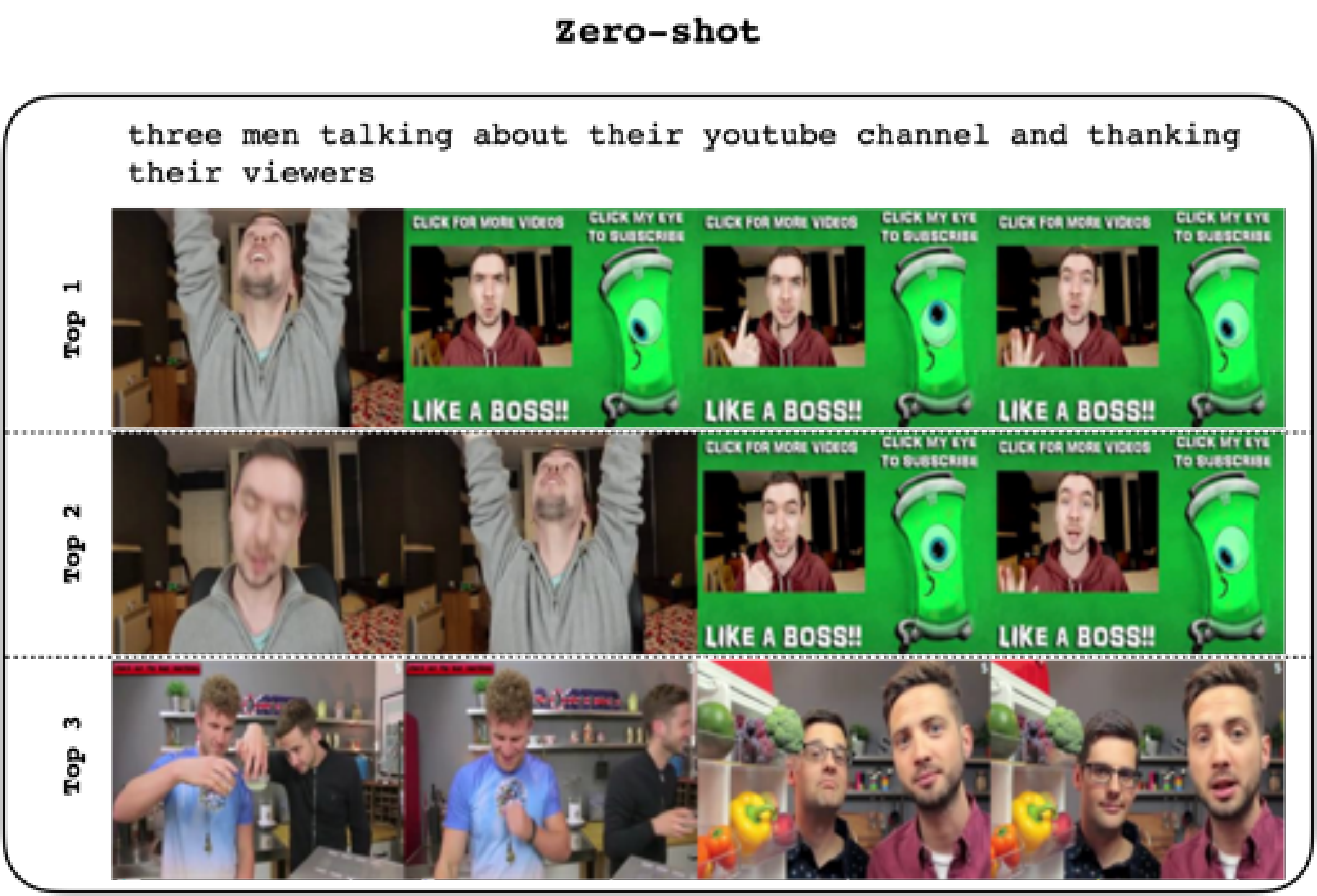}
\vspace{2mm}
\includegraphics[scale=0.19]{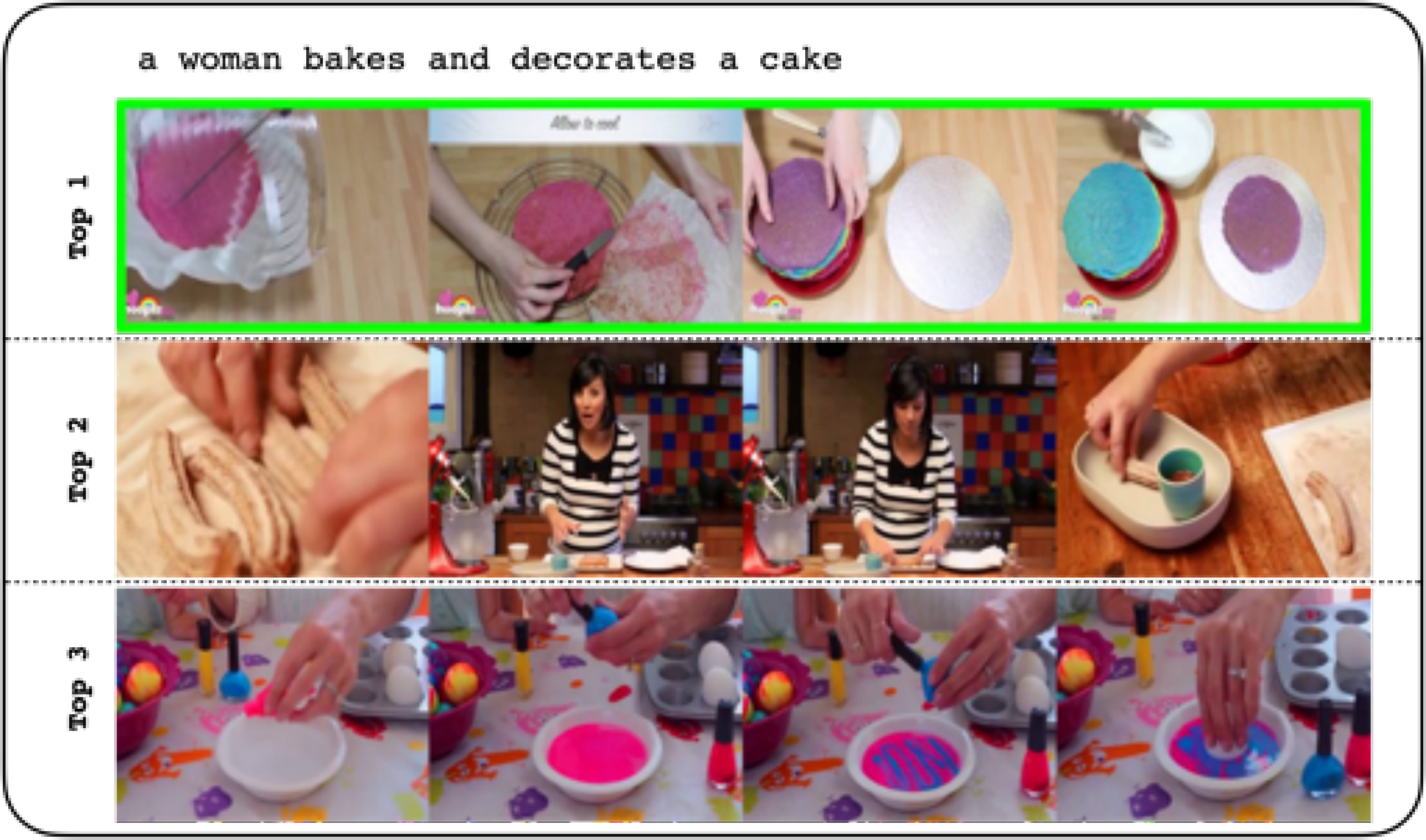}
\vspace{1mm}
\includegraphics[scale=0.19]{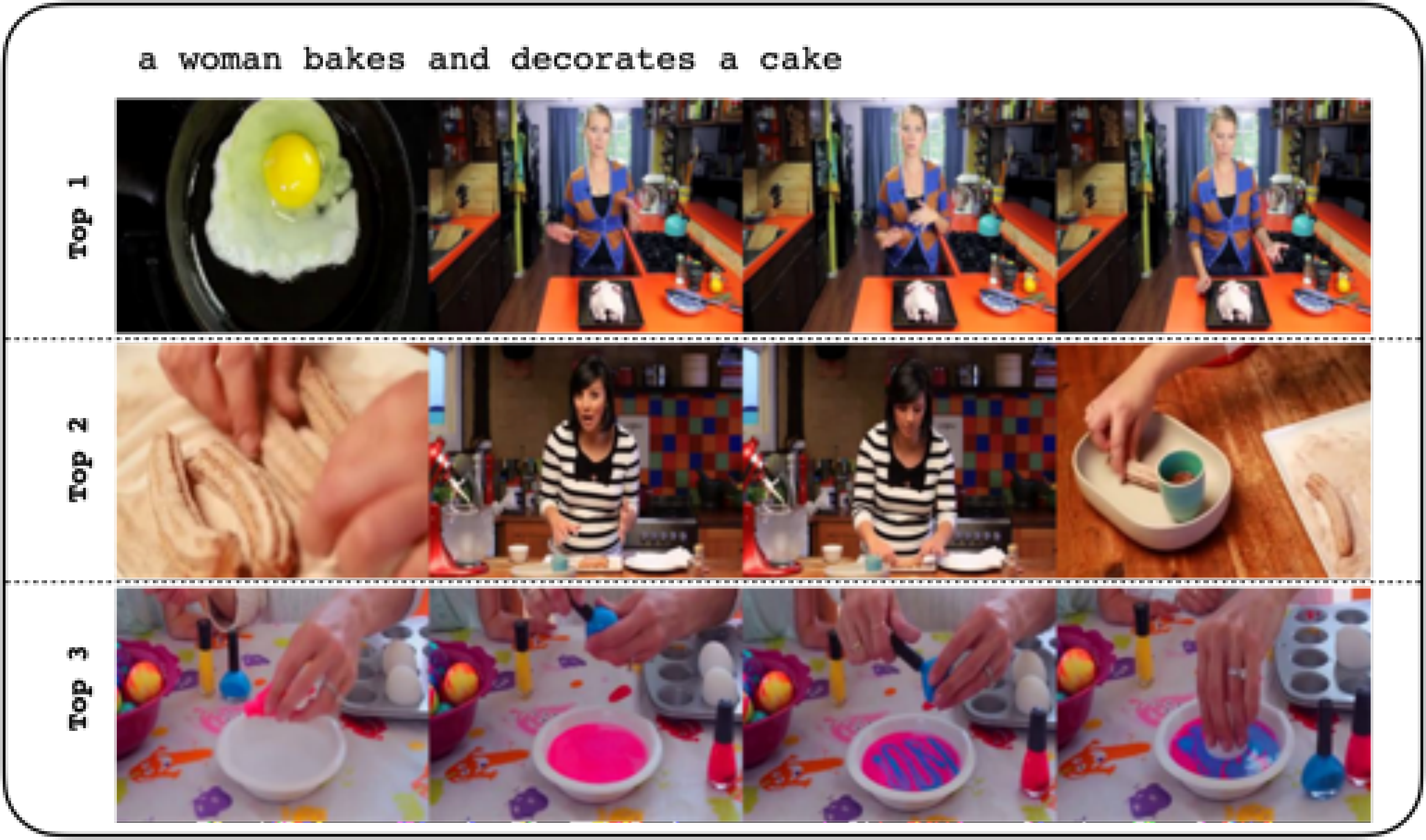}
\vspace{1mm}
\includegraphics[scale=0.19]{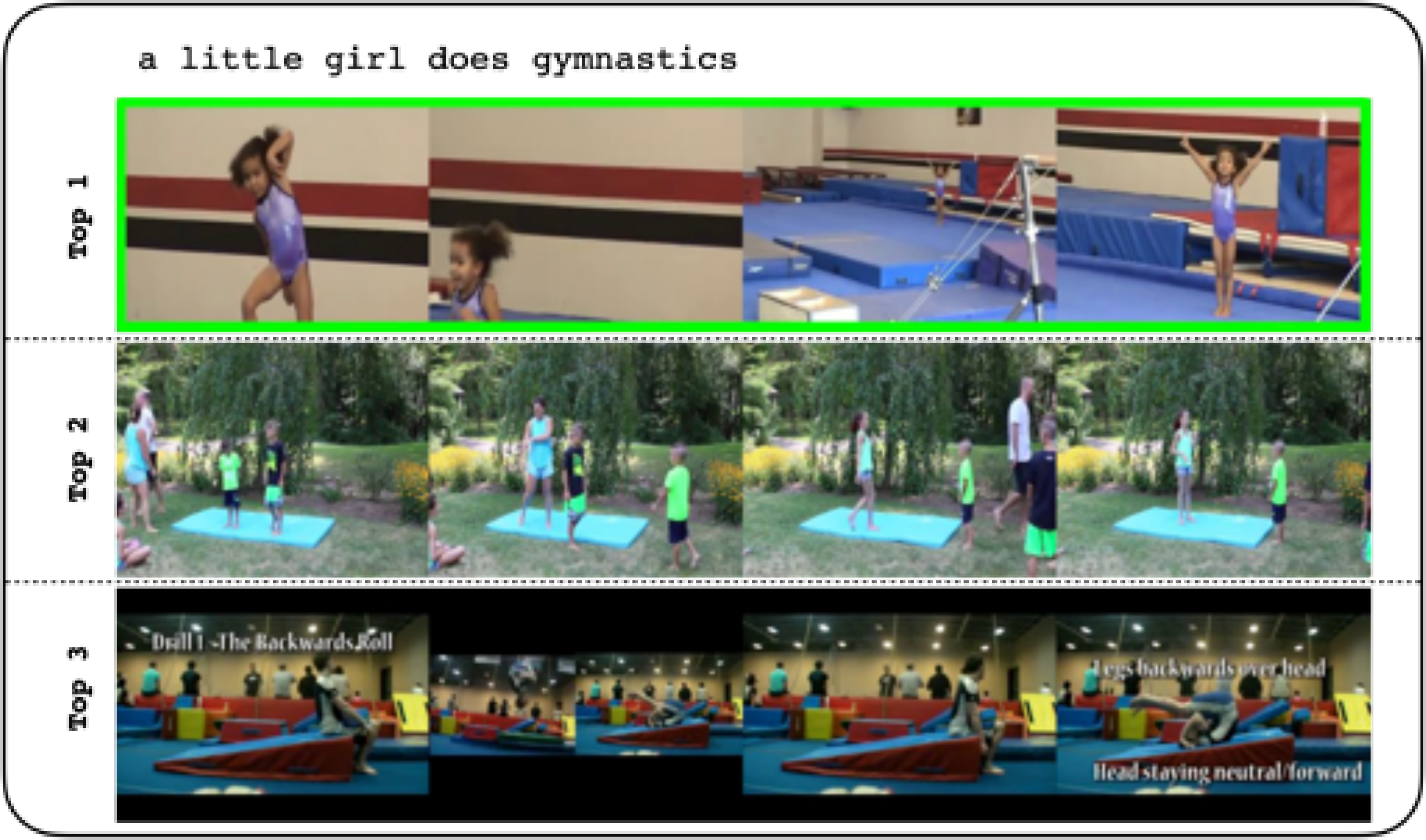}
\includegraphics[scale=0.19]{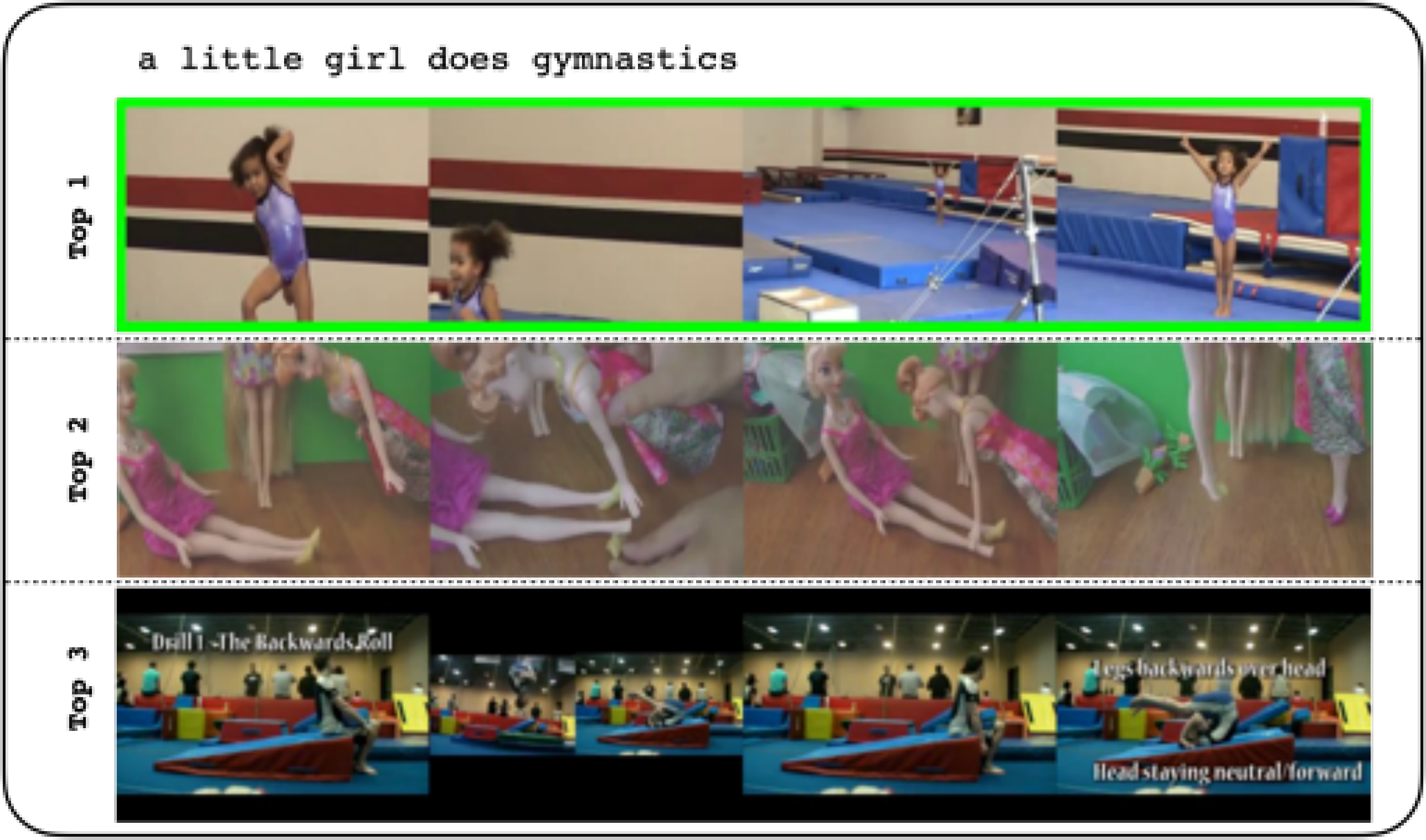}
\end{center}
\vspace{-1.1em}
\caption{\textbf{Qualitative evaluation of text-video retrieval on the MSR-VTT.} Retrieval examples for the proposed In-Style Model and zero-shot BLIP model. Each box shows the top-3 retrieved videos for a given text query. The correct video is highlighted with a green color.
}
\label{fig:msrvtt_retrieval_qual}
\end{figure*}

\begin{figure*}[!t]
\begin{center}
\includegraphics[scale=0.19]{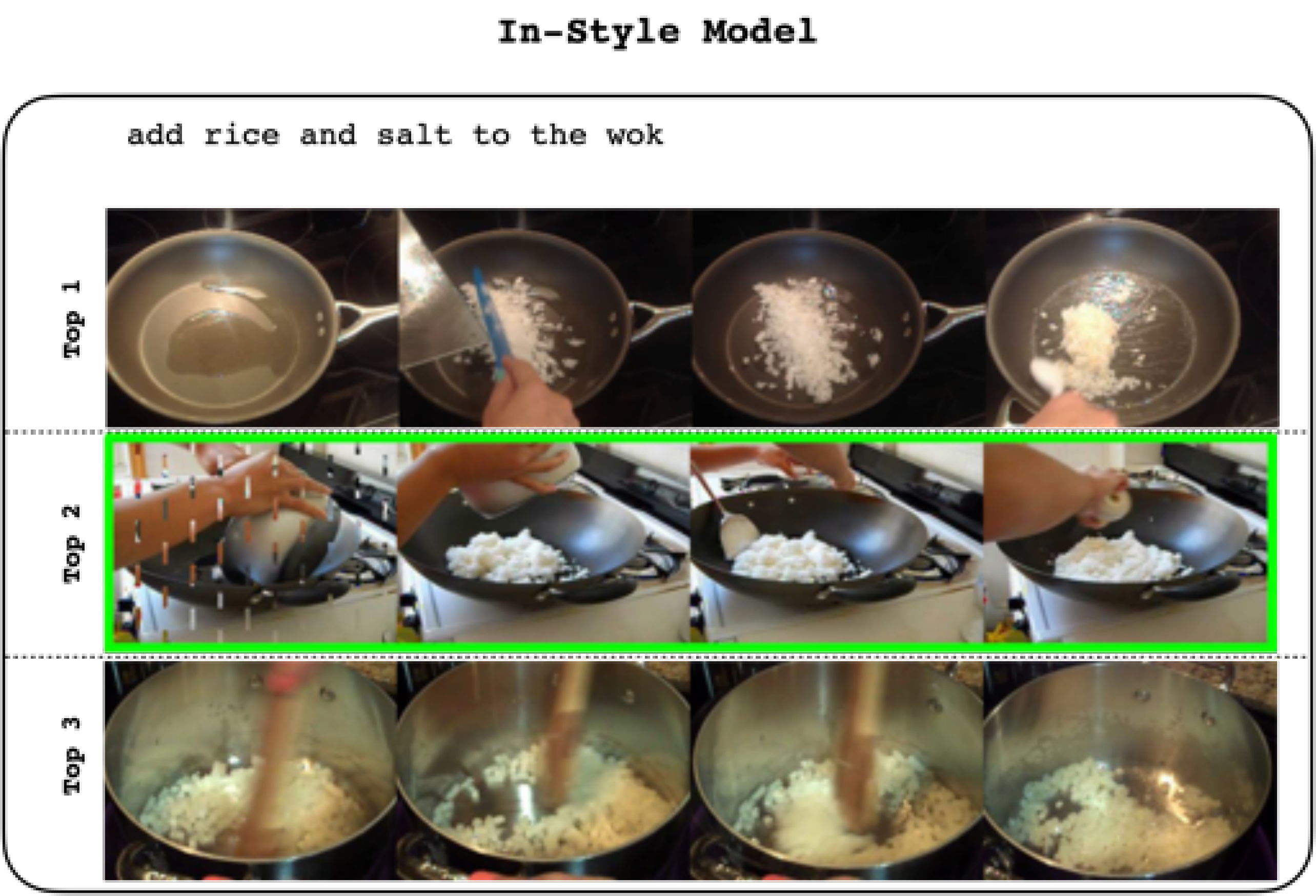}
\vspace{2mm}
\includegraphics[scale=0.19]{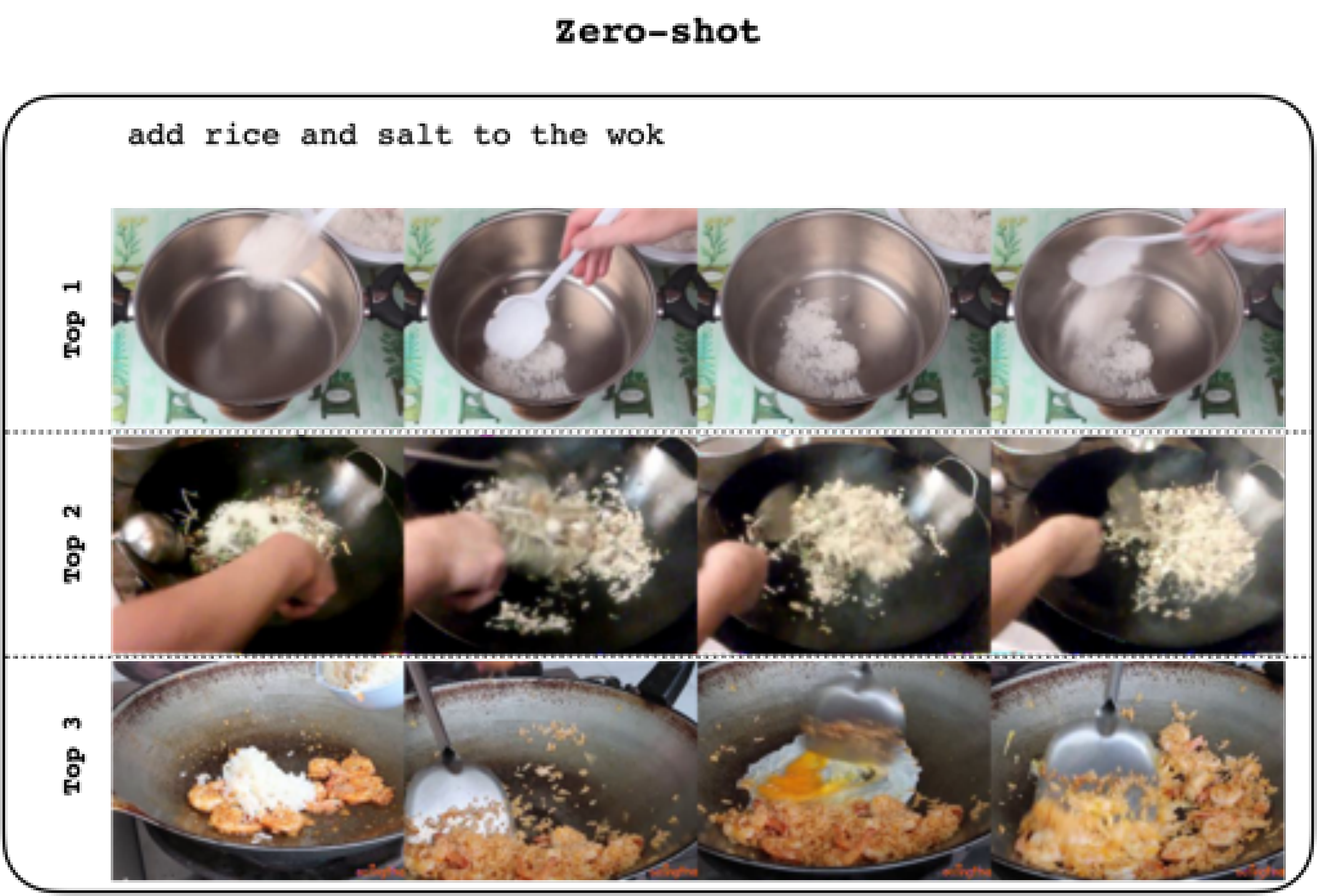}
\vspace{2mm}
\includegraphics[scale=0.19]{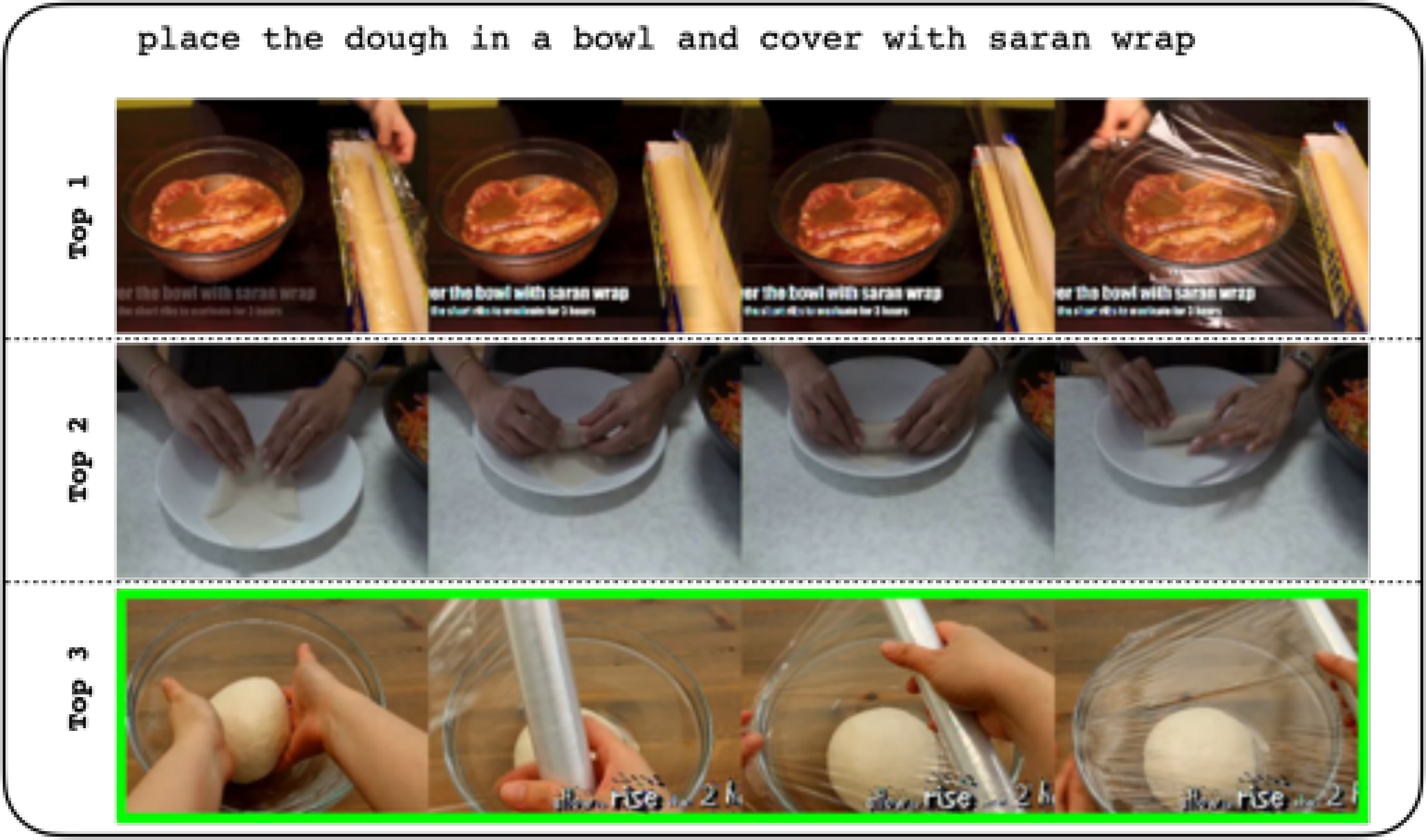}
\vspace{1mm}
\includegraphics[scale=0.19]{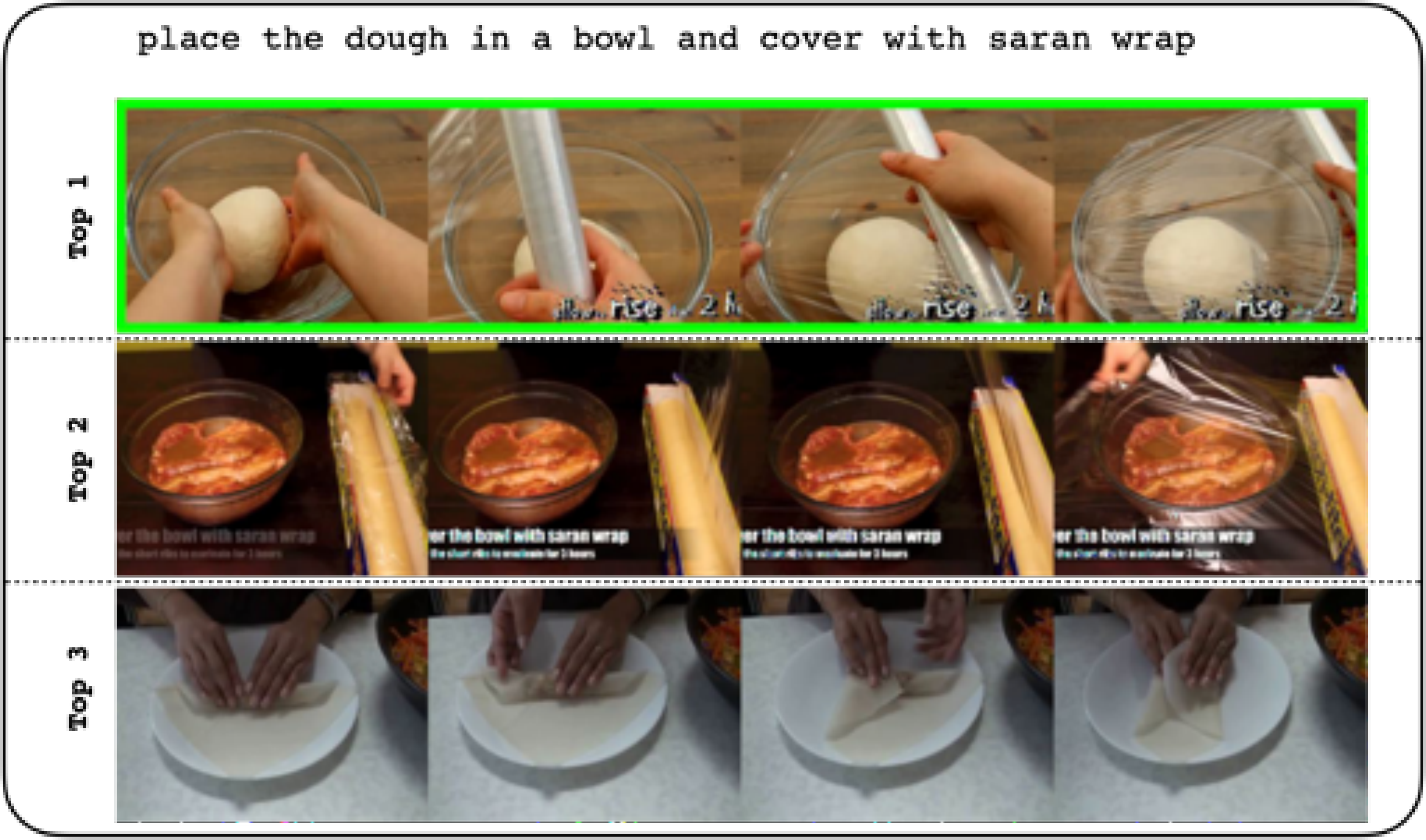}
\vspace{1mm}
\includegraphics[scale=0.19]{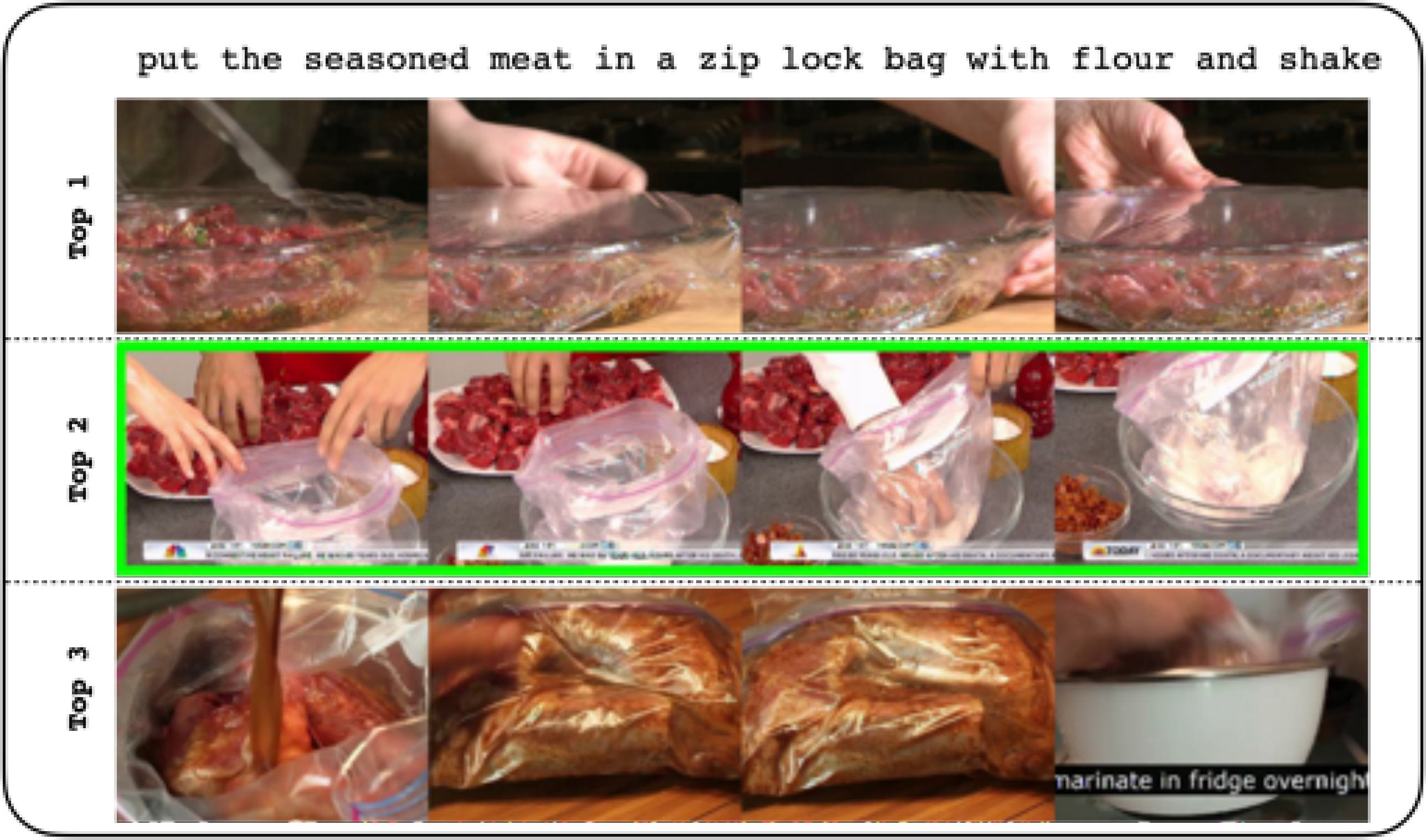}
\includegraphics[scale=0.19]{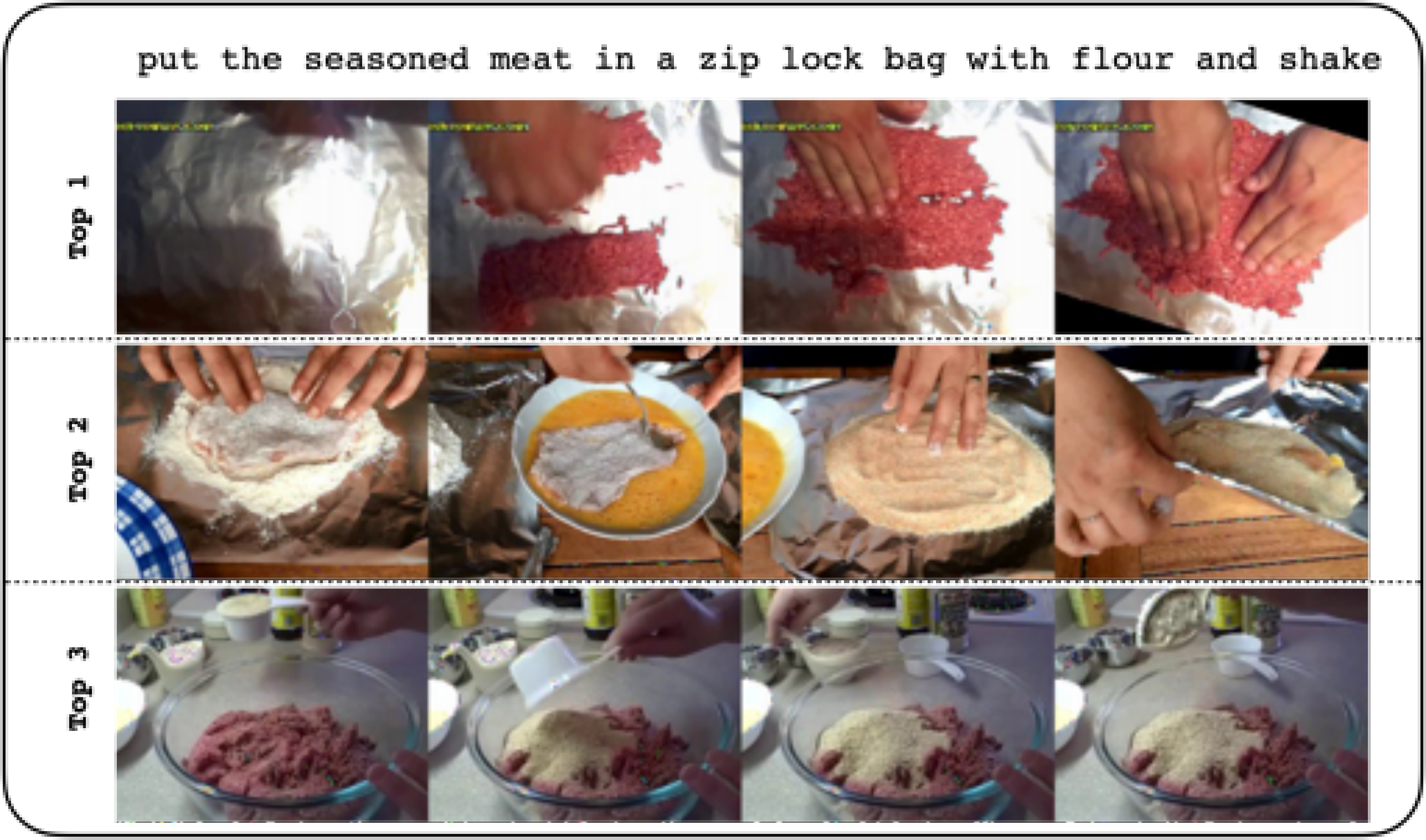}
\end{center}
\vspace{-1.1em}
\caption{\textbf{Qualitative evaluation of text-video retrieval on the YouCook2.} Retrieval examples for the proposed In-Style Model and zero-shot BLIP model. Each box shows the top-3 retrieved videos for a given text query. The correct video is highlighted with a green color.
}
\label{fig:youcook_retrieval_qual}
\end{figure*}

\begin{figure*}[!t]
\begin{center}
\includegraphics[scale=0.19]{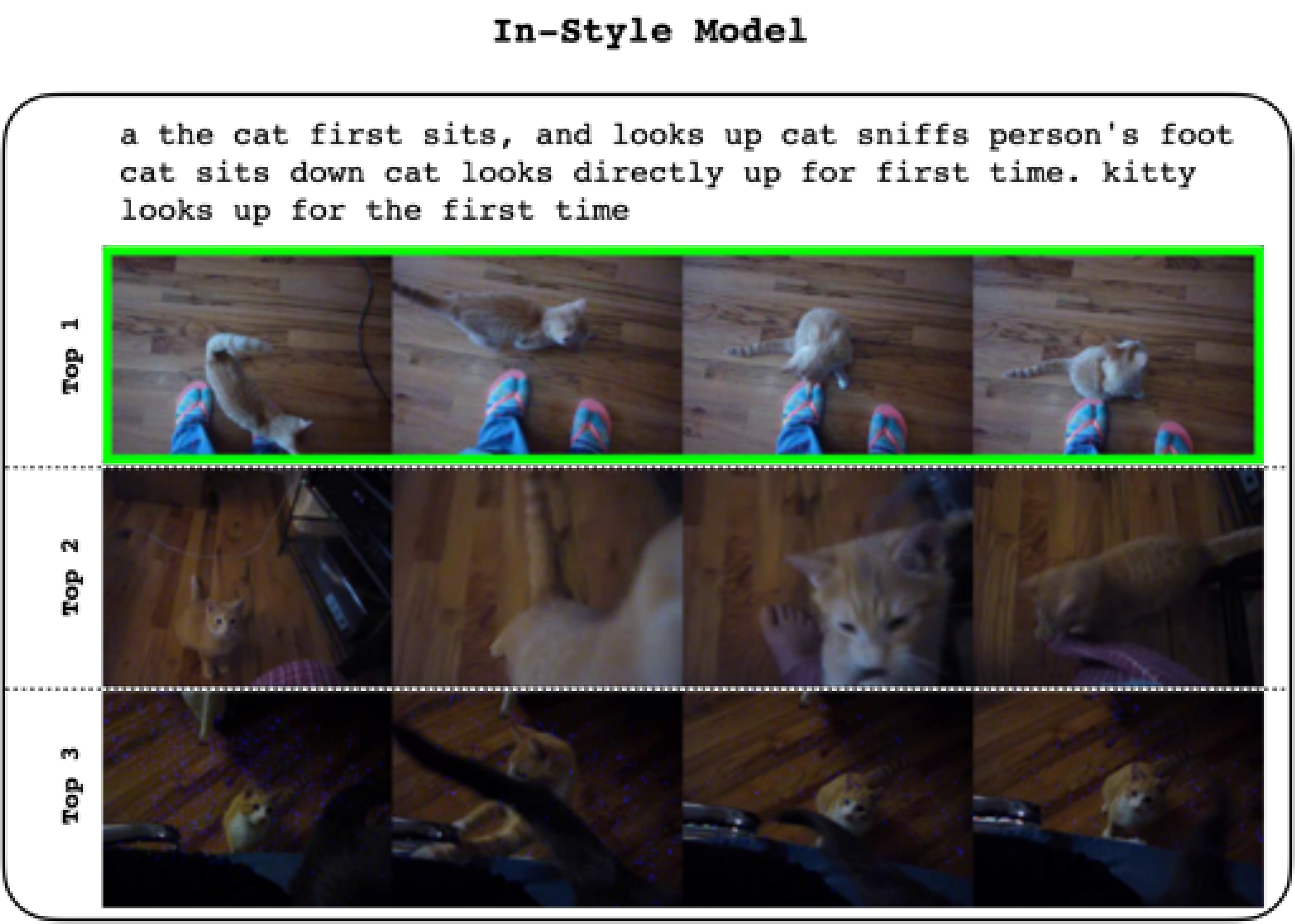}
\vspace{2mm}
\includegraphics[scale=0.19]{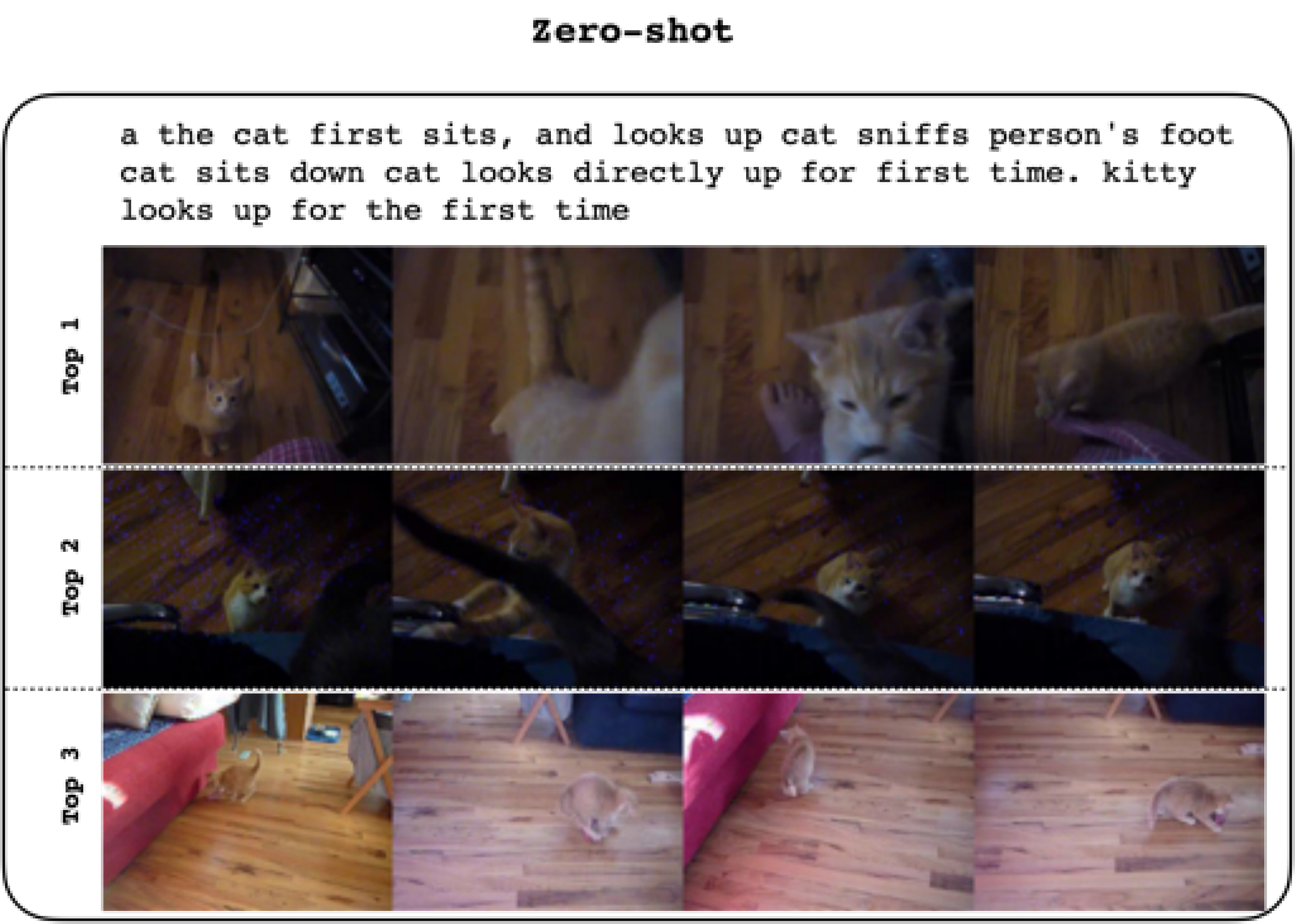}
\vspace{2mm}
\includegraphics[scale=0.19]{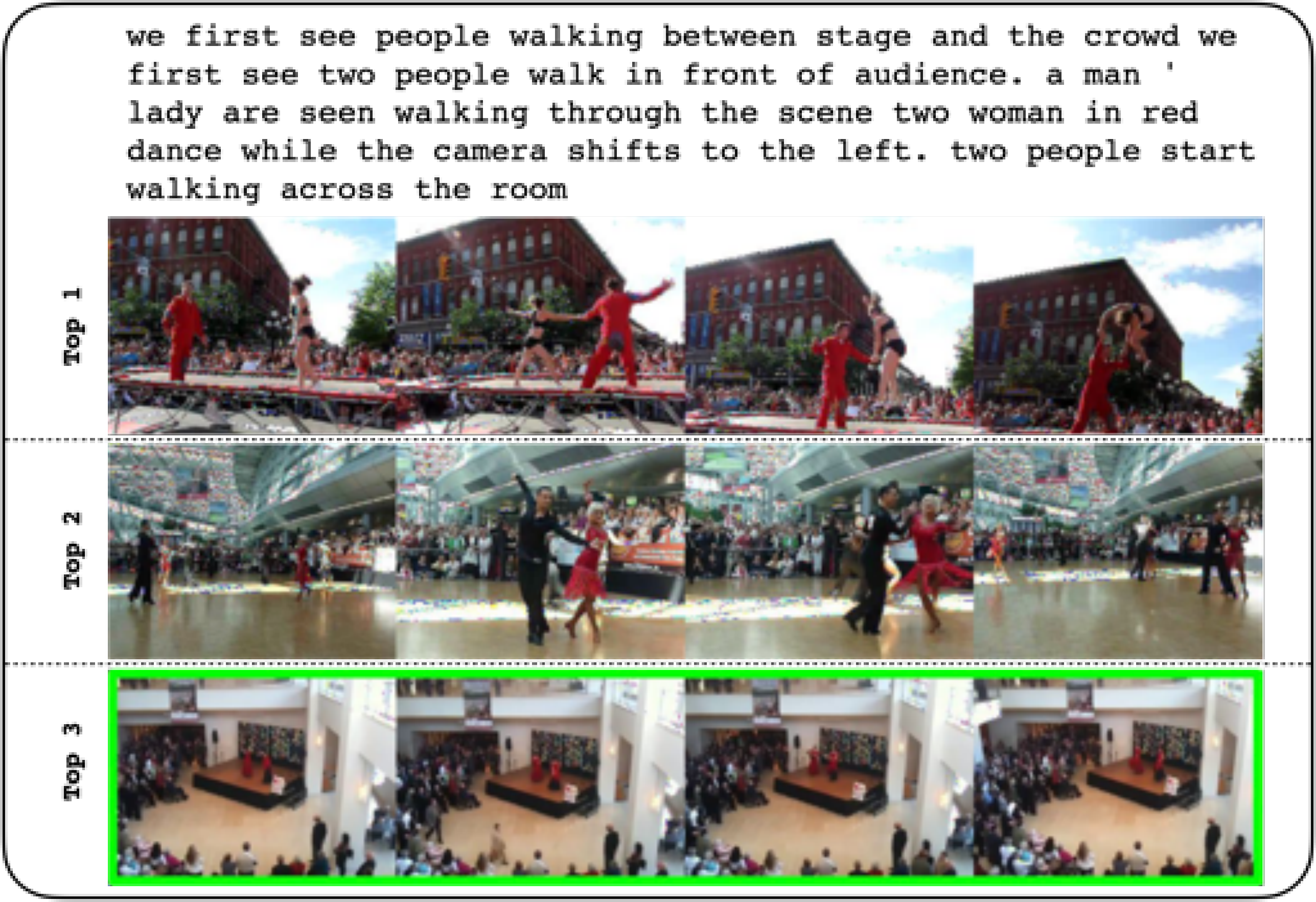}
\vspace{1mm}
\includegraphics[scale=0.19]{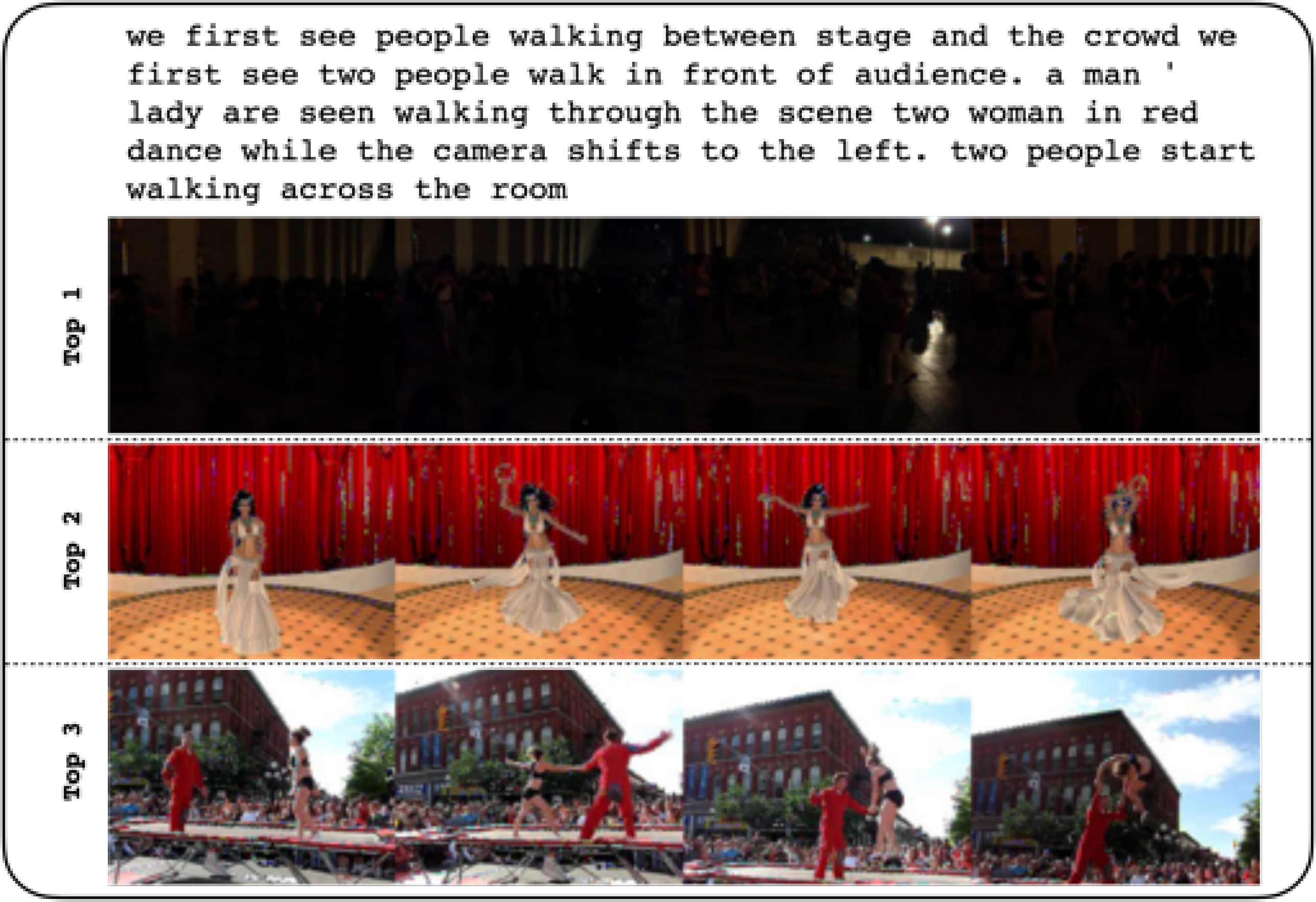}
\vspace{1mm}
\includegraphics[scale=0.19]{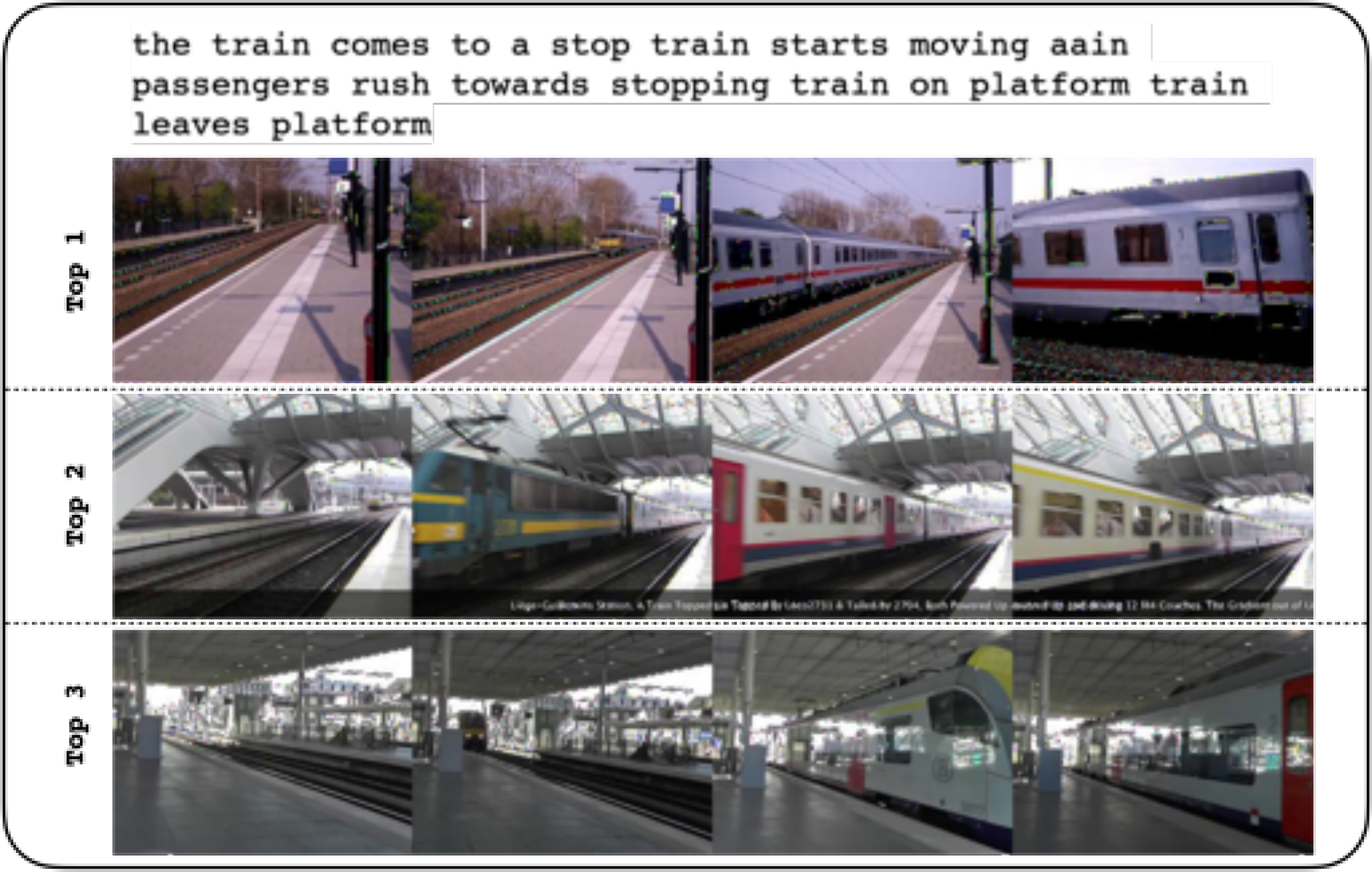}
\includegraphics[scale=0.19]{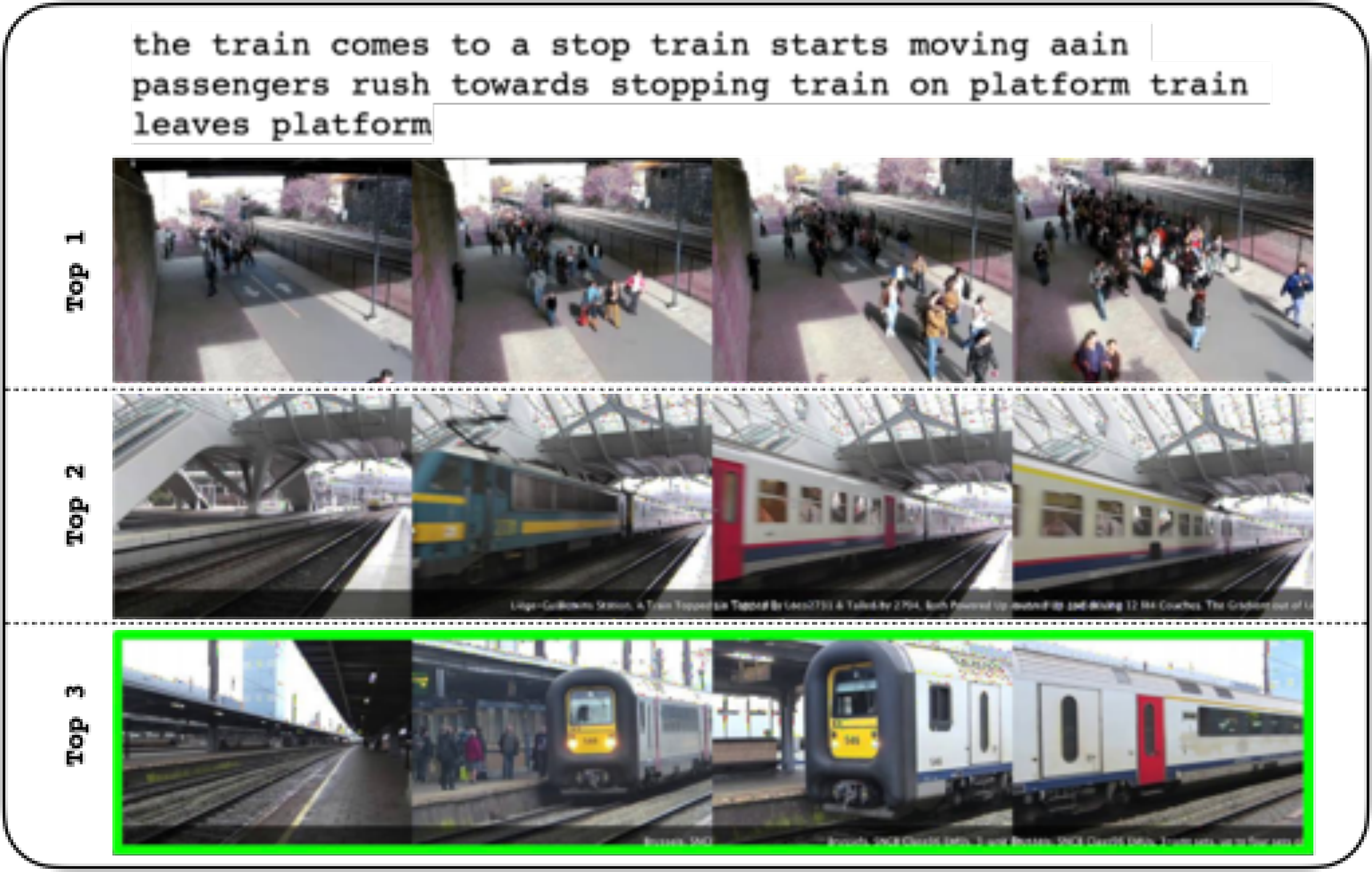}
\end{center}
\vspace{-1.1em}
\caption{\textbf{Qualitative evaluation of text-video retrieval on the DiDeMo.} Retrieval examples for the proposed In-Style Model and zero-shot BLIP model. Each box shows the top-3 retrieved videos for a given text query. The correct video is highlighted with a green color.
}
\label{fig:didemo_retrieval_qual}
\end{figure*}

\begin{figure*}[!t]
\begin{center}
\includegraphics[scale=0.19]{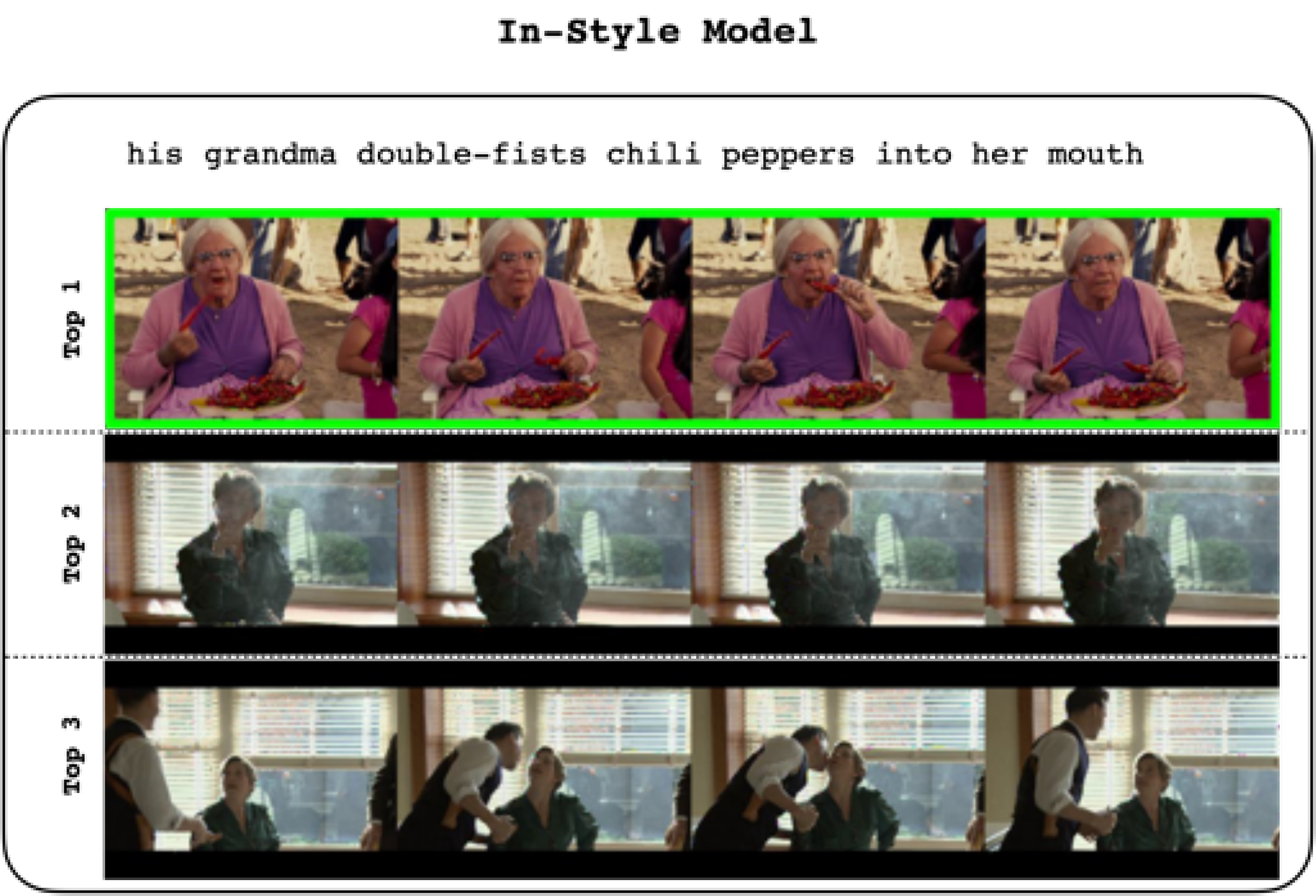}
\vspace{2mm}
\includegraphics[scale=0.19]{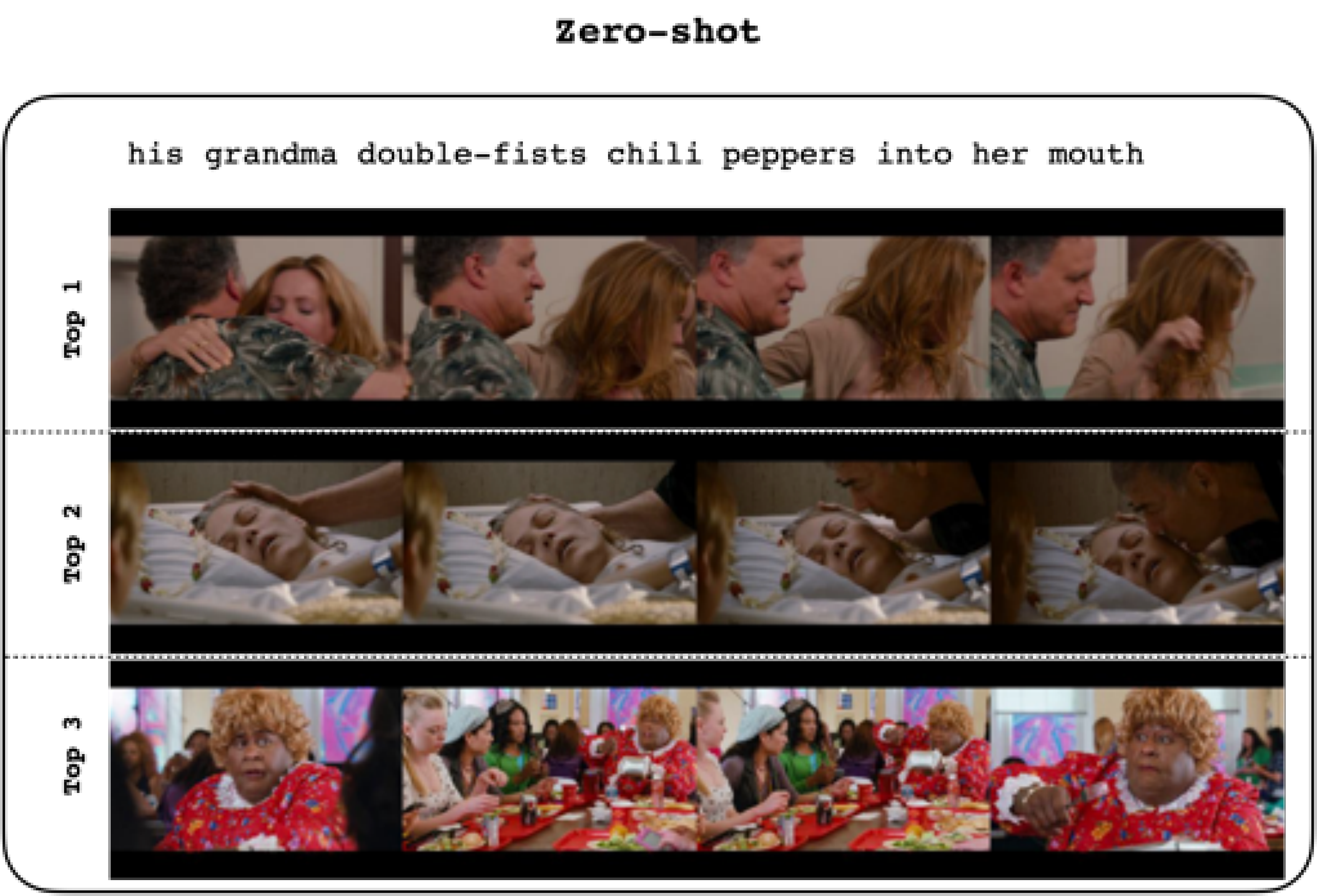}
\vspace{2mm}
\includegraphics[scale=0.19]{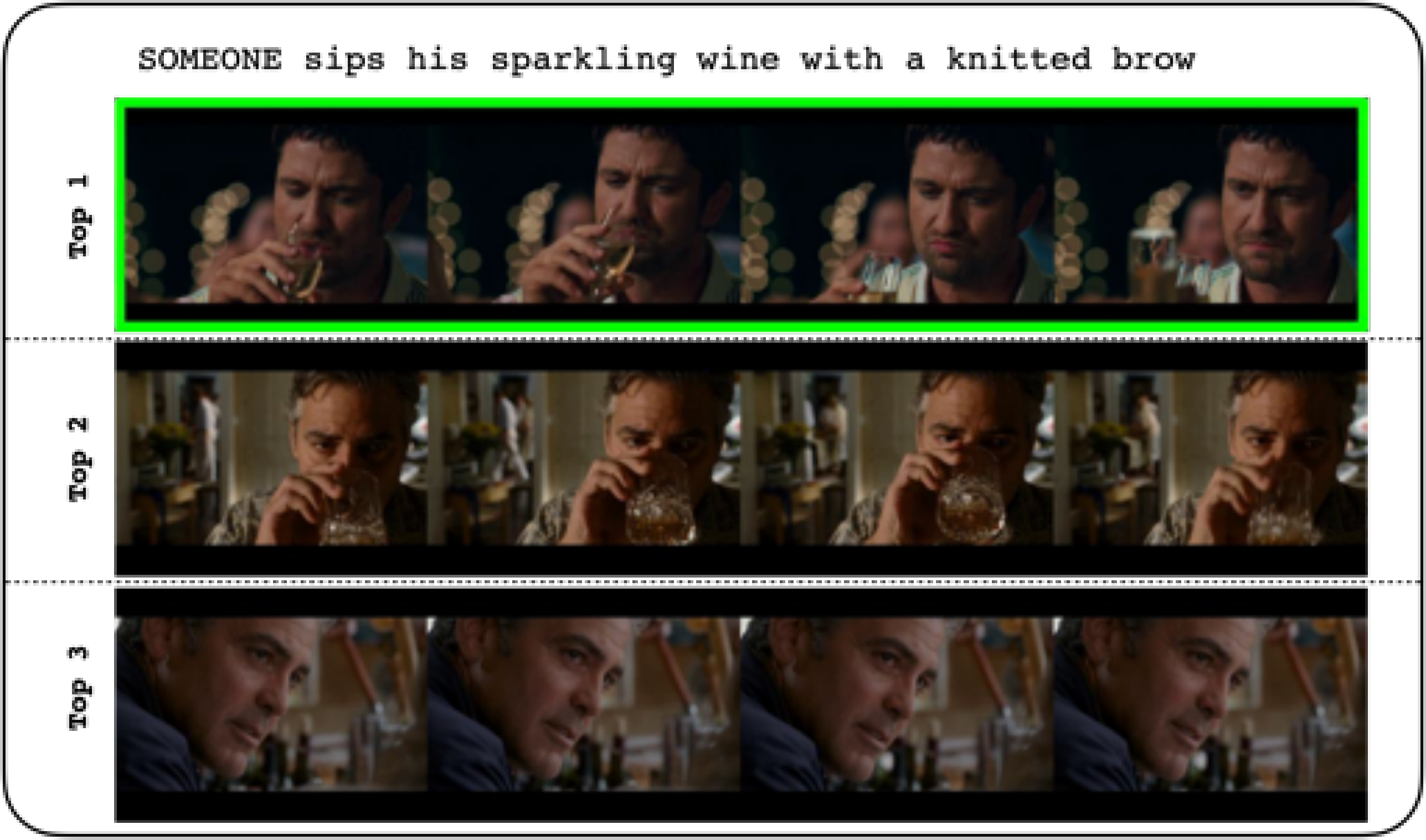}
\vspace{1mm}
\includegraphics[scale=0.19]{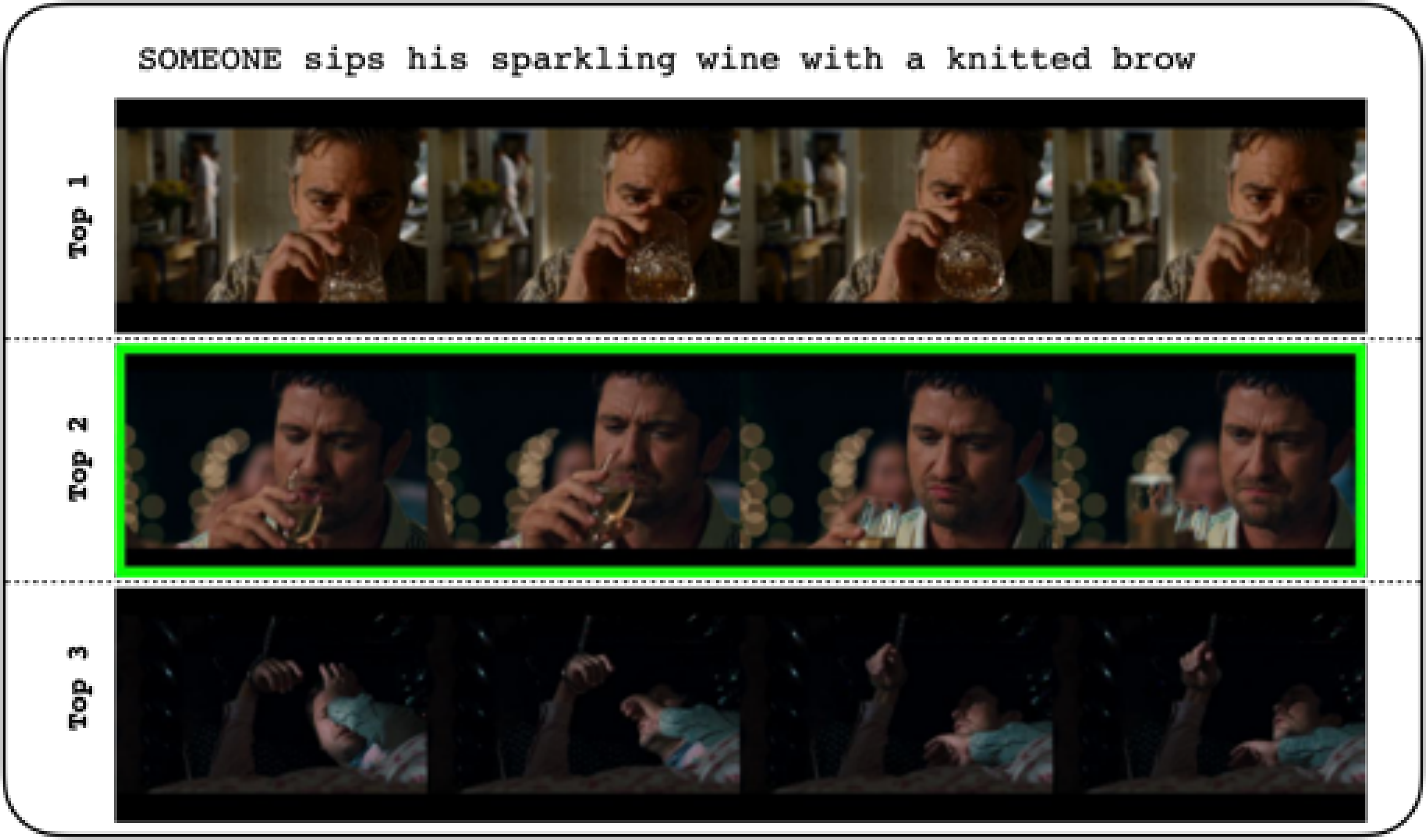}
\vspace{1mm}
\includegraphics[scale=0.19]{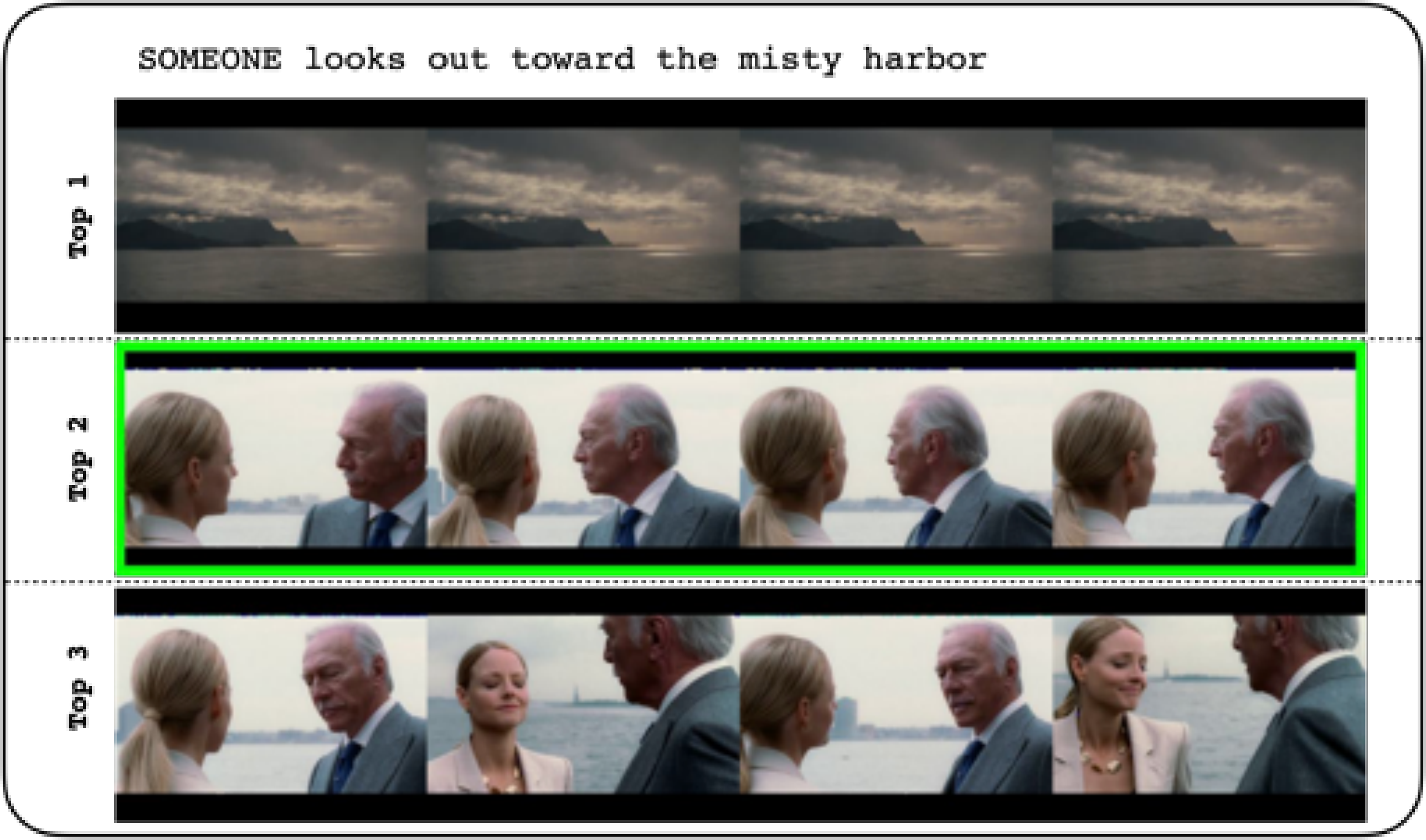}
\includegraphics[scale=0.19]{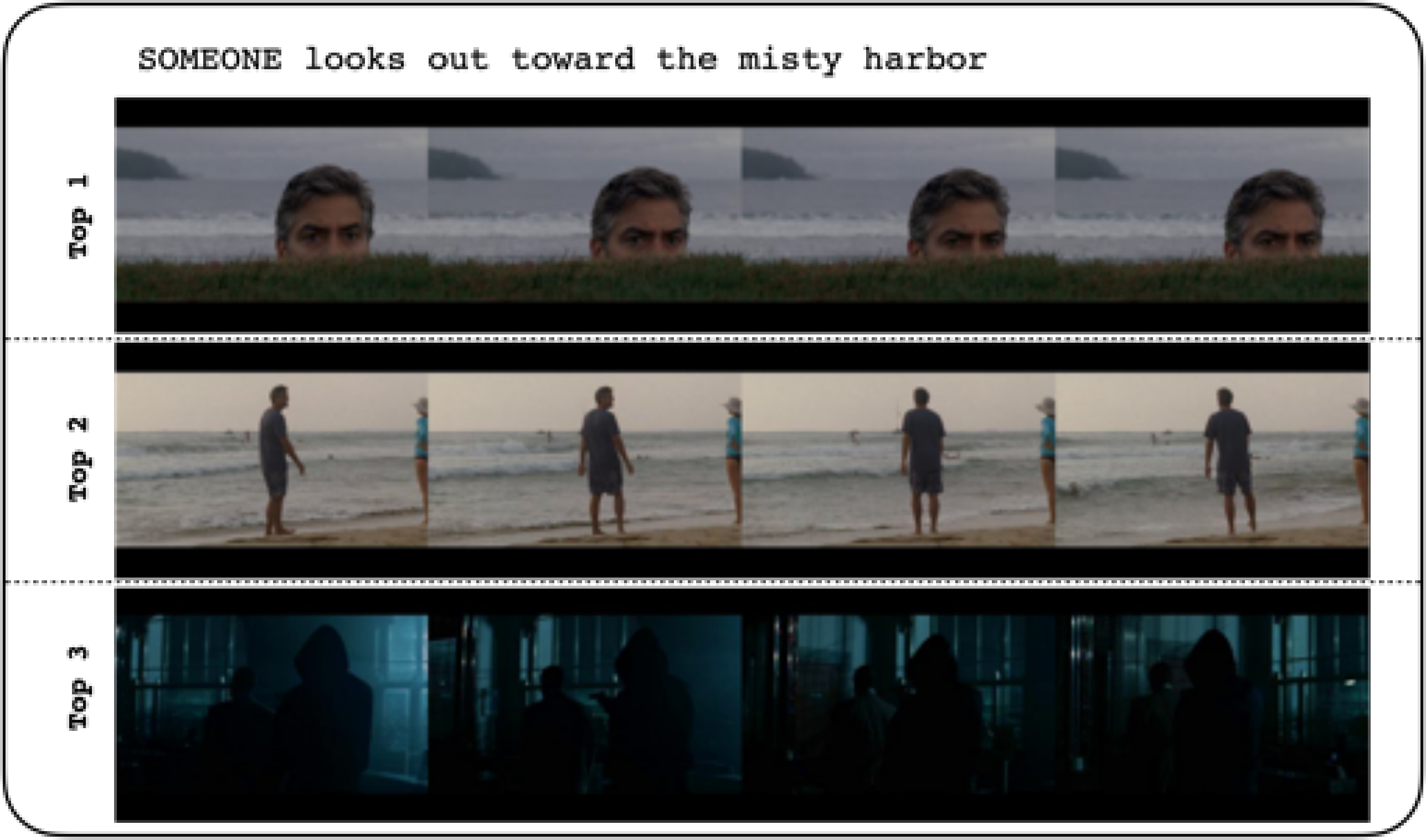}
\end{center}
\vspace{-1.1em}
\caption{\textbf{Qualitative evaluation of text-video retrieval on the LSMDC.} Retrieval examples for the proposed In-Style Model and zero-shot BLIP model. Each box shows the top-3 retrieved videos for a given text query. The correct video is highlighted with a green color.
}
\label{fig:lsmdc_retrieval_qual}
\end{figure*}

\paragraph{Text-video retrieval.} We also demonstrate qualitative results of text-video retrieval on the MSR-VTT (Figure~\ref{fig:msrvtt_retrieval_qual}), YouCook2 (Figure~\ref{fig:youcook_retrieval_qual}), DiDeMo (Figure~\ref{fig:didemo_retrieval_qual}), MSVD (Figure~\ref{fig:msrvtt_retrieval_qual}), and LSMDC (Figure~\ref{fig:lsmdc_retrieval_qual}) datasets. We found that the proposed In-Style model retrieves more semantically similar videos to a given query compared to the zero-shot BLIP model.

\section{Limitations}
\label{sec:limitations}

\begin{figure*}[!t]
\begin{center}
\includegraphics[scale=0.19]{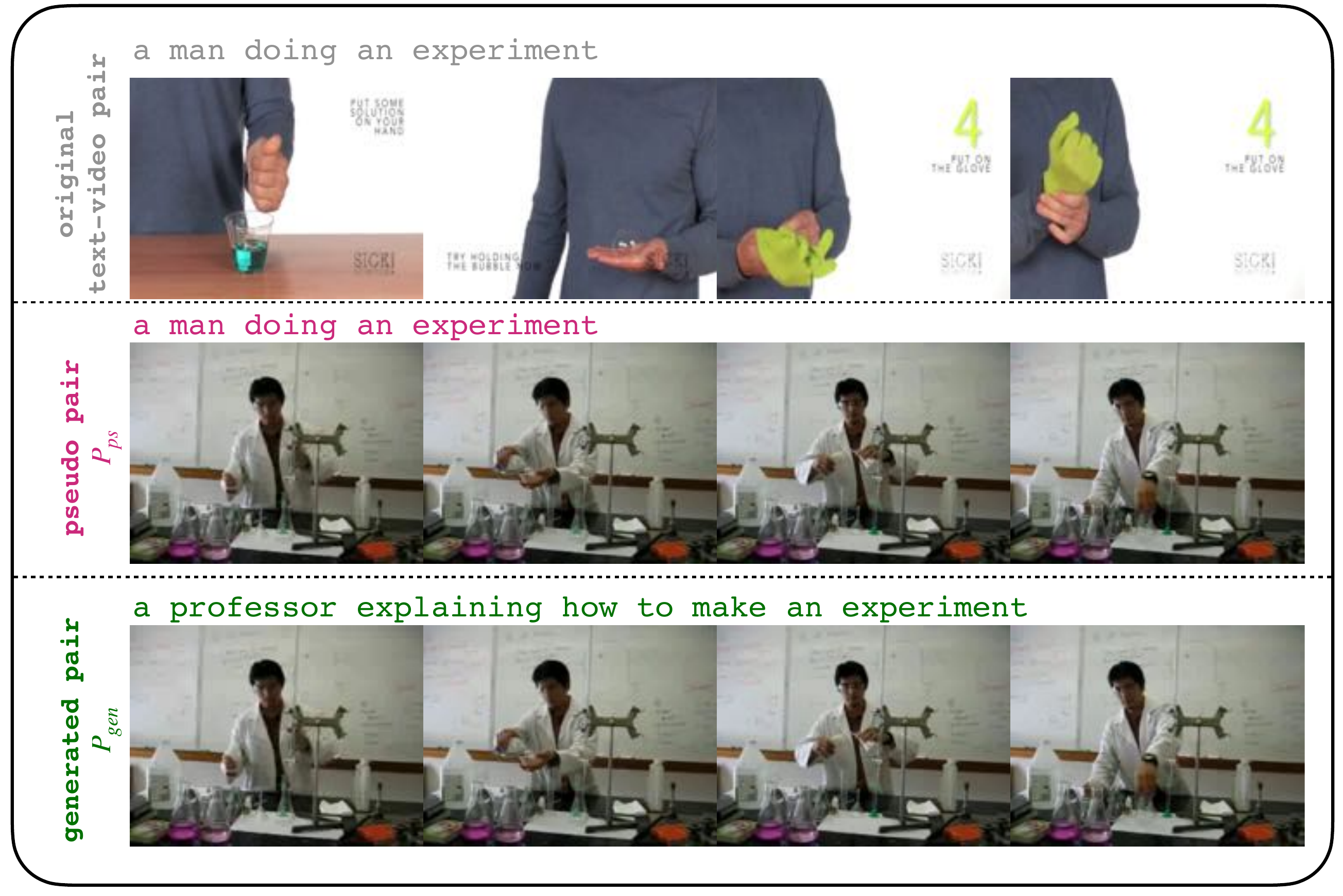}
\includegraphics[scale=0.19]{figs/msrvtt2.pdf}
\includegraphics[scale=0.19]{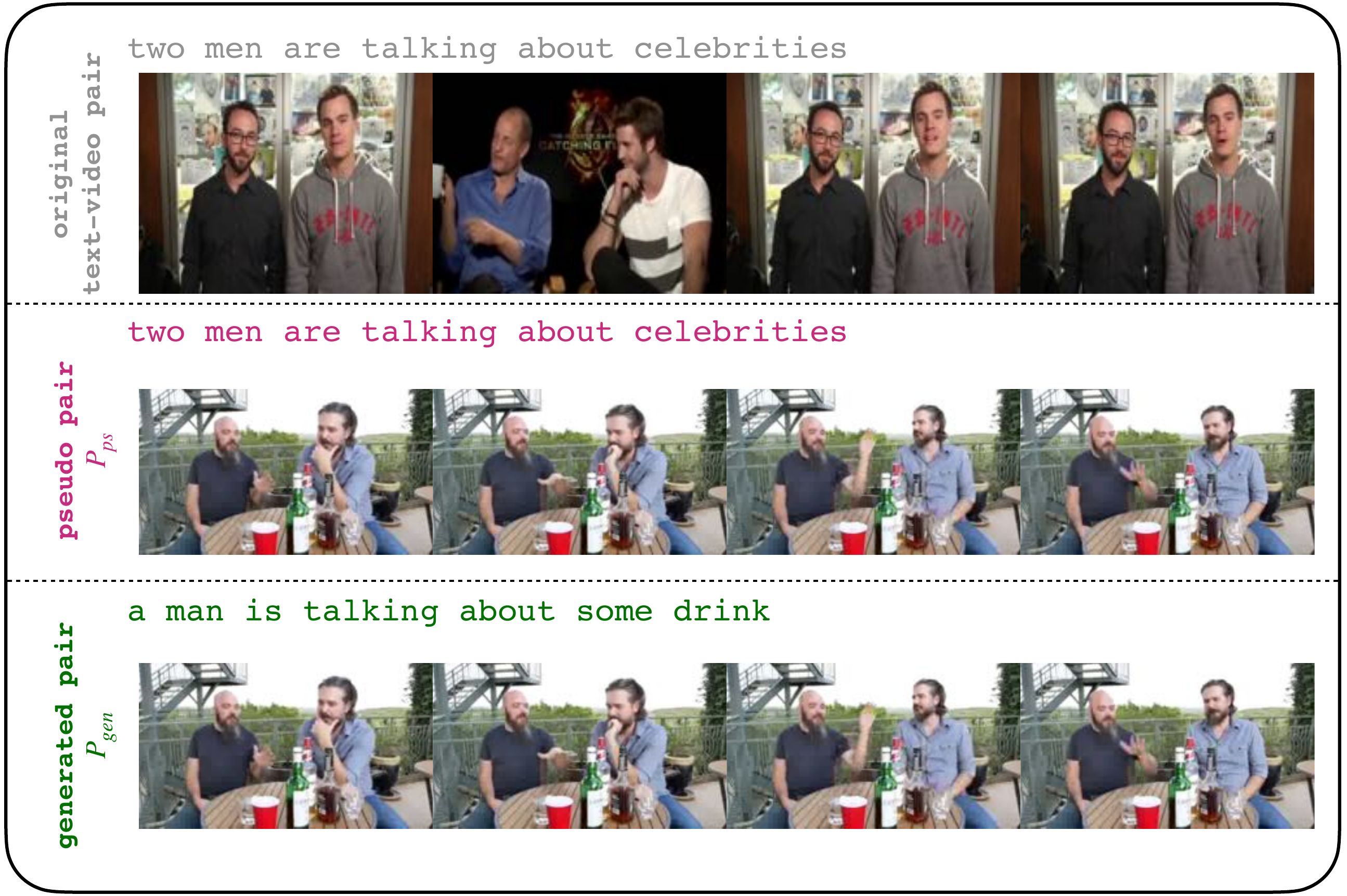}
\includegraphics[scale=0.19]{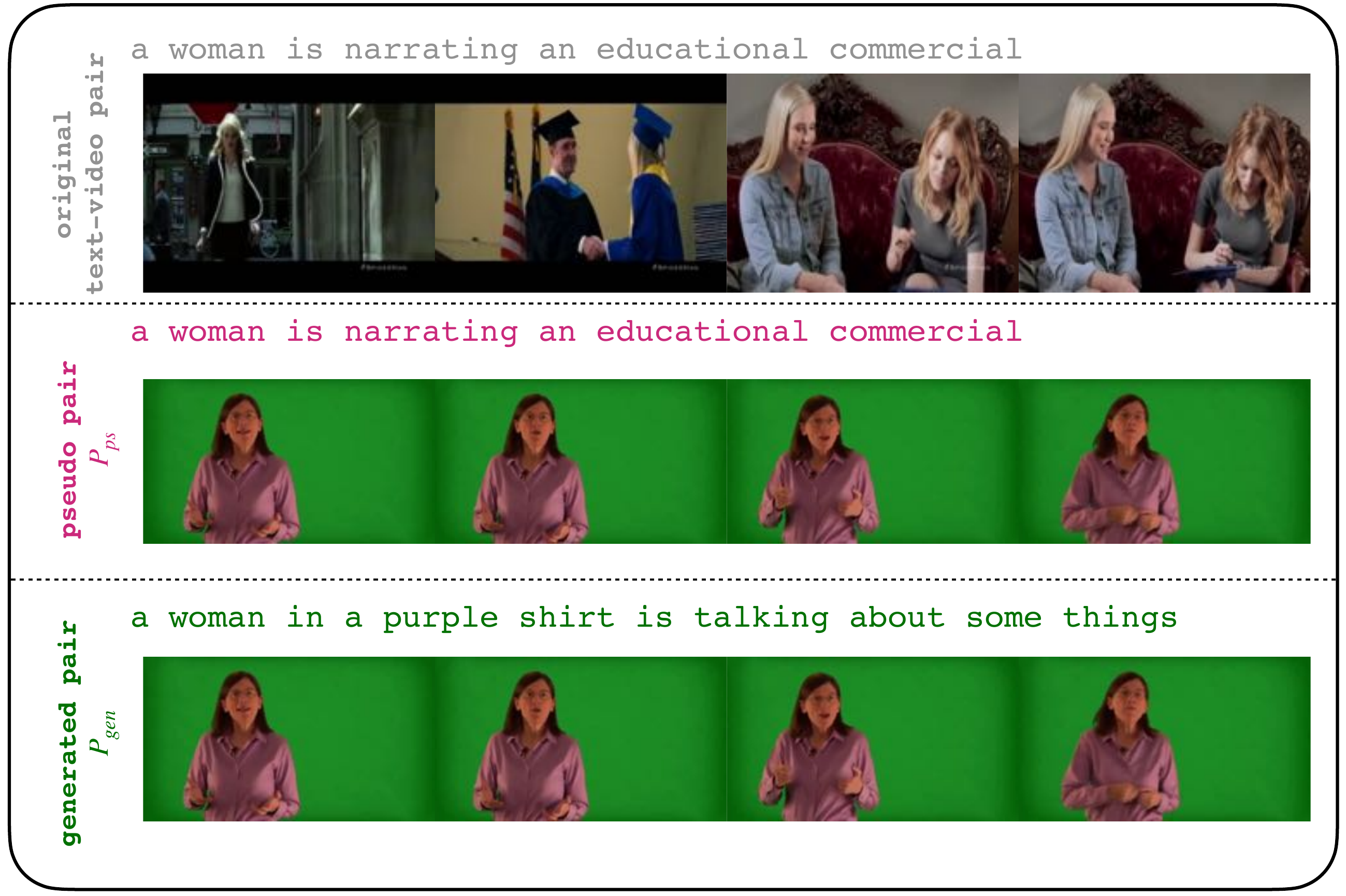}
\includegraphics[scale=0.19]{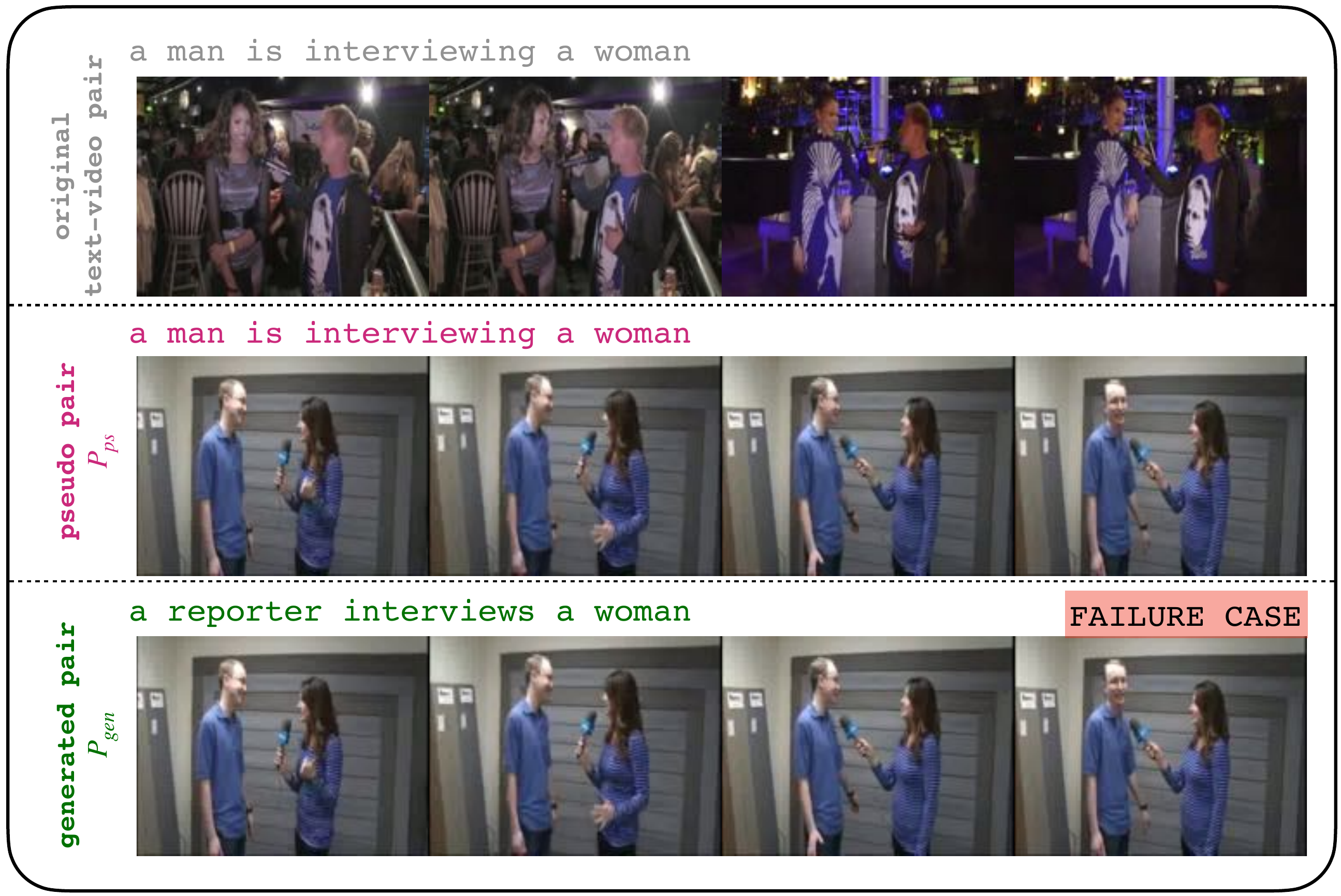}
\includegraphics[scale=0.19]{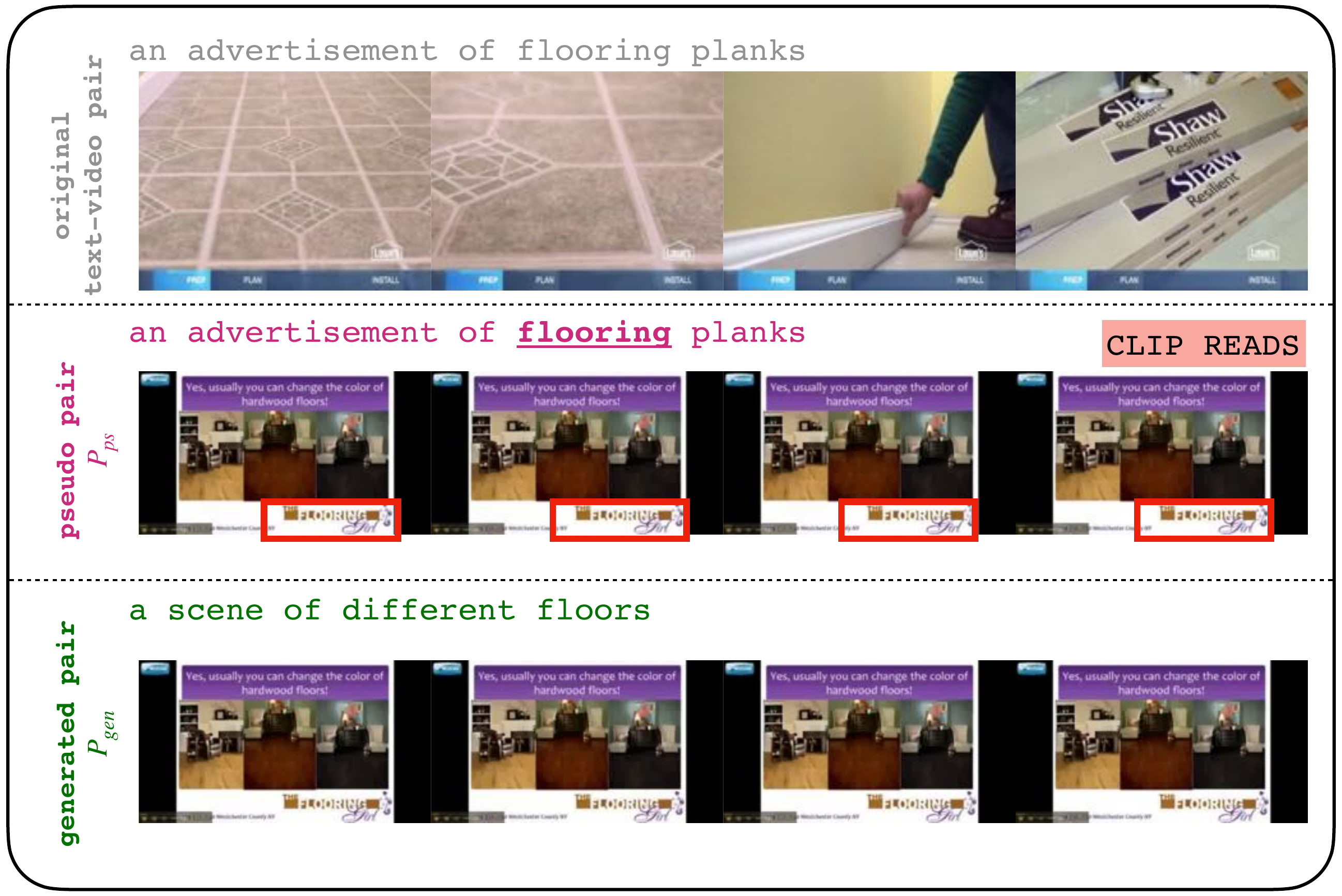}
\end{center}
\vspace{-1.1em}
\caption{ \small{\textbf{Qualitative evaluation of $P_{ps}$ and   $P_{gen}$ on the MSR-VTT.} First, a text query is matched with one of the videos (a pseudo pair $P_{ps}$), and then, after the style transfer step, for each video, a new caption is generated in the same style but with updated content (a generated pair $P_{gen}$)}
}
\label{fig:msrvtt_qual}
\end{figure*}

In this work, we rely on the pre-trained large image-language models such as CLIP~\cite{radford2021learning} and BLIP~\cite{li2022blip}. We consider this as an advantage and disadvantage at the same time. On one side, we show how to adapt such models to the input style of text queries, whereas, on the other side, we inherit all the biases that such models include~\cite{bommasani2021opportunities}. Moreover, CLIP has the property that it can read text from the images~\cite{materzynska2022disentangling}; therefore, matching or filtering steps could suffer from that because some unrelated text (e.g., advertisement) appears on the frames (e.g., see Figure~\ref{fig:msrvtt_qual}).

While our motivation is to avoid annotation costs for aligning text-video pairs, we still rely on a video collection; namely, we utilize a large-scale dataset of YouTube videos. However, we note that such videos are easy to collect as they are available on YouTube and are not preprocessed; therefore, these videos include various types of potential noise from the web, such as advertisements, camera motion, low-quality videos, and others. Moreover, even though the distribution of our input text queries is not the same as the distribution of web support videos, we empirically observe that there is always some overlap between the distributions. Hence, we do not assume that our In-Style method will be helpful when input queries and test sets are from absolutely different domains like medicine (input text queries) and wildlife (test). As we show in our experiments, we obtain the most gain when the input text queries and test sets are from the same distribution.


\begin{figure*}[!t]
\begin{center}
\includegraphics[scale=0.19]{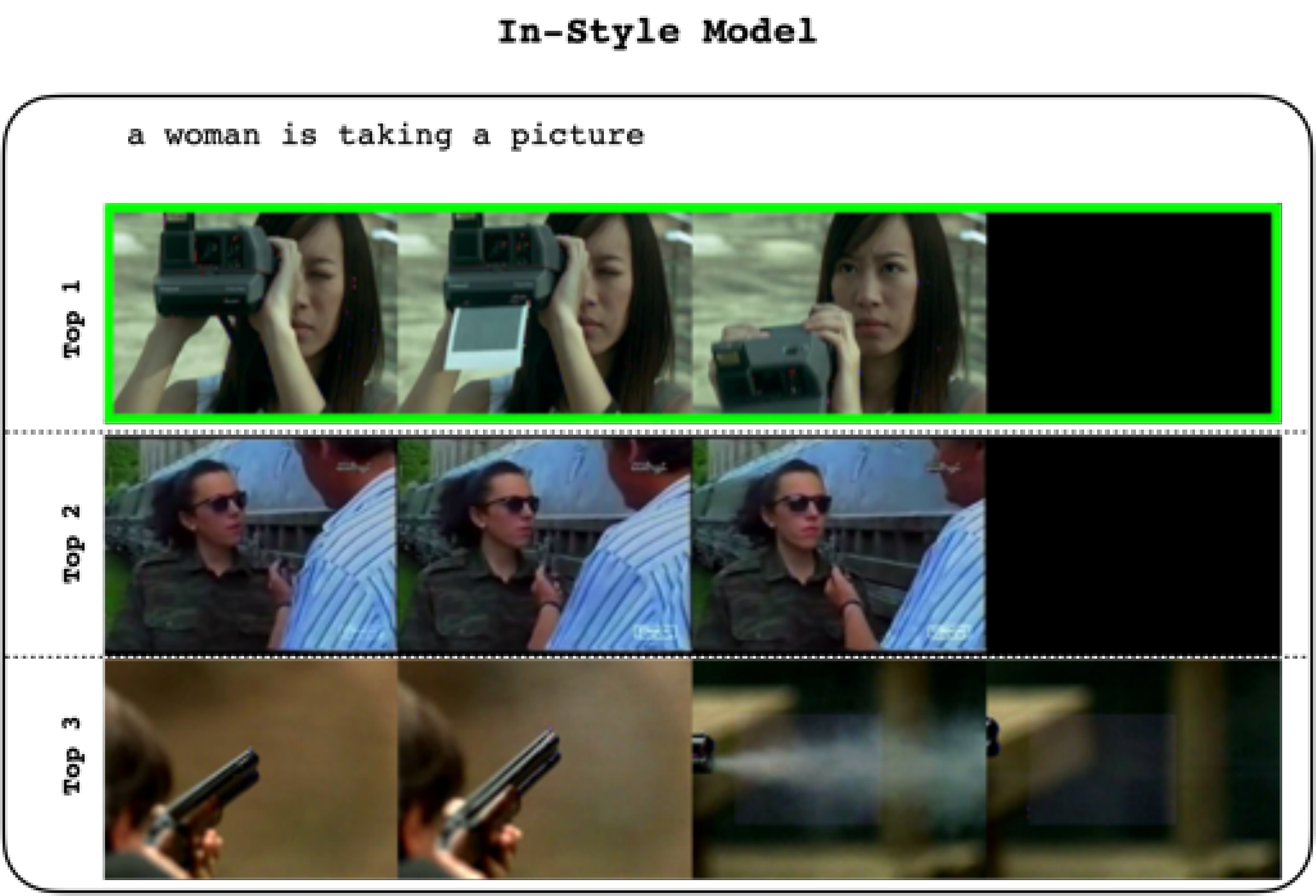}
\vspace{2mm}
\includegraphics[scale=0.19]{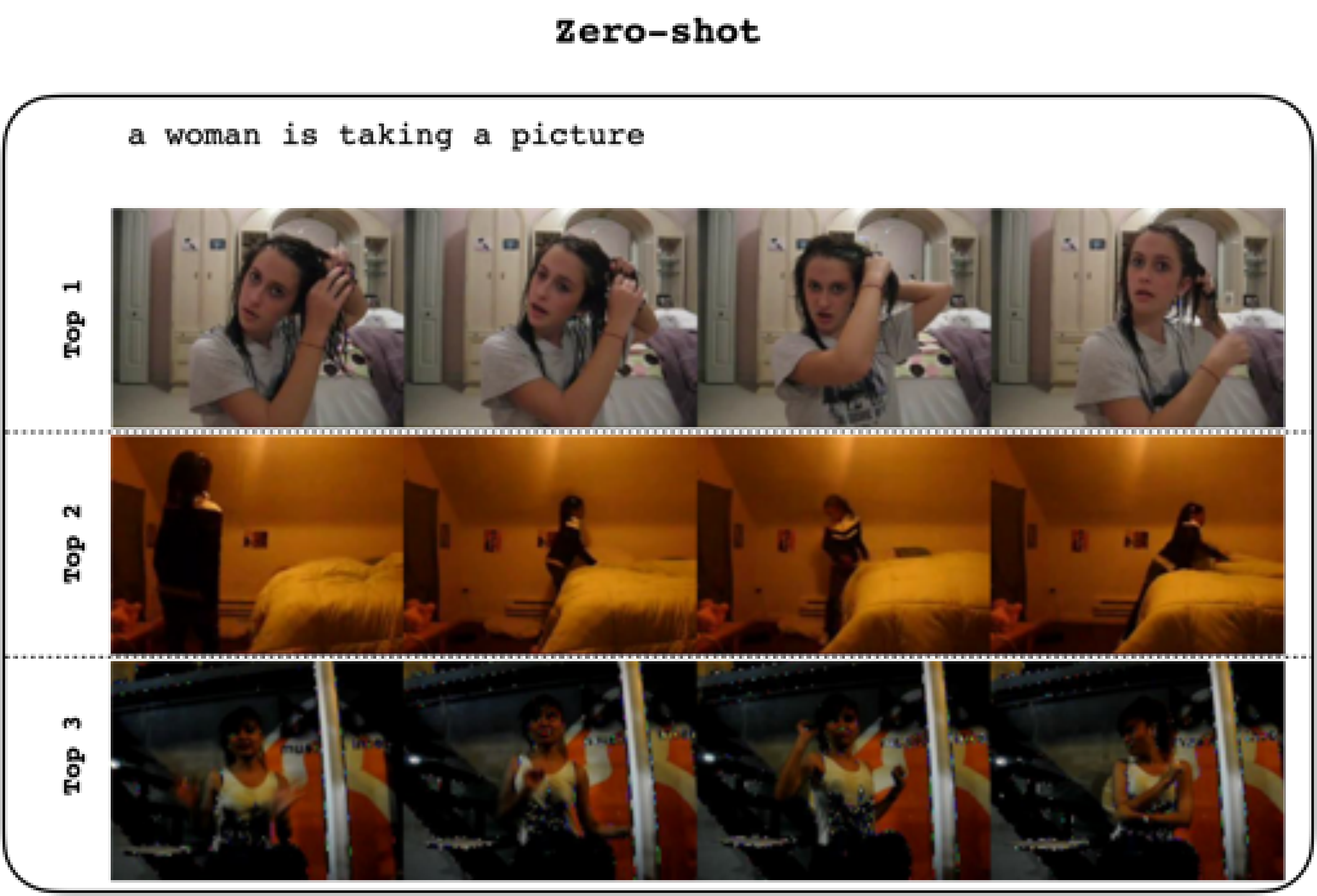}
\vspace{2mm}
\includegraphics[scale=0.19]{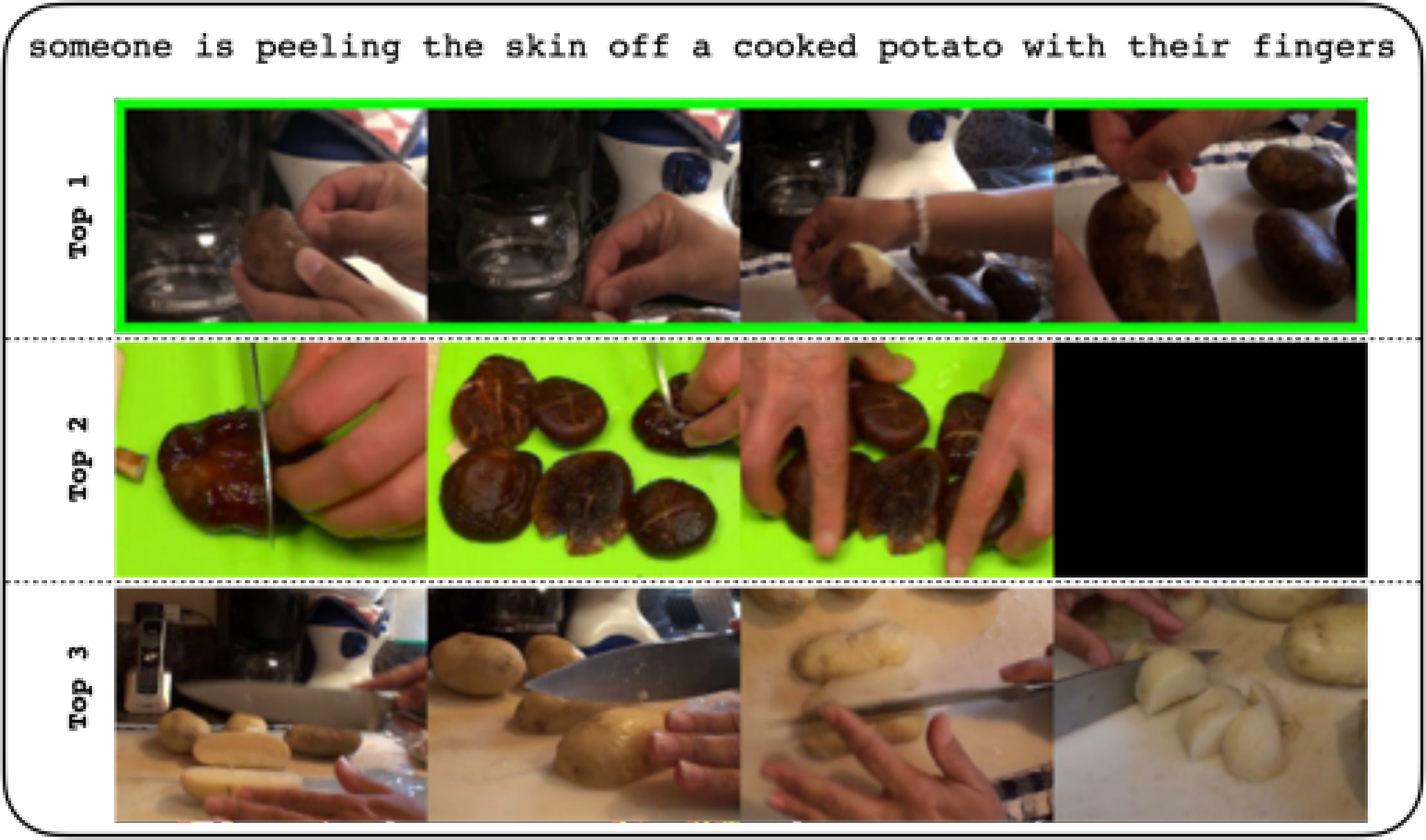}
\vspace{1mm}
\includegraphics[scale=0.19]{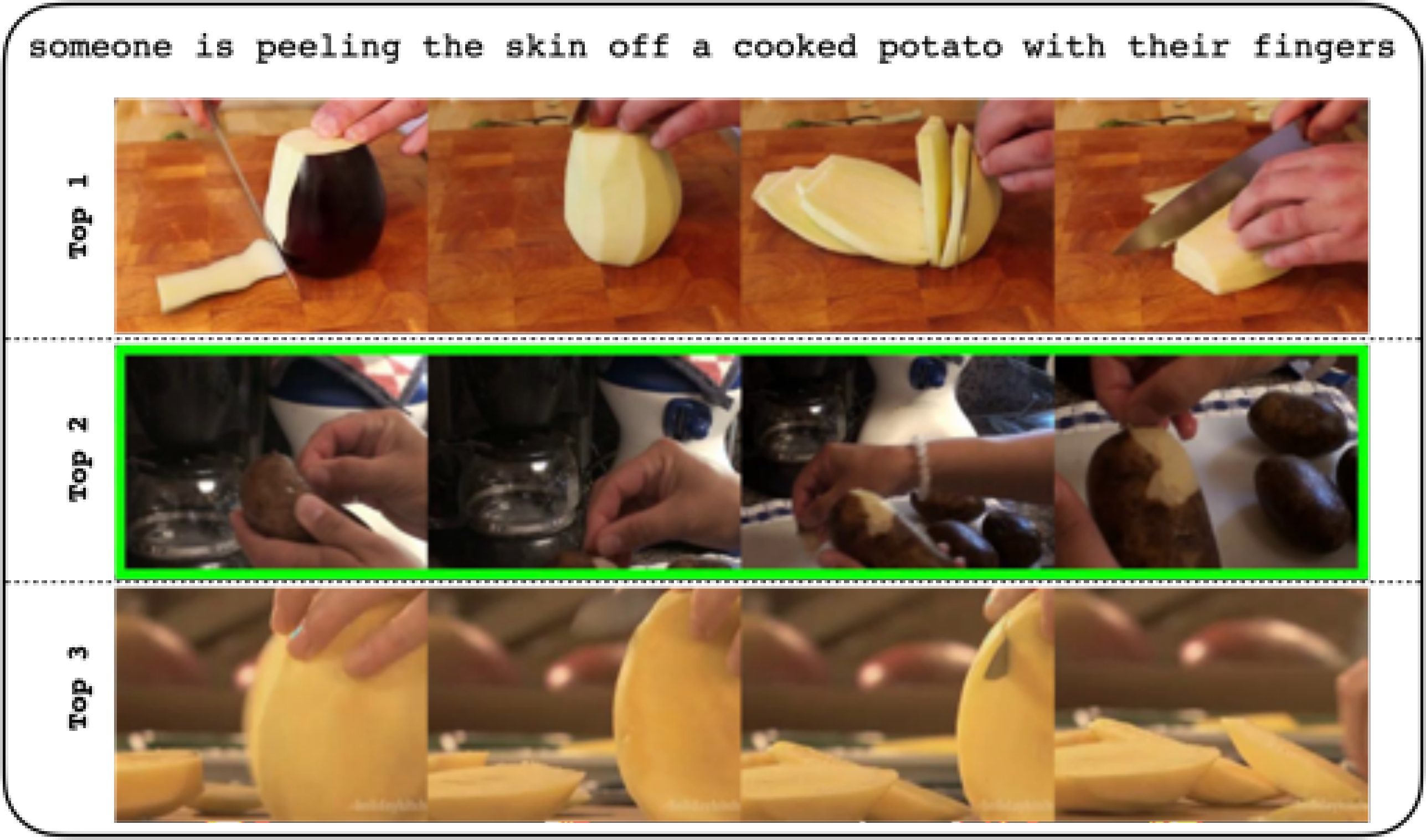}
\vspace{1mm}
\includegraphics[scale=0.19]{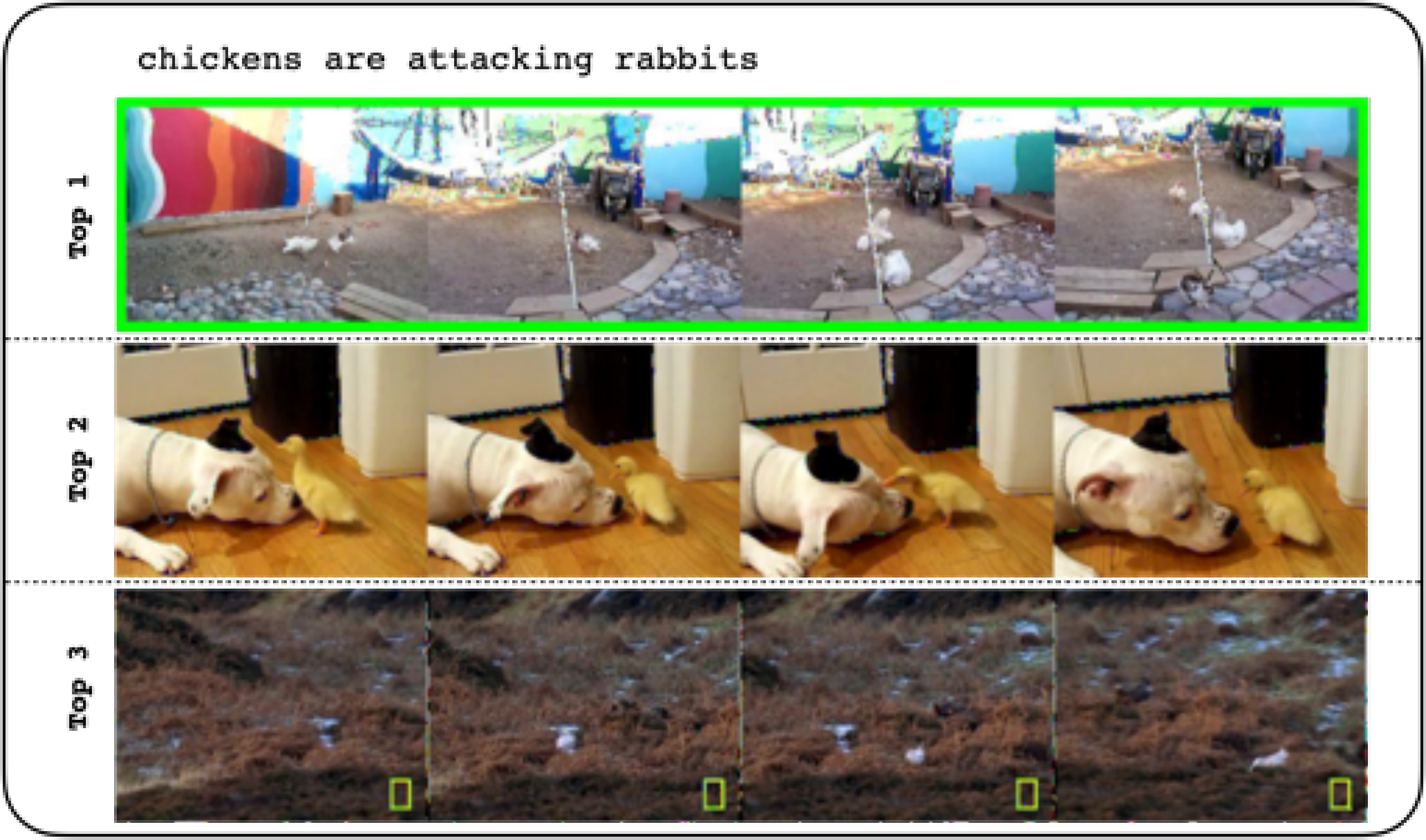}
\includegraphics[scale=0.19]{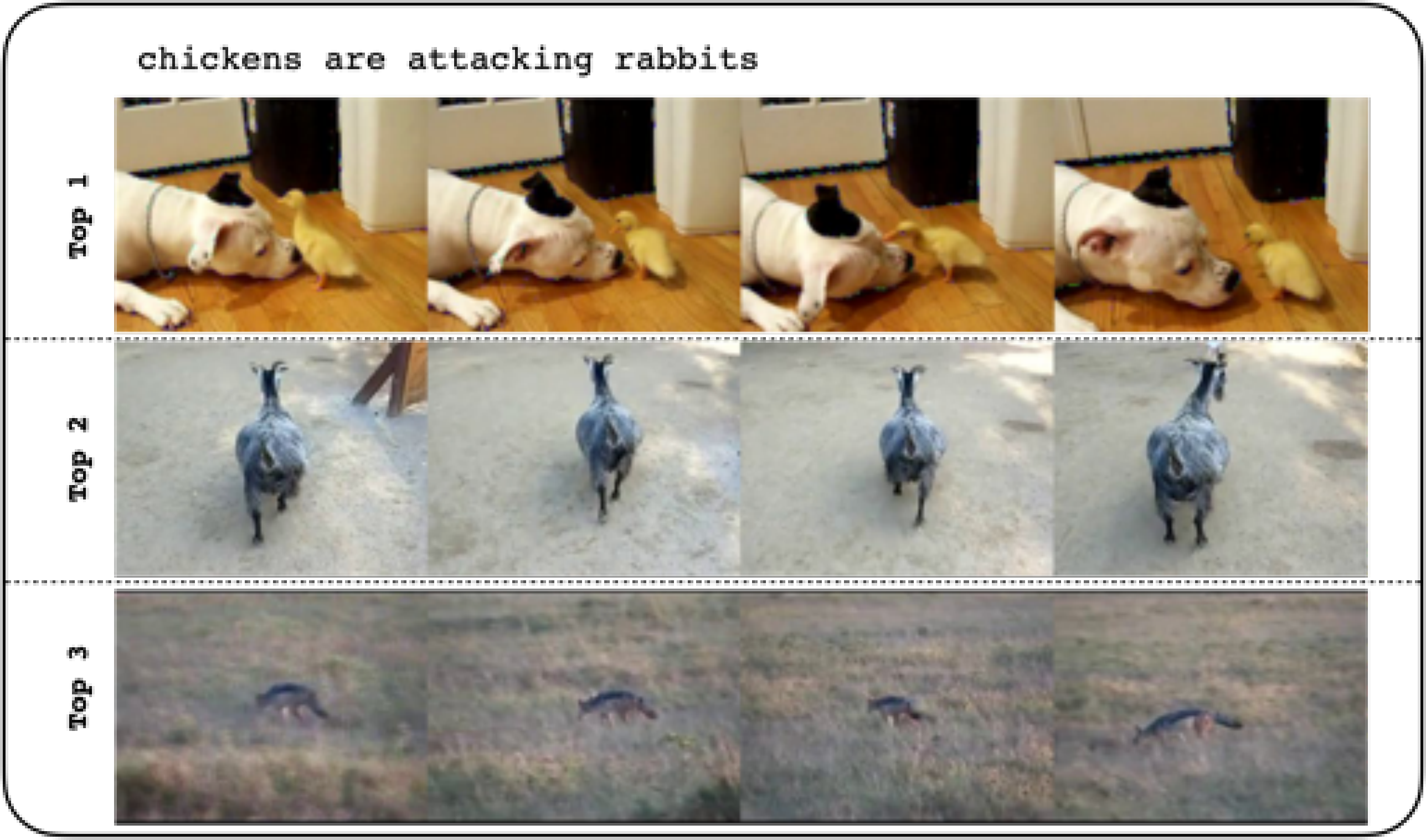}
\end{center}
\vspace{-1.1em}
\caption{\textbf{Qualitative evaluation of text-video retrieval on the MSVD.} Retrieval examples for the proposed In-Style Model and zero-shot BLIP model. Each box shows the top-3 retrieved videos for a given text query. The correct video is highlighted with a green color.
}
\label{fig:msvd_retrieval_qual}
\end{figure*}

\clearpage

\end{document}